\documentclass[10pt,journal,compsoc]{IEEEtran}
\usepackage[cmex10]{amsmath}
\usepackage{algorithmic}
\usepackage{algorithm}
\usepackage{array}
\usepackage{textcomp}
\usepackage{stfloats}
\usepackage{url}
\usepackage{verbatim}
\usepackage{graphicx}
\usepackage{hyperref}
\usepackage{color}
\usepackage{xcolor}
\usepackage{bbm}
\usepackage{booktabs}
\usepackage{multirow}
\usepackage{bm}
\usepackage{colortbl}
\usepackage{amssymb}
\usepackage{enumitem}
\newcommand{\fref}[1]{Fig. \ref{#1}}
\newcommand{\tref}[1]{Table \ref{#1}}
\newcommand{\eref}[1]{Eq. (\ref{#1})}
\ifCLASSOPTIONcompsoc
  \usepackage[nocompress]{cite}
\else
  \usepackage{cite}
\fi
\ifCLASSINFOpdf
\else
\fi
\ifCLASSOPTIONcompsoc
 \usepackage[caption=false,font=footnotesize,labelfont=sf,textfont=sf]{subfig}
\else
 \usepackage[caption=false,font=footnotesize]{subfig}
\fi
\begin{document}
%
\title{Unified Modality Separation: \\ A Vision-Language Framework for  Unsupervised Domain Adaptation}
%
%
%
%

\author{Xinyao Li,
        Jingjing Li*,~\IEEEmembership{Member,~IEEE,}
        Zhekai Du,
        Lei Zhu,
        and Heng Tao Shen,~\IEEEmembership{Fellow,~IEEE}
\IEEEcompsocitemizethanks{
\IEEEcompsocthanksitem This work was supported in part by the National Natural Science Foundation of China under Grant 52441801, in part by TCL Technology Innovation Funding SS2024105, and in part by the Fundamental Research Funds for the Central Universities (UESTC) under Grant ZYGX2024Z008.
\IEEEcompsocthanksitem Xinyao Li, Jingjing Li, Zhekai Du, and Heng Tao Shen are with University of Electronic Science and Technology of China, Chengdu 610054, China.
\IEEEcompsocthanksitem Lei Zhu is with Tongji University, Shanghai 200070, China.}
\thanks{* Corresponding author.}
\thanks{Manuscript received ; revised  .}}

%
%

\markboth{Journal of \LaTeX\ Class Files,~Vol.~13, No.~9, September~2014}%
{Shell \MakeLowercase{\textit{et al.}}: Bare Demo of IEEEtran.cls for Computer Society Journals}

\IEEEtitleabstractindextext{%
\begin{abstract}
Unsupervised domain adaptation (UDA) enables models trained on a labeled source domain to handle new unlabeled domains. 
Recently, pre-trained vision-language models (VLMs) have demonstrated promising zero-shot performance by leveraging semantic information to facilitate target tasks. By aligning vision and text embeddings, VLMs have shown notable success in bridging domain gaps. However, inherent differences naturally exist between modalities, which is known as \textit{modality gap}. Our findings reveal that direct UDA with the presence of modality gap only transfers modality-invariant knowledge, leading to suboptimal target performance. To address this limitation, we propose a unified modality separation framework that accommodates both modality-specific and modality-invariant components. During training, different modality components are disentangled from VLM features then handled separately in a unified manner.
At test time, modality-adaptive ensemble weights are automatically determined to maximize the synergy of different components. 
To evaluate instance-level modality characteristics, we design a modality discrepancy metric to categorize samples into modality-invariant, modality-specific, and uncertain ones. The modality-invariant samples are exploited to facilitate cross-modal alignment, while uncertain ones are annotated to enhance model capabilities. Building upon prompt tuning techniques, our methods achieve up to 9\% performance gain with 9 times of computational efficiencies. Extensive experiments and analysis across various backbones, baselines, datasets and adaptation settings demonstrate the efficacy of our design.
\end{abstract}

}

\maketitle

\IEEEdisplaynontitleabstractindextext

%
\IEEEpeerreviewmaketitle

\IEEEraisesectionheading{\section{Introduction}\label{sec:introduction}}

\IEEEPARstart{U}{nsupervised} domain adaptation (UDA) \cite{ganin2015unsupervised,saito2018maximum} transfers knowledge from a labeled source domain to a related but unlabeled target domain. The primary challenge in UDA is to mitigate the distribution shift between the source and target domains~\cite{ben2010theory}. Prior efforts either  optimize a criterion to minimize the  distribution discrepancy~\cite{long2015learning,long2017deep}, or employ adversarial learning~\cite{goodfellow2014generative} to learn domain-invariant features~\cite{ganin2015unsupervised,long2018conditional}. Another line of study resort to more capable  networks for better transferability~\cite{pmtrans,cdtrans}. While these methods have shown promising success, they have  concentrated on single vision modality, while neglecting the rich semantic structures in data. Recent advances in  vision-language models (VLMs)~\cite{clip,align} offer a promising solution. 
The contrastive learning in VLMs pulls closer paired image and text features, endowing them with rich knowledge on general semantic concepts. Such large-scale pretraining (e.g., CLIP \cite{clip} is trained on 400 million image-text pairs) also contributes  to the capability in closing domain gap, making them suitable for domain adaptation tasks.

\begin{figure}[!t]
  \centering
  \subfloat[Illustration of (i) conventional vision UDA, (ii) vision-language UDA, (iii) modality gap hindering adaptation on modality-specific data, and (iv) the proposed modality-aware adaptation concept.]{\includegraphics[width=0.475\textwidth]{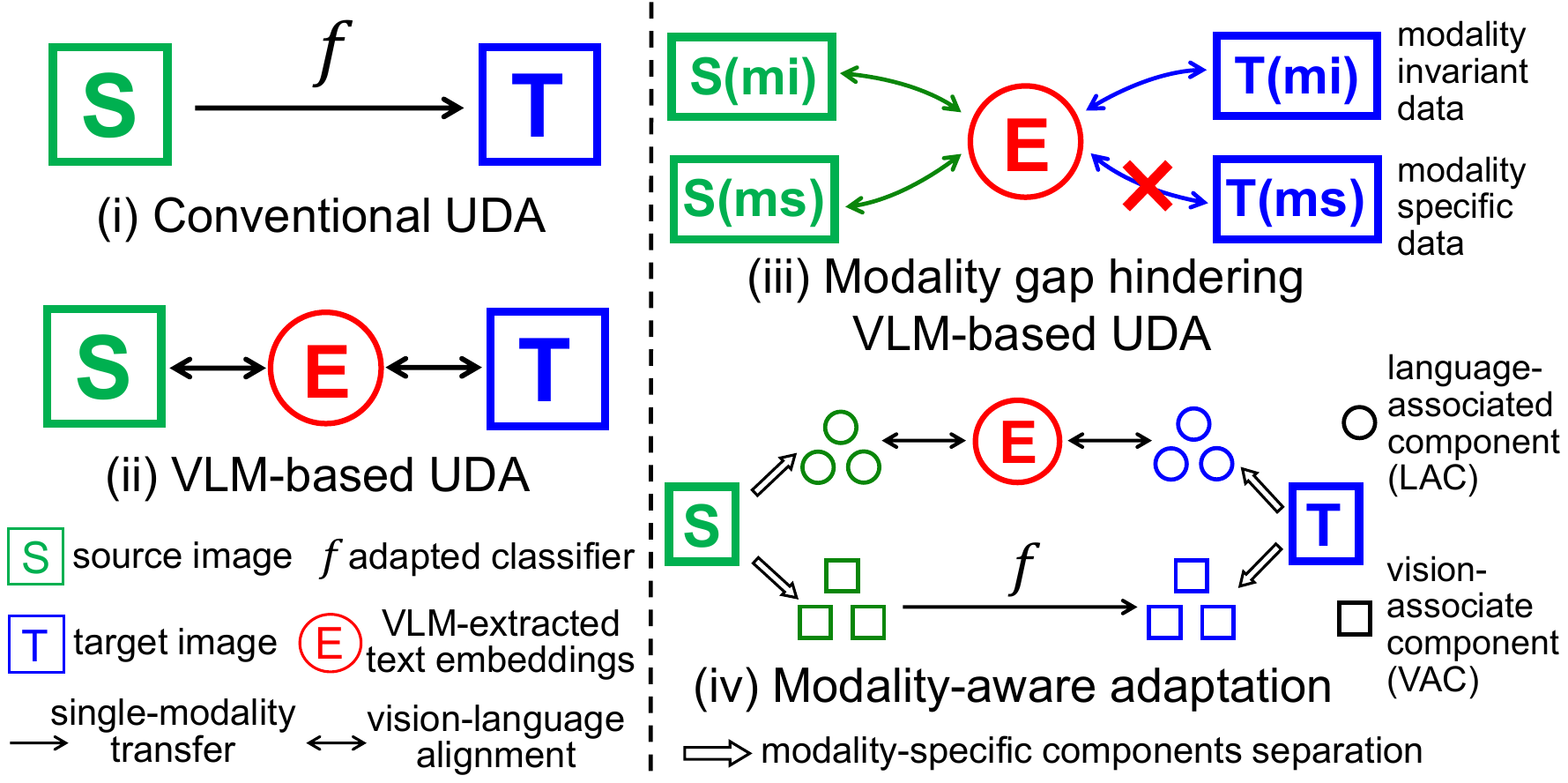}
  \label{fig1a}}
  \hfil
  \subfloat[Right: Examples on modality-specific information from OfficeHome-RealWorld. Left: Visualizations of the disentangled VAC and LAC.]{\includegraphics[width=0.475\textwidth]{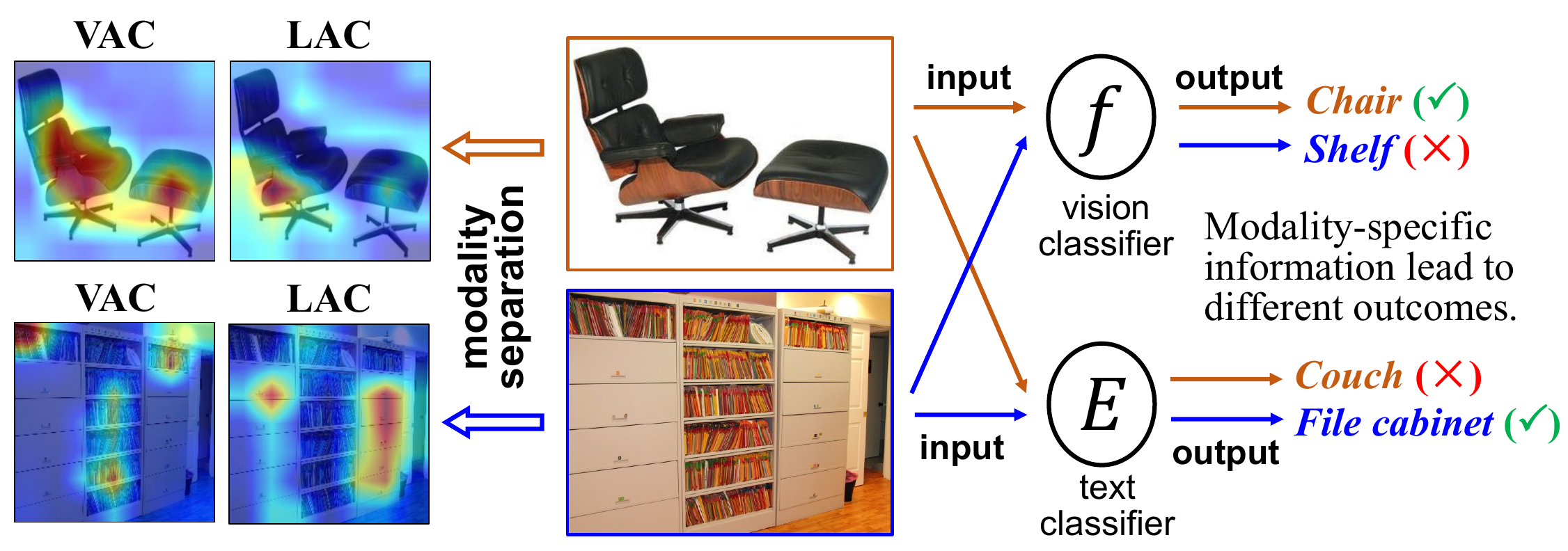}
  \label{fig1b}}
  \caption{Motivations and effects of our modality separation design. Vision-associated components (VAC), language-associated components (LAC) are inherent visual details and semantic cues in the image, each focused by VLM's modality-specific branches, respectively.}
  \label{fig1}
  \vspace{-14pt}
\end{figure}

Pioneering works on VLM-based UDA either adopt prompt tuning~\cite{lester2021power} to adapt text embeddings~\cite{du2024domain,daprompt}, or tune the pretrained parameters to match target distribution~\cite{padclip}. These methods all rely on similarities between vision and text embeddings to perform classification, as shown in \fref{fig1a}(ii). However, recent studies on the \textit{modality gap}~\cite{liang2022mind} phenomenon  reveal that, despite pretraining efforts, the vision and text features are distinctly distributed, each containing private modality-specific information.
\fref{fig1b}(right) offers clear examples on modality-specific knowledge. The vision and text classifiers are independently adapted to the target data\footnote{Vision classifier is adapted by linear-probing on pseudo labels from vision KMeans \cite{shot}. Text classifier includes prompts tuned on text pseudo labels following method in \cite{daprompt}.}, whose classification results vary across different modalities. The vision classifier excels at identifying visually straightforward items like a chair, while the text classifier better interprets semantically rich pictures like a file cabinet.
Such modality-specific information cannot be well aligned by current methods with the existence of modality gap, as illustrated in \fref{fig1a}(iii).
Rather than attempting to reduce \cite{menonvisual} or bridge \cite{zhang2024connect} the modality gap, we propose to bypass it with \textit{modality separation} on CLIP features, so that modality information vital for classification can be preserved. 
Specifically, a multimodal framework is designed to transfer and exploit modality-specific information for finer adaptation, as depicted in \fref{fig1a}(iv).

As fine-tuning CLIP parameters is expensive and may disturb the pre-aligned multi-modal space~\cite{gao2024clip}, we opt to the feature disentanglement and adaptation with a Unified Modality Separation (UniMoS) framework. 
During training, the modality separation network disentangles the pre-extracted vision features into 
vision- and language-associated components (VAC, LAC) to capture respective modality-specific characteristics. \fref{fig1b}(left) visualizes disentangled VAC and LAC using GradCAM \cite{selvaraju2017grad}. VAC and LAC focus on distinct image areas that represent modality-specific information.
Such distinction is ensured by an orthogonal regularization loss.
LAC and VAC are then handled separately in a unified manner, with a learnable weight to balance their contributions. 
To transfer source modality-specific knowledge to target domain, VAC and LAC are aligned separately with a modality discriminator.
During inference, we combine the strengths of each modality with a Modality-adaptive Ensemble (MaE) mechanism. MaE first evaluates ensemble thresholds for target samples, then determines optimal weights by approximating the vision and text distributions. 
These weights can adapt dynamically to various training phases and tasks automatically.

\begin{figure}[!t]
  \centering
  \includegraphics[width=0.46\textwidth]{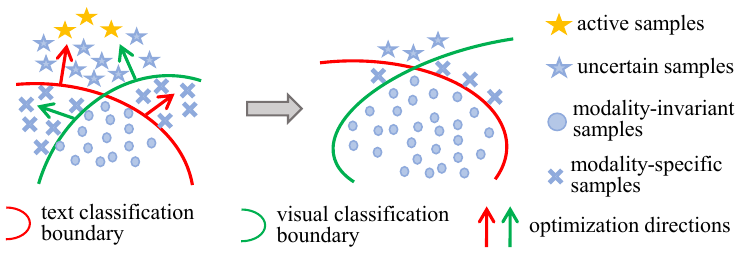}
  \vspace{-10pt}
  \caption{Illustration of the MDI metric and model optimization goal. Both modality classifiers are optimized towards the active samples and each other, achieving maximum harmony on target samples.}
  \label{fig2}
  \vspace{-9pt}
\end{figure}

To handle samples with various modality gap, we further design a Modality DIscrepancy (MDI) metric to categorize target data according to their cross-modal prediction confidences and consistencies.
The resultant MDI categories are conceptualized in \fref{fig2}. The modality-invariant samples help learn more robust decision boundaries. On the other hand, modality-specific information is extracted and utilized by the modality ensemble processes. 
Uncertain samples might interfere the modality separation and adaptation processes. The absence of target labels makes them hard to learn. Therefore, we draw inspirations from active domain adaptation (ADA)~\cite{xie2022active,du2023diffusion} to annotate these hard samples, which enhances the adaptation process with limited annotation costs.
\fref{fig2} indicates that exploitation of annotated uncertain samples extend VLM's representation boarders. The main contributions of this work can be summarized as: 
\begin{enumerate}[leftmargin=21pt]
\item We identify the negative effects of the modality gap on domain adaptation tasks, emphasizing the modality-specific information that is often overlooked by existing VLM-based UDA methods. 
\item We propose a  unified modality separation framework to handle both modality-specific and modality-invariant components for comprehensive UDA.
\item We present a novel MDI score that quantifies instance-level modality characteristics, which not only enhances training but also provides insights and solutions for multimodal active learning. 
\item Our method is compatible with various prompt tuning techniques and   adaptation settings including UDA, ADA, source-free ADA and multi-source DA. Extensive evaluations on various benchmarks, backbones and prompt tuning methods demonstrate the efficacy of our approach.  Comprehensive  analysis further demonstrate the validness and efficacy of our design.
\end{enumerate}

This work extends UniMoS\footnote{https://github.com/TL-UESTC/UniMoS} in our previous paper~\cite{unimos} to UniMoS++ with substantial technical contributions and experimental analysis. (1) We update the fixed test-time ensemble weight in UniMoS with a modality-adaptive ensemble weight for better flexibility and practicability. (2) We extend the modality-level alignment process of UniMoS to customized adaptation strategies tailored to instance-level modality characteristics, which are evaluated by a novel MDI metric. 
(3) UniMoS++ is evaluated on various adaptation settings including multi-source DA, ADA, source-free ADA, and OOD generalization, achieving up to 9\% performance gain with 9 times of computational efficiencies. 
(4) UniMoS++ is compatible with existing prompt-tuning and parameter-efficient fine-tuning techniques, which offers great flexibility and applicability for adapting VLMs. 

\section{Related Work}
\subsection{Unsupervised Domain Adaptation}
Unsupervised domain adaptation (UDA) is a practical problem setting in transfer learning~\cite{pan2009survey}. The main challenge in UDA is to match representations across the source and target domain. Discrepancy-based approaches~\cite{long2015learning,long2017deep,li2020maximum} design divergence metrics to explicitly measure and minimize the distribution gap. Popular metrics include MMD~\cite{yan2017mind}, MDD~\cite{li2020maximum}, etc. Adversarial-based methods~\cite{ganin2015unsupervised,long2018conditional} learn domain-invariant features implicitly via a min-max game between the feature extractor and domain discriminator. To ensure consistency among feature and semantic spaces, recent works have focused on learning and exploiting  target data structures~\cite{yang2021exploiting,yang2023trust,shot}. 
With developments in model scale and capacity, some works adapt larger vision models like vision transformer (ViT)~\cite{vit} and its variants. 
SSRT~\cite{ssrt} perturbs target domain data to refine ViT, and adaptively adjusts training configurations to prevent model collapse. PMTrans~\cite{pmtrans}  adapts SwinTransformer~\cite{liu2021swin} by adopting patch mixup as an intermediate domain bridge. CDTrans~\cite{cdtrans} aligns attention scores in features extracted by DeiT~\cite{deit}. Vision-language foundation models like CLIP~\cite{clip} enable adaptation in text modality, further boosting UDA performance. Taking inspiration from prompt tuning~\cite{lester2021power}, DAPrompt~\cite{daprompt} learns class-wise domain-specific and domain-agnostic knowledge for text embeddings. DAMP~\cite{du2024domain} learns domain-invariant visual and textual prompts from mutual-prompting between visual and textual representations. As an alternative, PADCLIP~\cite{padclip} tunes pretrained CLIP parameters with adjusted learning rate to prevent catastrophic forgetting. UniMoS~\cite{unimos} explicitly disentangles CLIP-extracted vision features for modality-aware adaptation and prediction, achieving fine adaptation effects with low computation overheads. 

\subsection{Active Domain Adaptation}
Active domain adaptation (ADA) enhances UDA by actively labeling a small amount of target data.  TQS \cite{tqs} proposes to evaluate sample informativeness by transferable committee, transferable uncertainty, and transferable domainness. EADA \cite{eada} leverages energy-based models \cite{lecun2006tutorial} to detect out-of-distribution target samples. DUC \cite{duc} and DAPM \cite{dapm} propose to first calibrate the model against distribution shift before uncertainty estimation. DiaNA \cite{diana} partitions target samples into four categories according to certainty and consistency, and trains a Gaussian mixture model to sample unlabeled data. LADA \cite{lada} exploits the neighborhood information of active samples by augmenting confident neighbors in a class-balanced manner. Recently, some researches combine source-free UDA \cite{shot} with ADA, proposing source-free ADA (SFADA) \cite{wang2023mhpl, li2022source}. SFADA methods do not directly access source data during adaptation, but are provided a source-pretrained model, making them practical in data-sensitive scenarios. All forementioned methods are designed for vision models like ImageNet-pretrained ResNet \cite{he2016deep}, thus cannot understand the precise annotation needs for VLMs.

\subsection{Vision-language Models}
Trained on large-scale multimodal datasets, vision-language models (VLMs)~\cite{clip,align} have shown great generalization abilities. For example, CLIP~\cite{clip} is trained on 400 million text-image pairs while ALIGN~\cite{align} leverages more than one billion text-image pairs. The majority of researches adapt VLMs to the target dataset in a few-shot manner. Similar to prompt tuning~\cite{lester2021power}, CoOp~\cite{coop} propose to learn textual embeddings for each category instead of manually designing prompt descriptions. CoCoOp~\cite{cocoop} improves CoOp by conditioning on instances. MaPLe~\cite{maple} learns multimodal embeddings by injecting learnable  parameters into the vision branch. DenseCLIP~\cite{denseclip} adapts prompts posteriorly for dense prediction. Considering the large amount of parameters in VLMs, another line of studies learn adapters~\cite{houlsby2019parameter} to achieve parameter-efficient knowledge transfer~\cite{sung2022vl,clip-adapter}. To further reduce training and transfer costs, some methods build key-value cache models~\cite{grave2017unbounded} on vision and text features to achieve training-free adaptation of VLMs~\cite{tip-adapter,udandarao2023sus}. 
Despite the contrastive pretraining efforts, Liang \textit{et al.}~\cite{liang2022mind} reveal the existence of modality gap between vision and text features, which leads to imperfect alignment of multimodal features. Follow-up works investigate the root cause of modality gap~\cite{fahim2024its}, build better latent spaces with the presence of modality gap~\cite{ramasinghe2024accept}, or attempt to bridge the modality gap~\cite{zhang2024connect}. Our work aim to achieve modality-aware adaptation by protecting modality-specific information.

\begin{figure*}[!t]
  \centering
  \includegraphics[width=0.89\textwidth]{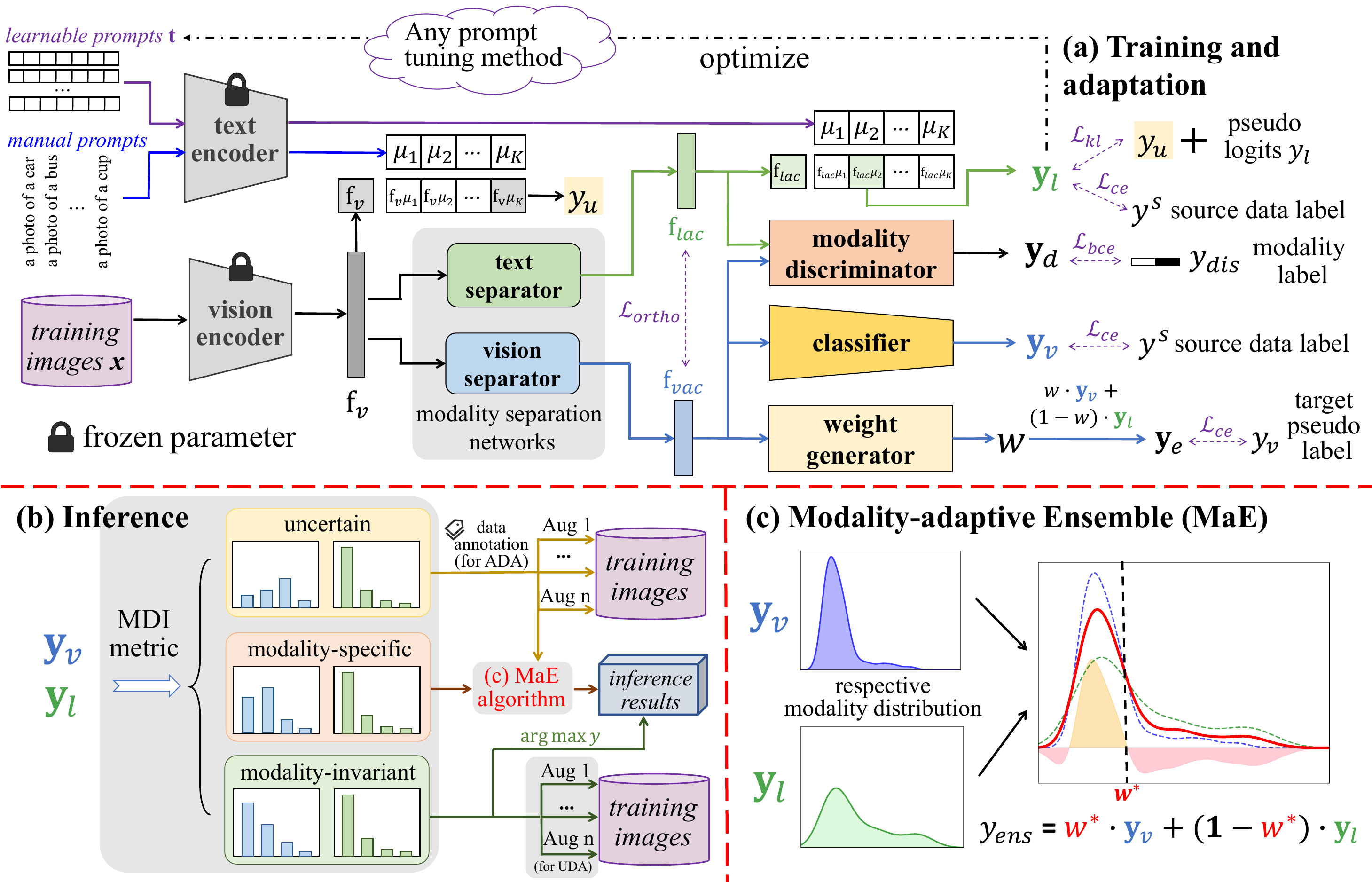}
  \vspace{-5pt}
  \caption{Illustration of UniMoS++ framework. (a) Training procedure of UniMoS++. CLIP-extracted vision features are disentangled into LAC ($\mathrm{f}_{lac}$) and VAC ($\mathrm{f}_{vac}$) by the modality separation network. The vision classifier generates vision  outputs from VAC, while text  outputs are the similarities between LAC and text embeddings of learnable prompts. The modality discriminator aligns LAC and VAC from the source and target domain. A weight generator learns train-time ensemble weights to balance contributions of different modalities. (b) MDI scores are computed from each pair of modality outputs. UN samples are annotated and augmented to complement the training set in ADA. MS and MI samples are optimally combined by test-time ensemble. MI samples are augmented to ensure semantic consistency during training. (c) MaE algorithm. Modality ensemble thresholds are calculated and sorted to approximate modality distributions, which generates the test-time ensemble weight.}
  \label{fig3}
  \vspace{-8pt}
\end{figure*}

\section{Method}
In this study, we focus on $K-$category classification problems.
In UDA, we have access to a labeled source domain $\mathcal{D}^s=\{(x_i^s,y_i^s)\}_{i=1}^{N^s}$ with $N^s$ samples and unlabeled target domain $\mathcal{D}^t=\{(x_i^t)\}_{i=1}^{N^t}$ with $N^t$ samples. The data distribution of source and target samples are different: $P(X^s) \ne P(X^t)$, and the goal of UDA is to learn a function $f: X^t \rightarrow Y^t$ to accurately predict target labels under domain shift. 
ADA introduces additional labeling on certain target samples $\mathcal{T}_L$, where $|\mathcal{T}_L| \ll N^t$.
ADA aims to find and exploit the most informative $\mathcal{T}_L$  to help learn a better  $f$.

\subsection{Framework Overview}
\label{prel}
We adopt CLIP~\cite{clip} as our foundation VLM. 
CLIP features a vision encoder and a text encoder. 
The vision encoder takes an \textit{image} as input and produces corresponding vision feature $\mathrm{f}_v$. For the text branch, CLIP provides a manual prompt template for zero-shot inference: \textit{A photo of a [CLASS]}, where \textit{[CLASS]} denotes the category to describe. Following \cite{padclip}, we construct manual prompts $\mathbf{t}=\{t_i\}_{i=1}^K$ for the domain adaptation task as: \textit{A [DOMAIN] photo of a [CLASS$_i$]}, where \textit{[DOMAIN]} indicates domain specifics (e.g., real) and $i$ is category index.
The text encoder takes these \textit{text} prompts as input and produces text feature $\mu_i$ for class $i$.
Prompt tuning in VLMs~\cite{coop} presents an alternative approach to obtain prompts. By constructing $t_i = [V]_1,...,[V]_M[CLASS_i]$ where $[V]_k$ are learnable parameters, prompts can be optimized towards the target domain. With vision feature $\mathrm{f}_v$ of an image, and text features $\{\mu_i\}_{i=1}^K$ of all categories, we compute the cosine similarities between the vision  and text features as CLIP classification probability:
\begin{align}
    P({y}=c) = \frac{\mathrm{exp}(\cos(\mathrm{f}_v, \mu_c)/\tau)}{\sum_{i=1}^K \mathrm{exp}(\cos(\mathrm{f}_v, \mu_i)/\tau)},
    \label{zero-shot}
\end{align}
where $c$ is the investigated categories and $\tau$ is temperature.

As discussed in Sec.\ref{sec:introduction}, classification results from cross-modal similarity in \eref{zero-shot} might  be unreliable due to modality gap, especially for OOD images \cite{padclip}. To tackle this, we conceptualize vision features as a composite of vision-associated components (VAC) and language-associated components (LAC): $\mathrm{f}_v = \{\mathrm{f}_{vac}, \mathrm{f}_{lac}\}$. LAC are text-correlated components introduced by the multimodal pretraining, which include semantic information that connects both modalities. VAC are vision-specific features extracted by the vision backbone.  By explicitly disentangling VAC and LAC for modality-specific prediction results $\mathbf{y}_{v}$ and $\mathbf{y}_{l}$, we can bypass the negative effects of modality gap. Finally, the classification results are integrated for cross-modality inference of target data:
\begin{equation}
  y_{ens} = w^* \cdot \mathbf{y}_{v} + (1-w^*) \cdot \mathbf{y}_{l},
  \label{ensemble1}
\end{equation}
where $w^*$ is test-time weight combining the  strengths of VAC and LAC. \fref{fig3}(a) presents UniMoS++ overviews. 

\subsection{Modality Separation Networks}
\label{sec_2}
Modality separation networks is the core of UniMoS++. The networks consist of  dual separation branches that project vision features into VAC and LAC. Denote the vision and text separator as $Q_{v}$ and $Q_{t}$, we obtain the separated components: $\mathrm{f}_{vac}=Q_{v}(\mathrm{f}_v)$, $\mathrm{f}_{lac}=Q_{t}(\mathrm{f}_v)$.  Taking inspiration from domain separation networks~\cite{bousmalis2016domain}, we apply an orthogonal loss to ensure the distinction between VAC and LAC:
\begin{equation}
  \mathcal{L}_{ortho} = ||\mathrm{diag}(\mathbf{F}_{lac}^{s} \,  {\mathbf{F}_{vac}^{s}}^{\top }) ||_{F}^{2} + ||\mathrm{diag}(\mathbf{F}_{lac}^{t} \, {\mathbf{F}_{vac}^{t}}^{\top } )||_{F}^{2}, 
\label{l_ortho}
\end{equation}
where matrices $\mathbf{F} \in \mathbb{R}^{B \times d_v}$ are batched VAC and LAC from both domains, and $\mathrm{diag}(\cdot )$ takes the diagonal elements of a matrix. Unlike \cite{bousmalis2016domain} that reduces similarities among all sample pairs, our $\mathcal{L}_{ortho}$ enforces separation  between \textit{paired} VAC and LAC components. This design allows modality disentanglement while maintaining original cross-modal correspondences.
Different modality outputs are then generated from VAC and LAC. For the text modality, we obtain  classification logits via \eref{zero-shot} but without normalization:
\begin{equation}
  \tilde{\mathbf{y}}_{l} = ({l}_1, {l}_2, \cdots {l}_k), \quad {l}_i = \cos(\mu_{i}, \mathrm{f}_{lac})/\tau.
  \label{lac_hat}
\end{equation}
To mitigate imbalances in text modality predictions~\cite{padclip, debiaspl}, we  implement Approximated Controlled Direct Effect~\cite{debiaspl} to adjust similarity scores  in \eref{lac_hat}:
\begin{align}
  \mathbf{y}_{l} = \tilde{\mathbf{y}}_{l} - \eta  \log \hat{p},\;
  \hat{p} \leftarrow m\hat{p} + (1-m)\frac{1}{B}\sum_{i=1}^{B}p_i,
  \label{debias1}
\end{align}
where $m, \eta$ are debias factors, $B$ is the batch size, and $p_i=\mathrm{softmax}(\tilde{\mathbf{y}}_{l})$ denotes classification probability of LAC.
For the vision modality, we follow the design in \cite{shot} to introduce a vision classifier with linear layers $\Phi=\{\Phi_{1},\Phi_{2}\}$, 
which first produces the bottleneck features $\mathrm{f}_b = \Phi_1 (\mathrm{f}_{vac})$ and then generates vision outputs: $\mathbf{y}_{v} = \Phi_2 (\mathrm{f}_b)$. For simplicity, below we refer to the vision (text) output of input $x$ as $\mathbf{y}_v|x$ ($\mathbf{y}_l|x$), and omit the $x$ when there is no ambiguity.

\subsection{Modality Discrepancy Metric}
In this section we introduce MDI metric to evaluate instance-level modality characteristics and their potential contributions in the ensemble process. 
We start by analyzing original output logits of vision and text branch, which reflect the prediction confidence and class-wise relations that dominate the ensemble processes.
We first define function $\mathrm{top}_k(\mathbf{y})$ that refers to the \textbf{index} of the $k_{\mathrm{th}}$ largest logit of vector $\mathbf{y}$, and function $\mathrm{val}_k(\mathbf{y})=\mathbf{y}[{\mathrm{top}_k(\mathbf{y})}]$ that outputs the \textbf{value} of the  $k_{\mathrm{th}}$ largest logit.
MDI first broadly categorizes target data into modality-invariant (MI), modality-specific (MS) and uncertain (UN) samples. Subsequently, finer numerical metrics are tailored to each MDI category to quantify their modality characteristics. \fref{fig3}(b) illustrates the concept of MDI metric, where the bars represent classification logits of different modalities.

1) \textbf{Modality-invariant samples.} We define MI samples as those with minimum modality gap and high prediction confidence. Such samples are either simple and straightforward to describe, or have been well-learned  by the current model. With minimum modality gap, prediction results from different modalities should be consistent on MI samples. Based on the notations in Sec. \ref{sec_2}, we define MI samples as:
\begin{align}
  \mathcal{T}_\mathrm{MI} = \{ x^t | \mathrm{top}_1(\mathbf{y}_v)= \mathrm{top}_1(\mathbf{y}_l) \}.
\end{align}
However, we observe that such categorization neglects prediction confidence of different modalities. This oversight can lead to unreliable MI samples that lead both modalities to produce overly similar results. We term such phenomenon as \textit{modality collapse}, which jeopardizes modality ensemble effects.
\textbf{MI-score} is then proposed to quantify both the consistency and confidence of paired modality outputs, with larger scores representing higher confidence:
\begin{align}
  \mathrm{MI}(\mathbf{y}_{v}, \mathbf{y}_{l}) = \mathrm{val}_1(\mathbf{y}_{v}) \mathrm{val}_1(\mathbf{y}_{l}) - \mathrm{val}_2(\mathbf{y}_{v}) \mathrm{val}_2(\mathbf{y}_{l}),
\end{align}
where  $\mathbf{y}_{v}$ and $\mathbf{y}_{l}$ are paired modality outputs of MI samples.
Existing methods that rely on threshold filtering~\cite{daprompt,du2024domain}  to select confident samples could be unreliable due to modality gap. MI-score comprehensively considers prediction consistency confidence across modalities, serving as a more effective metric for assessing sample reliability of VLMs. We select the top-$m\%$ most confident MI samples by:
\begin{align}
  \mathcal{T}_C=\{ x^t |x^t \in \mathcal{T}_\mathrm{MI} \; \mathrm{and} \; 
  \mathrm{MI}(\mathbf{y}_{v}, \mathbf{y}_{l}) > \mathrm{th}_c \},
\label{t_c}
\end{align}
where $\mathrm{th}_c$ is the $m\%$ largest MI-score. Predictions on $\mathcal{T}_C$ are highly accurate. During training, we promote model robustness and better  representation structures by ensuring prediction consistency on $\mathcal{T}_C$, which is elaborated in Sec.\ref{met}.

2) \textbf{Modality-specific samples.} Samples apart from MI are inconsistent across modalities. Specifically, samples with symmetrical modality predictions on top-2 most confident classes are denoted as \textit{modality-specific (MS)} :
\begin{align}
  \mathcal{T}_\mathrm{MS} = \{ x^t | \mathrm{top}_1(\mathbf{y}_v)= \mathrm{top}_2(\mathbf{y}_l) \; \mathrm{and} \; \mathrm{top}_2(\mathbf{y}_v)= \mathrm{top}_1(\mathbf{y}_l) & \}.
\end{align}
The inherent modality-specific information in MS samples makes them better interpreted under a certain modality. We observe high accuracy by selecting the correct modality classifier for MS samples. Therefore, our goal is to find the optimal test-time ensemble weight that emphasizes the `correct' modality, while minimizing the effects of the `wrong' modality. However, as the training progresses, the MS samples evolve rapidly, making a fixed weight ineffective. Manually adjusting the weights for each task is both labor-intensive and impractical. To address this, we propose a modality-adaptive ensemble (MaE) algorithm to automatically determine ensemble weight by analyzing the modality distributions, as detailed in Sec.\ref{mae}

3) \textbf{Uncertain samples.} 
Target samples not categorized as MI or MS samples are Uncertain (UN): $\mathcal{T}_\mathrm{UN}=\{x^t|x^t \notin \mathcal{T}_\mathrm{MI} \; \mathrm{and} \; x^t \notin \mathcal{T}_\mathrm{MS} \}$. UN samples are either underexplored by current model, or lacks modality-shared information, making their predictions highly unreliable. Without labeling on target data, learning UN samples is harmful to the modality-aware training process. Therefore, we opt to actively annotating these samples to enhance representation capabilities in both modalities. 

UN samples with completely different outcomes from the two modalities are further defined as UN-annotate (UN-a), while the rest 
of the UN samples are UN-ensemble (UN-e) samples.
We further design a UN-score to quantify their informativeness. Lower  scores stand for higher annotation priorities:
\begin{align}
  \mathrm{UN}(\mathbf{y}_{v}) = \mathrm{val}_1(\mathbf{y}_{v}) - \mathrm{val}_2(\mathbf{y}_{v}).
\end{align}

We propose to select UN-a samples with the lowest UN-scores for active annotation:
\begin{align}
  \mathcal{T}_{L}=\{ (x^t, y^t) \,| \, x^t \in \mathcal{T}_\mathrm{UN-a} \; \mathrm{and} \; \mathrm{UN}(\mathbf{y}_{v}) < \mathrm{th}_l \},
  \label{active}
\end{align}
where $th_l$ is the UN-score threshold controlling active sample amount.  UN-e samples are also annotated in case UN-a samples run out.

\begin{figure}[!t]
  \centering
  \includegraphics[width=0.3\textwidth]{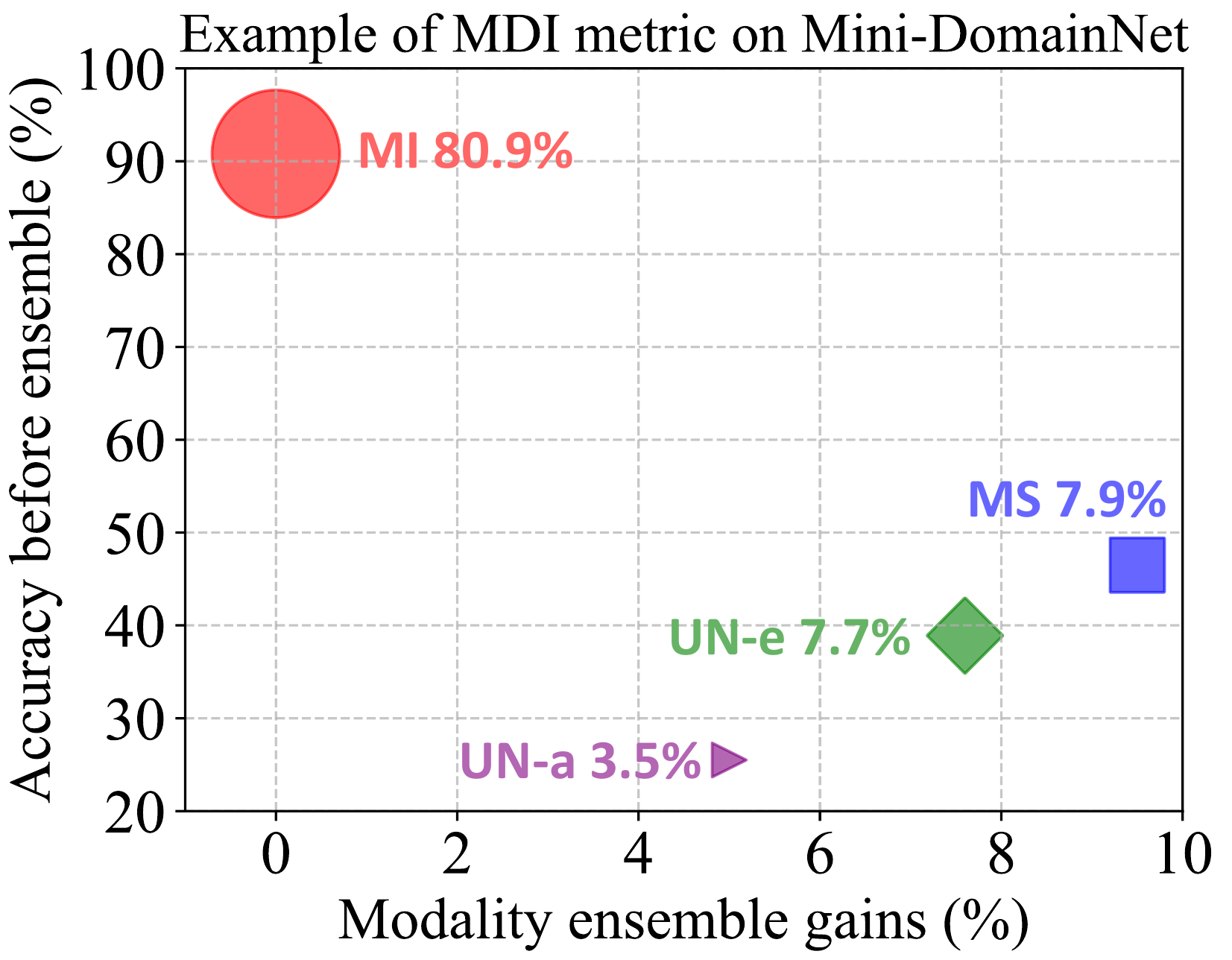}
  \caption{Example of MDI categorization results on clp$\rightarrow$rel of Mini-DomainNet. X-axis shows the accuracy improvement after test-time modality ensemble. Y-axis shows the original vision modality accuracy. Marker sizes are correlated with the proportion (exact number given by texts) of the MDI type they represent.}
  \label{fig4}
  \vspace{-8pt}
\end{figure}

\fref{fig4} provides a real-world example to understand the characteristics of  MDI categories introduced above. The majority of samples are easy MI samples with high prediction accuracy. Note that proportion of MI samples may decrease  on harder tasks. The most significant ensemble effects (+7.9\%) are observed on MS samples, which complies with our theory on their modality-specific information. UN-a samples present the lowest classification accuracies (25\%), making their annotation and exploitation desirable. Accuracy and ensemble effects on UN-e samples are between UN-a and MS samples. To conclude, the performance of MDI metric on real-world dataset is consistent with our design purposes, supporting its efficacy and validness.

\subsection{Modality-aware Training and Adaptation}
\label{met}
We perform modality-aware training and knowledge transfer based on disentangled VAC and LAC.

1) \textbf{Training.} 
Pretrained VLMs preserve rich semantic knowledge that are helpful to identify hard samples, which is distilled to optimize LAC. For \textbf{unlabeled} target data, we obtain zero-shot similarity scores as teacher knowledge:
\begin{align}
  y_u = (l_1-\overline{l} , l_2-\overline{l}, \cdots, l_k-\overline{l}), \quad l_i = \cos(\mu_{i}, \mathrm{f}_{v})/\tau,
\label{y_zs}
\end{align}
where $\overline{l}= \frac{1}{K} \sum_{k=1}^{K} l_k$ normalizes similarity scores and $\mu_{i}$ are extracted from manual prompts. For annotated target samples in $\mathcal{T}_L$ (\eref{active}), we handcraft \textit{pseudo logits} to emphasize the ground-truth $y^t$: 
\begin{align}
  y_l = (2 e_{y^t} - \mathbf{1})h,
\label{y_pslogi}
\end{align}
where $e_y$ is a base vector of length $K$ with only the $y_{\text{th}}$ component being 1, $\mathbf{1}$ is all-one vector, and $h$ is a positive constant. We distill pretrained knowledge in  $y_u$ and exploit annotation information in $y_l$ by ensuring semantic consistency among different strongly  augmented versions:
\begin{align}
  \mathcal{L}_{lac}^t = \underset{x,y \sim \mathcal{T}_L}{\mathbb{E}} \mathrm{KL}(\mathbf{y}_l|\mathcal{A}(x), y_l) + \underset{x \sim \mathcal{T}_U}{\mathbb{E}} \mathrm{KL}(\mathbf{y}_l|x, y_u),
\label{l_lac_t}
\end{align}
where $\mathcal{T}_L$ is the target annotation set, $\mathcal{T}_U = \mathcal{D}^t \setminus  \mathcal{T}_L$, $\mathrm{KL}(,)$ is Kullback-Leibler divergence and $\mathcal{A}(\cdot)$ is augmentation operation. In UDA tasks, $\mathcal{T}_L$ is replaced by $\mathcal{T}_C$. For labeled source data, we optimize cross-entropy loss:
\begin{align}
  \mathcal{L}_{lac}^s = \underset{x,y \sim \mathcal{D}^s}{\mathbb{E}} - \sum_{k=1}^{K} \mathbbm{1}[y=k] \cdot \log P_k(\mathbf{y}_l),
\label{l_lac_s}
\end{align}
where $P_k(y)$ performs $\mathrm{softmax}$ operation on $y$ then takes the $k_{th}$ component. Combining \eref{l_lac_t} and \eref{l_lac_s}, we define the overall training loss for LAC as:
\begin{align}
  \mathcal{L}_{lac} = \mathcal{L}_{lac}^t + \alpha \mathcal{L}_{lac}^s,
\label{l_lac}
\end{align}
where $\alpha$ adjusts the effects of source data supervision.

For the optimization of VAC, we aim to utilize the locality structure of vision representations via a K-means-based deep clustering approach \cite{deepcluster,shot}.  We calculate the clustering centroids for class $k$ as follows:
\begin{align}
  \phi_k = \frac{\sum_{x^t} P_k (y_{ens}) \cdot \mathrm{f}_b}{\sum_{x^t} P_k (y_{ens})},
\end{align}
where $y_{ens}$ introduces test-time ensemble output in \eref{ensemble1} for complementary text-specific knowledge.  For any given target bottleneck feature $\mathrm{f}_b$, we compute its cosine similarity with all centroids, assigning the class with the highest similarity as the  pseudo-label:
\begin{equation}
  y_{v} = \arg \underset{k}{\max} \, \cos(\mathrm{f}_b, \phi_k).
\label{cluster}
\end{equation}
The similarity scores can also represent class-wise relations and replace the logits in \eref{y_zs} in late training stages.
VAC is then optimized separately in a unified manner with LAC. 

Similar to \eref{ensemble1}, train-time ensemble output is defined:
\begin{align}
  \mathbf{y}_e = w \cdot \mathbf{y}_v + (1-w) \cdot \mathbf{y}_l.
\label{ensemble2}
\end{align}
Note the train-time weights $w$ here differ from $w^*$ in \eref{ensemble1}. Train-time weights are learnable to assist the modality-aware training: $w=\mathcal{W}(\mathrm{f}_{vac})$, where $\mathcal{W}$ is the weight generator in \fref{fig3}(a). We separate VAC by \textit{detaching}  $\mathbf{y}_l$ during the optimization of \eref{ensemble2}. The goal is to utilize the text-specific information in LAC while not disturbing the vision branch. The training loss of target VAC are then defined as:
\begin{align}
  \mathcal{L}_{vac}^t = \underset{x,y \sim \mathcal{T}_L}{\mathbb{E}}& - \sum_{k=1}^{K} \mathbbm{1}[y=k] \cdot \log P_k(\mathbf{y}_e|\mathcal{A}(x)) + \nonumber \\
  \underset{x \sim \mathcal{T}_U}{\mathbb{E}}& - \sum_{k=1}^{K} \mathbbm{1}[y_v=k] \cdot \log P_k(\mathbf{y}_e|x),
\label{l_vac_t}
\end{align}
We present gradient analysis on the optimization of \eref{l_vac_t} to show the necessity and efficacy of the train-time ensemble in \eref{ensemble2}. For simplicity, we only analyze the second term:
\begin{align}
  \frac{\partial \mathcal{L}_{vac}^t}{\partial \theta_v} 
  = \frac{\partial \mathcal{L}_{vac}^t}{\partial \mathbf{y}_e} 
  \cdot \frac{\partial \mathbf{y}_e}{\partial \mathbf{y}_v} 
  \cdot \frac{\partial \mathbf{y}_v}{\partial \theta_v} = [P_{y_v}(\mathbf{y}_e)-1] \cdot w \cdot \frac{\partial \mathbf{y}_v}{\partial \theta_v}.\label{gra_v}
\end{align}
\eref{gra_v} presents the gradient for learning VAC, where $\theta_v$ are trainable parameters in the vision branch. Compared with directly optimizing $\mathbf{y}_v$, the advantages include: (1) On language-specific samples, text-correlated knowledge introduced  by the first term allows smooth optimization without forcing destructive cross-modal alignment on VAC, thus protecting vision-specific knowledge. (2) The second term $w$ dynamically controls the importance of vision and text outputs. When encountering unconfident pseudo labels, smaller weights can alleviate the negative effects. 
\begin{align}
  \label{gra_w}
  \frac{\partial \mathcal{L}_{vac}^t}{\partial \theta_w} &= \frac{\partial \mathcal{L}_{vac}^t}{\partial \mathbf{y}_e} \cdot \frac{\partial \mathbf{y}_e}{\partial w} \cdot \frac{\partial w}{\partial \theta_w}  \\
  &= [P_{y_v}(\mathbf{y}_e)-1] \cdot [\mathbf{y}_v - \mathbf{y}_l] \cdot \frac{\partial w}{\partial \theta_w}.\nonumber
\end{align}
\eref{gra_w} presents the gradient for learning $w$, where $\theta_w$ are parameters in $\mathcal{W}$. The first term is ensemble error, which helps learn  $w$ that best combines knowledge from both modalities. The second term measures difference between vision and text outputs. Large inconsistencies between the two modalities generally indicate uncertain or modality-specific samples, which leads to larger gradient for adjusting the weights towards the proper modality. On labeled source data we directly optimize cross-entropy loss for VAC:
\begin{align}
  \mathcal{L}_{vac}^s = \underset{x,y \sim \mathcal{D}^s}{\mathbb{E}}& - \sum_{k=1}^{K} \mathbbm{1}[y=k] \cdot \log P_k(\mathbf{y}_v).
\label{l_vac_s}
\end{align}

To enhance individual discriminability and global diversity,  we follow state-of-the-art~\cite{shot,ahmed2021unsupervised} to apply an information maximization loss $\mathcal{L}_{im}$ for regularization:
\begin{align}
  \mathcal{L}_{im} = - \underset{x \sim \mathcal{D}^t}{\mathbb{E}}  \sum_{k=1}^{K} P_k (\mathbf{y}_e) \log P_k (\mathbf{y}_e)
  +
  \sum_{k=1}^{K} \overline{q}_k \log \overline{q}_k
  ,
\label{l_im}
\end{align}
where $\overline{q}_k=-\mathbb{E}_{x \sim \mathcal{D}^t} P_k(\mathbf{y}_e)$. 
Combining \eref{l_vac_t}, \eref{l_vac_s} and \eref{l_im}, the overall training loss for VAC is given :
\begin{align}
  \mathcal{L}_{vac} = \mathcal{L}_{vac}^t + \beta \mathcal{L}_{vac}^s + \mathcal{L}_{im},
\label{l_vac}
\end{align}
where $\beta$ adjusts the effects of source data supervision.

2) \textbf{Adaptation.}
To achieve domain adaptation on separated modalities, we introduce a modality discriminator $f_d$ to align the distribution of source and target VAC (LAC). The discriminator is first trained on source domain to differentiate LAC from VAC, where it learns distributions of different modalities on the source domain. The discriminator then freezes and applies to differentiate target VAC and LAC. By optimizing the discrimination loss, the modality separator learns to generate target VAC and LAC aligned with source ones, to which the source knowledge are transferred. We adopt binary cross-entropy loss for modality discrimination:
\begin{align}
  \mathcal{L}_{bce}(\mathbf{y}_d;\theta) = -[y_{dis} \log \mathbf{y}_d + (1-y_{dis}) \log (1 - \mathbf{y}_d)],
\end{align}
where $y_{dis}$ are binary modality labels and $\mathbf{y}_d = f_d(x,\mathbf{t})$ is discriminator output. Denoting parameters for the modality separator, discriminator and learnable prompts as $\theta_{Q}, \theta_d, \theta_t$, we define the discrimination loss as:
\begin{align}
  \mathcal{L}_{d} = \underset{x \sim \mathcal{D}^s}{\mathbb{E}} \mathcal{L}_{bce}(\mathbf{y}_d ; \theta_{Q},\theta_d) 
  +
  \underset{x \sim \mathcal{D}^t}{\mathbb{E}} \mathcal{L}_{bce}(\mathbf{y}_d ; \theta_{Q}, \theta_t). 
\label{l_dis}
\end{align}

\begin{figure}[!t]
  \centering
  \includegraphics[width=0.48\textwidth]{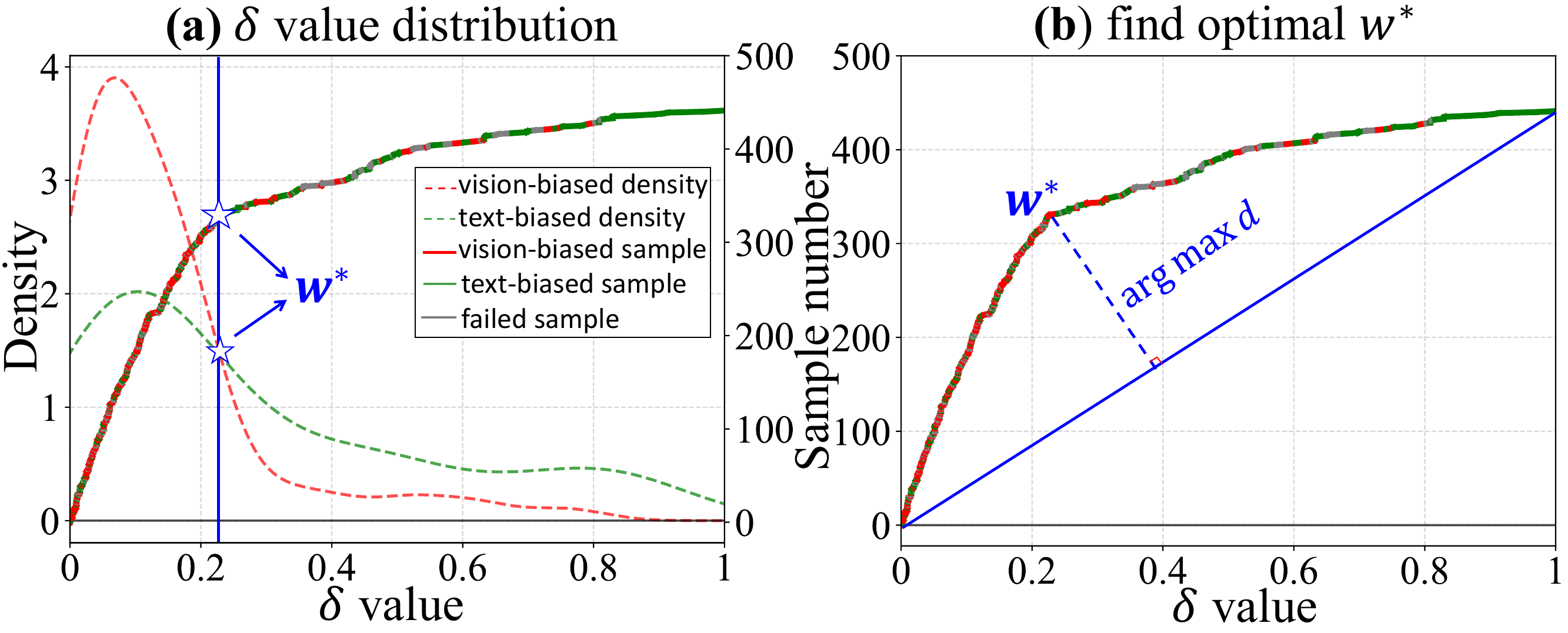}
  \vspace{-8pt}
  \caption{Illustration of MaE algorithm. (a) Modality-biased distribution of sorted $\delta$ on task A-P from OfficeHome. (b) Finding  optimal $w^*$.}
  \label{fig5}
  \vspace{-8pt}
\end{figure}

\subsection{Modality-adaptive Ensemble}
\label{mae}
UniMoS++ introduces train-time ensemble in \eref{ensemble2} with \textit{learnable} weights $w$ to protect modality-specific information during optimization. On the contrary, \textit{predefined} test-time ensemble weights $w^* \in (0,1)$ in \eref{ensemble1} combine modality information for higher accuracy. However, it is challenging to set a suitable $w^*$ for all adaptation tasks and different training stages. An inappropriate $w^*$ leads to low-quality pseudo labels in \eref{cluster} and negative ensemble effects. Therefore, we propose modality-adaptive ensemble (MaE) algorithm to automatically determine $w^*$. We first make a reasonable assumption that the ensemble result is consistent with either the vision or text output. Denote vision classification result as $i=\mathrm{top}_1(\mathbf{y}_{v})$, text classification results as $j=\mathrm{top}_1(\mathbf{y}_{l})$, their logits as $v_i=\mathbf{y}_{v}[i],v_j=\mathbf{y}_{v}[j]$ and $l_i=\mathbf{y}_{l}[i],l_j=\mathbf{y}_{l}[j]$. If the ensemble outputs are consistent with vision ones, $w^*$ should satisfy the following inequality:
\begin{align}
  w^* v_i + (1-w^*)l_i &> w^* v_j + (1-w^*) l_j \nonumber \\
  w^*(v_i-l_i) + l_i &> w^*(v_j - l_j) + l_j.
\end{align}
Denote $\Delta_i=v_i-l_i$, $\Delta_j=v_j-l_j$, $\Delta_l=l_j-l_i$. In practice, $\Delta_i-\Delta_j>0$ always holds, therefore we have:
\begin{align}
  w^* > \frac{\Delta_l}{\Delta_i-\Delta_j}.
\label{w}
\end{align}

The ensemble threshold $\delta=\Delta_l/(\Delta_i-\Delta_j)$ decides which modality dominates the ensemble output: $w^*>\delta$ for vision output and $w^*<\delta$ for text output. Samples with correct vision (text) outputs are termed  \textit{vision(text)-biased}. Our goal is to find a $w^*$ greater than most of $\delta$ values in vision-biased samples, and smaller than most of $\delta$ values in text-biased samples. The solid curve in \fref{fig5}(a) presents sorted $\delta$ values for all target data, with red parts representing vision-biased samples and green parts for text-biased ones. The dotted curves show distribution density of vision and text-biased samples over different $\delta$ values. The intersection of two density curves represent the optimal $w^*$.

If the ensemble thresholds for all target data distribute evenly, the sorted $\delta$ value should form a straight line. But due to inherent modality-specific characteristics and different optimization processes in vision and text branch, we can observe that in vision-biased region ($\delta<w^*$) sample number increases sharply, while in text-biased region ($\delta>w^*$) sample number increases mildly, forming a convex curve. We propose to find the most deviated point in the sorted $\delta$ line to approximate the optimal $w^*$, as shown in \fref{fig5}(b). For target samples sorted by $\delta$ value, we can denote the point coordinate with the $k_\mathrm{th}$ largest $\delta$ value as $(\delta_k,N_k)$. Define function $d(x,y,a)$ that calculates the distance from point $(x,y)$ to line $ax-y=0$, we have optimal $w^*$ by:
\begin{align}
  w^*=\delta_{k^*}, \; k^*=\underset{k}{\arg \max } \, d(\delta_k,N_k,N^t).
\label{w_star}
\end{align}

\begin{algorithm}[t]
    \caption{Training algorithm of UniMoS++ for ADA task}
    \label{alg1}
    \textbf{Input}: Labeled source data $\mathcal{D}^s$, unlabeled target data $\mathcal{D}^t$, maximum epoch number $max\_epoch$, pretrained CLIP, annotation budget $b$, annotation count in each round $b_r$.\\
    \textbf{Output}: Trained modality separator $Q_{t},Q_{v}$, trained prompts $\mathbf{t}$ and linear layers $\Phi$.
    \begin{algorithmic}[1] 
    \STATE $epoch \leftarrow 0$, $c \leftarrow 0$, $\mathcal{T}_L \leftarrow \emptyset $.
    \WHILE{$epoch < max\_epoch$}
    \STATE Obtain text teacher knowledge $y_u$ and $y_l$ via \eref{y_zs} and \eref{y_pslogi}.
    \STATE Obtain target vision pseudo label $y_{v}^t$ via \eref{cluster}.
    \STATE Obtain CLIP's vision features $\mathrm{f}_v$ and text features $\mu_i$, and disentangled $\mathrm{f}_{vac}=Q_{v}(\mathrm{f}_v)$, $\mathrm{f}_{lac}=Q_{t}(\mathrm{f}_v)$.
    \STATE Obtain train-time ensemble outputs  by \eref{ensemble2}.
    \STATE Calculate modality discrimination losses by \eref{l_dis}.
    \STATE Update network parameters by optimizing \eref{l_all}.
    \IF {$c < b$}
    \STATE  Select  active target samples by \eref{active} to form current annotation set $\mathcal{T}_r$
    \STATE  Update $\mathcal{T}_L \leftarrow \mathcal{T}_L + \mathcal{T}_r$, $c \leftarrow c+b_r$
    \ENDIF
    \STATE Obtain inference results by \eref{w_star} and \eref{ensemble1}. 
    \STATE $epoch \leftarrow epoch+1$
    \ENDWHILE
    \end{algorithmic}
  \end{algorithm}

\subsection{Overall Algorithm}
The overall training and inference process of UniMoS++ on ADA task are in Algorithm \ref{alg1}. 
For UDA tasks, simply set $b=0$ and replace $\mathcal{T}_L$ in \eref{l_lac_t}, \eref{l_vac_t} with $\mathcal{T}_C$. 
Combining \eref{l_ortho}, \eref{l_lac}, \eref{l_vac}, \eref{l_dis}, we have the training loss:
\begin{align}
  \mathcal{L} = \mathcal{L}_{lac} + \mathcal{L}_{vac} + \gamma \mathcal{L}_{ortho} + \gamma \mathcal{L}_{d},
  \label{l_all}
\end{align}
where $\gamma$ is regularization hyperparameter.  Note that CLIP-pretrained encoders remain frozen the whole time, and that all trainable modules in UniMoS++ are linear layers.

\begin{table*}[t]
    \caption{UDA results on OfficeHome with ResNet50 (upper part) and ViT-B (lower part). Best results are marked bold. Methods with `*' are based on CLIP.}
    \centering
    \label{officehome}
    \vspace{-5pt}
    \resizebox{0.9\linewidth}{!}{
    \setlength{\tabcolsep}{4.5pt}
    \begin{tabular}{p{2.1cm}|p{1.3cm}|cccccccccccc|c}
    \toprule
    Method & Venue  & A$\to$C & A$\to$P & A$\to$R & C$\to$A & C$\to$P & C$\to$R & P$\to$A & P$\to$C & P$\to$R & R$\to$A & R$\to$C & R$\to$P & Avg. \\ \midrule
    CLIP*~\cite{clip} & ICML'21  & 51.7 & 81.5 & 82.3 & 71.7 & 81.5 & 82.3 & 71.7 & 51.7 & 82.3 & 71.7 & 51.7 & 81.5 & 71.8 \\
    MSGD~\cite{msgd} & TPAMI'22 & 58.7 & 76.9 & 78.9 & 70.1 & 76.2 & 76.6 & 69.0 & 57.2 & 82.3 & 74.9 & 62.7 & 84.5 & 72.3 \\
    KUDA~\cite{kuda} & ECCV'22 & 58.2 & 80.0 & 82.9 & 71.1 & 80.3 & 80.7 & 71.3 & 56.8 & 83.2 & 75.5 & 60.3 & 86.6 & 73.9 \\
    EIDCo~\cite{eidco} & ICCV'23 & \textbf{63.8} & 80.8 & 82.6 & 71.5 & 80.1 & 80.9 & 72.1 & 61.3 & 84.5 & \textbf{78.6} & 65.8 & 87.1 & 75.8 \\
    ICON~\cite{yue2023make} & NeurIPS'23 & 63.3 & 81.3 & 84.5 & 70.3 & 82.1 & 81.0 & 70.3 & \textbf{61.8} & 83.7 & 75.6 & \textbf{68.6} & 87.3 & 75.8 \\ 
    DAPrompt*~\cite{daprompt} & TNNLS'23 & 54.1 & 84.3 & 84.8 & 74.4 & 83.7 & 85.0 & 74.5 & 54.6 & 84.8 & 75.2 & 54.7 & 83.8 & 74.5 \\
    PDA*~\cite{bai2024prompt} & AAAI'24 & 55.4 & 85.1 & 85.8 & 75.2 & 85.2 & 85.2 & 74.2 & 55.2 & 85.8 & 74.7 & 55.8 & 86.3 & 75.3 \\
    \rowcolor[RGB]{221,235,247} 
    UniMoS*(0.3) & CVPR'24 & 59.5 & 89.4 & 86.9 & 75.2 & 89.6 & 86.8 & 75.4 & 58.4 & 87.2 & 76.9 & 59.5 & \textbf{89.7} & 77.9 \\ 
    \rowcolor[RGB]{221,235,247} 
    UniMoS*(0.5) & CVPR'24 & 58.1 & 88.1 & 86.3 & 73.7 & 88.9 & 85.9 & 73.8 & 57.1 & 86.3 & 75.8 & 59.1 & 88.9 & 76.8 \\
    \rowcolor[RGB]{221,235,247} 
    UniMoS*(0.7) & CVPR'24 & 56.7 & 86.8 & 84.8 & 73.4 & 88.7 & 85.4 & 73.1 & 57.1 & 85.8 & 74.9 & 58.3 & 88.6 & 76.1 \\
    \rowcolor[RGB]{221,235,247} 
    \textbf{UniMoS++*} & TPAMI'25 & 59.6 & \textbf{89.6} & \textbf{87.0} & 75.0 & \textbf{89.8} & \textbf{86.9} & 75.5 & 59.0 & \textbf{87.3} & 77.2 & 59.4 & 89.6  & \textbf{78.0}  \\ 
    \rowcolor[RGB]{217,217,217} 
    PADCLIP(FT)*~\cite{padclip} & ICCV'23 & 57.5 & 84.0 & 83.8 & \textbf{77.8} & 85.5 & 84.7 & 76.3 & 59.2 & 85.4 & 78.1 & 60.2 & 86.7 & 76.6 \\ \midrule

    CLIP*~\cite{clip} & ICML'21 & 67.8 & 89.0 & 89.8 & 82.9 & 89.0 & 89.8 & 82.9 & 67.8 & 89.8 & 82.9 & 67.8 & 89.0 & 82.4 \\
    TVT~\cite{tvt} & WACV'23 & 74.9 & 86.8 & 89.5 & 82.8 & 88.0 & 88.3 & 79.8 & 71.9 & 90.1 & 85.5 & 74.6 & 90.6 & 83.6 \\
    DAPrompt*~\cite{daprompt} & TNNLS'23 & 70.6 & 90.2 & 91.0 & 84.9 & 89.2 & 90.9 & 84.8 & 70.5 & 90.6 & 84.8 & 70.1 & 90.8 & 84.0 \\
    PDA*~\cite{bai2024prompt} & AAAI'24 & 73.5 & 91.4 & 91.3 & 86.0 & 91.6 & 91.5 & 86.0 & 73.5 & 91.7 & 86.4 & 73.0 & 92.4 & 85.7 \\
    \rowcolor[RGB]{221,235,247} 
    \textbf{UniMoS*(0.3)} & CVPR'24 & 74.9 & 94.0 & \textbf{92.5} & 86.4 & \textbf{94.3} & \textbf{92.5} & 86.0 & 73.9 & \textbf{93.0} & 86.4 & 74.2 & 94.5 & \textbf{86.9} \\
    \rowcolor[RGB]{221,235,247} 
    UniMoS*(0.5) & CVPR'24 & 73.5 & 93.8 & 91.8 & 85.6 & 93.9 & 91.9 & 84.8 & 72.9 & 92.7 & 85.6 & 73.8 & \textbf{94.6} & 86.2 \\ 
    \rowcolor[RGB]{221,235,247} 
    UniMoS*(0.7) & CVPR'24 & 72.6 & 93.5 & 91.3 & 85.2 & 93.7 & 91.8 & 84.5 & 72.9 & 92.7 & 85.3 & 73.7 & 94.4 & 86.0 \\
    \rowcolor[RGB]{221,235,247} 
    UniMoS++* & TPAMI'25 & 74.6 & \textbf{94.4} & 92.2 & \textbf{86.8} & 94.0 & 92.3 & 85.8 & 73.6 & 92.8 & 86.2 & 74.5 & 94.5 & 86.8 \\ 
    \rowcolor[RGB]{217,217,217} 
    PADCLIP(FT)*~\cite{padclip} & ICCV'23 & \textbf{76.4} & 90.6 & 90.8 & 86.7 & 92.3 & 92.0 & 86.0 & \textbf{74.5} & 91.5 & \textbf{86.9} & \textbf{79.1} & 93.1 & 86.7 \\ \bottomrule

    \end{tabular}}
    \vspace{-8pt}
\end{table*}
\begin{table*}[ht]
    \caption{Class-wise UDA results on VisDA-2017 with ResNet101. Best results are marked bold. Methods with `*' are based on CLIP.}
    \vspace{-5pt}
    \label{visda}
    \footnotesize
    \centering
    \resizebox{0.9\linewidth}{!}{
    \setlength{\tabcolsep}{5pt}
    \begin{tabular}{p{2.1cm}|p{1.3cm}|cccccccccccc|c}
    \toprule
    Method & Venue & plane & bicycle & bus & car & horse & knife & mcycl & person & plant & sktbrd & train & truck & Avg. \\ \midrule
    CLIP*~\cite{clip} & ICML'21 & 98.2 & 83.9 & 90.5 & 73.5 & 97.2 & 84.0 & 95.3 & 65.7 & 79.4 & 89.9 & 91.8 & 63.3 & 84.4 \\
    MSGD~\cite{msgd} & TPAMI'22 & 97.5 & 83.4 & 84.4 & 69.4 & 95.9 & 94.1 & 90.9 & 75.5 & 95.5 & 94.6 & 88.1 & 44.9 & 84.5 \\
    DAPrompt*~\cite{daprompt} & TNNLS'23 & 97.8 & 83.1 & 88.8 & 77.9 & 97.4 & 91.5 & 94.2 & 79.7 & 88.6 & 89.3 & \textbf{92.5} & 62.0 & 86.9 \\
    PDA*~\cite{bai2024prompt} & AAAI'24 & 97.2 & 82.3 & 89.4 & 76.0 & 97.4 & 87.5 & 95.8 & 79.6 & 87.2 & 89.0 & 93.3 & 62.1 & 86.4 \\
    \rowcolor[RGB]{221,235,247} 
    UniMoS*(0.3) & CVPR'24 & \textbf{98.5} & 88.1 & 90.1 & 74.0 & 96.8 & 95.0 & 92.7 & 84.5 & 89.8 & 87.8 & \textbf{92.5} & 65.1 & 87.9 \\ 
    \rowcolor[RGB]{221,235,247} 
    UniMoS*(TLR) & CVPR'24 & 97.7 & 88.2 & 90.1 & 74.6 & 96.8 & 95.8 & 92.4 & 84.1 & 90.8 & 89.0 & 91.8 & 65.3 & 88.1 \\ 
    \rowcolor[RGB]{221,235,247} 
    \textbf{UniMoS++*} & TPAMI'25 & 98.1 & 88.6 & 90.0 & 75.4 & 96.9 & 95.6 & 92.8 & 83.5 & 89.6 & 88.7 & 92.4 & \textbf{65.5} & 88.1 \\ 
    \rowcolor[RGB]{217,217,217} 
    PADCLIP(FT)*~\cite{padclip} & ICCV'23 & 96.7 & \textbf{88.8} & 87.0 & 82.8 & 97.1 & 93.0 & 91.3 & 83.0 & 95.5 & 91.8 & 91.5 & 63.0 & \textbf{88.5} \\ \bottomrule
    \end{tabular}}
    \vspace{-8pt}
\end{table*}

The  UniMoS++ in Algorithm \ref{alg1} can handle multiple adaptation settings with minimal changes. MDI metric reveals instance-level modality characteristics that contributes to effective active selection for ADA (line 10),  and more reliable pseudo labels for UDA (\eref{t_c}). MaE (line 13) not only boosts ensemble results in UDA and ADA, but also endows UniMoS++ with the ability to tackle multi-source DA (MSDA). For MSDA, we combine all source domains to form a mixed domain, and then learn dynamic adaptation weights to fit the mixed distribution with MaE. Given only pretrained weights in the source-free ADA (SFADA) setting, UniMoS++ can quickly adapt to target domain by tuning prompts on unlabeled target data. The results presented in Sec.\ref{sec_mainres} highlight the abilities of UniMoS++.

\section{Experiments}
\subsection{Dataset and Implementation Details}
UniMoS++ is evaluated on 5  mainstream domain adaptation  datasets. \textbf{OfficeHome}~\cite{venkateswara2017deep} is partitioned into 4 distinct domains (\textbf{A}rt, \textbf{C}lipart, \textbf{P}roduct and \textbf{R}ealworld), each containing 65 categories of office items. On \textbf{VisDA-2017}~\cite{peng2017visda}, the goal is to transfer knowledge from 152k synthetic images  to 55k images of real items. \textbf{DomainNet}~\cite{domainnet} is the most challenging UDA benchmark so far, containing 0.6 million samples from 345 categories divided into 6 distinct domains: \textbf{cl}i\textbf{p}art, \textbf{inf}ograph, \textbf{p}ai\textbf{nt}ing, \textbf{q}uick\textbf{dr}aw, \textbf{re}a\textbf{l}, \textbf{sk}e\textbf{t}ch. We additionally provide results on \textbf{Mini-DomainNet}~\cite{saito2019semi,litrico2023guiding}, a subset of DomainNet with 4 domains and 126 categories. 
\textbf{WILDS}   \cite{koh2021wilds} is a challenging in-the-wild adaptation benchmark composed of 10 sub-tasks, and we select 3 image classification tasks for evaluation following \cite{chi2024adapting}.

All experiments are conducted on one NVIDIA RTX 3090Ti GPU. The CLIP-extracted vision and text features are obtained once and saved in memory to save computation costs. For all tasks, we adopt SGD optimizer with batch size 32, and set $m=0.99, \eta=0.5$ in \eref{debias1}. We set initial learning rate 3e-3 with annealing strategy for learning rate decay. Regularization weight $\gamma$ in \eref{l_all} is 0.01 across all datasets and tasks. For UDA tasks, we train 60 epochs on OfficeHome, DomainNet, Mini-DomainNet and 10 epochs on VisDA. For ADA tasks, we train 60 epochs on all datasets. We annotate $0.5b$ UN-a samples each active round. If amount of UN-a samples is not enough, UN-e samples with lowest UN-scores are selected for complement.  $th_c$ in \eref{t_c} is decided as the lowest MI-score of the top-10\% MI samples. 
The dimension $d_b$ of bottleneck feature is 256. The modality discriminator consists of two linear layers $(d_v,256)$ and (256,1) with a ReLU activation layer. The weight generator  consists of two linear layers $(d_v,256)$ and (256,1) with a Sigmoid activation layer.

\begin{table*}[ht]
    \centering
    \caption{UDA results on DomainNet with vision transformers. Best results are marked in bold font. Methods with `*' are based on CLIP.}
    \vspace{-5pt}
    \label{domainnet}
    \resizebox{0.95\linewidth}{!}{ 
    \setlength{\tabcolsep}{3pt}
    \begin{tabular}{@{}c|ccccccc||c|ccccccc||c|ccccccc}
    \hline
    \begin{tabular}[c]{@{}c@{}}DeiT\\ -B~\cite{deit}\end{tabular} & clp & inf & pnt & qdr & rel & skt & avg & \begin{tabular}[c]{@{}c@{}}ViT\\ -B~\cite{vit}\end{tabular} & clp & inf & pnt & qdr & rel & skt & avg & \begin{tabular}[c]{@{}c@{}}SSRT\\ -B~\cite{ssrt}\end{tabular} & clp & inf & pnt & qdr & rel & skt & avg \\ 
    \hline
    clp & - & 24.3 & 49.6 & 15.8 & 65.3 & 52.1 & 41.4 & clp & - & 27.2 & 53.1 & 13.2 & 71.2 & 53.3 & 43.6 & clp &  & 33.8 & 60.2 & 19.4 & 75.8 & 59.8 & 49.8 \\ 
    inf & 45.9 & - & 45.9 & 6.7 & 61.4 & 39.5 & 39.9 & inf & 51.4 & - & 49.3 & 4.0 & 66.3 & 41.1 & 42.4 & inf & 55.5 & - & 54.0 & 9.0 & 68.2 & 44.7 & 46.3 \\ 
    pnt & 53.2 & 23.8 & - & 6.5 & 66.4 & 44.7 & 38.9 & pnt & 53.1 & 25.6 & - & 4.8 & 70.0 & 41.8 & 39.1 & pnt & 61.7 & 28.5 & - & 8.4 & 71.4 & 55.2 & 45.0 \\ 
    qdr & 31.9 & 6.8 & 15.4 & - & 23.4 & 20.6 & 19.6 & qdr & 30.5 & 4.5 & 16.0 & - & 27.0 & 19.3 & 19.5 & qdr & 42.5 & 8.8 & 24.2 & - & 37.6 & 33.6 & 29.3 \\ 
    rel & 59.0 & 25.8 & 56.3 & 9.2 & - & 44.8 & 39.0 & rel & 58.4 & 29.0 & 60.0 & 6.0 & - & 45.8 & 39.9 & rel & 69.9 & 37.1 & 66.0 & 10.1 & - & 58.9 & 48.4 \\ 
    skt & 60.6 & 20.6 & 48.4 & 16.5 & 61.2 & - & 41.5 & skt & 63.9 & 23.8 & 52.3 & 14.4 & 67.4 & - & 44.4 & skt & 70.6 & 32.8 & 62.2 & 21.7 & 73.2 & - & 52.1 \\ 
    avg & 50.1 & 20.3 & 43.1 & 10.9 & 55.5 & 40.3 & \colorbox{lightgray}{36.7} & avg & 51.5 & 22.0 & 46.1 & 8.5 & 60.4 & 40.3 & \colorbox{lightgray}{38.1} & avg & 60.0 & 28.2 & 53.3 & 13.7 & 65.3 & 50.4 & \colorbox{lightgray}{45.2} \\ \hline
    \begin{tabular}[c]{@{}c@{}}CDTrans\\ -DeiT~\cite{cdtrans}\end{tabular} & clp & inf & pnt & qdr & rel & skt & avg & \begin{tabular}[c]{@{}c@{}}PMTrans\\ -Swin~\cite{pmtrans}\end{tabular} & clp & inf & pnt & qdr & rel & skt & avg & \begin{tabular}[c]{@{}c@{}}DAPrompt\\ -B*~\cite{daprompt}\end{tabular} & clp & inf & pnt & qdr & rel & skt & avg \\ \hline
    clp & - & 29.4 & 57.2 & 26.0 & 72.6 & 58.1 & 48.7 & clp & - & 34.2 & 62.7 & 32.5 & 79.3 & 63.7 & 54.5 & clp & - & 73.0 & 73.8 & 72.6 & 73.9 & 73.5 & 73.4  \\
    inf & 57.0 & - & 54.4 & 12.8 & 69.5 & 48.4 & 48.4 & inf & 67.4 & - & 61.1 & 22.2 & 78.0 & 57.6 & \textbf{57.3} & inf & 50.8 & - & 50.1 & 49.6 & 50.6 & 50.3 & 50.3  \\ 
    pnt & 62.9 & 27.4 & - & 15.8 & 72.1 & 53.9 & 46.4 & pnt & 69.7 & 33.5 & - & 23.9 & 79.8 & 61.2 & 53.6 & pnt & 71.1 & 70.6 & - & 70.0 & 72.7 & 71.7 & 71.2 \\ 
    qdr & 44.6 & 8.9 & 29.0 & - & 42.6 & 28.5 & 30.7 & qdr & 54.6 & 17.4 & 38.9 & - & 49.5 & 41.0 & \textbf{40.3} & qdr & 17.2 & 14.4 & 13.9 & - & 14.3 & 13.9 & 14.7  \\ 
    rel & 66.2 & 31.0 & 61.5 & 16.2 & - & 52.9 & 45.6 & rel & 74.1 & 35.3 & 70.0 & 25.4 & - & 61.1 & 53.2 & rel & 84.9 & 84.8 & 84.9 & 84.7 & - & 84.6 & 84.8  \\ 
    skt & 69.0 & 29.6 & 59.0 & 27.2 & 72.5 & - & 51.5 & skt & 73.8 & 33.0 & 62.6 & 30.9 & 77.5 & - & 55.6 & skt & 65.8 & 65.4 & 65.8 & 64.9 & 65.9 & - & 65.6  \\ 
    avg & 59.9 & 25.3 & 52.2 & 19.6 & 65.9 & 48.4 & \colorbox{lightgray}{45.2} & avg & 67.9 & 30.7 & 59.1 & 27.0 & 72.8 & 56.9 & \colorbox{lightgray}{52.4} & avg & 57.8 & 61.4 & 57.7 & 68.1 & 55.0 & 58.4 & \colorbox{lightgray}{59.8}\\ \hline

    \begin{tabular}[c]{@{}c@{}}DAMP\\ -B*~\cite{du2024domain}\end{tabular} & clp & inf & pnt & qdr & rel & skt & avg & \begin{tabular}[c]{@{}c@{}}UniMoS\\ -B* (TLR)\end{tabular} & clp & inf & pnt & qdr & rel & skt & avg & \begin{tabular}[c]{@{}c@{}}\textbf{UniMoS}\\ \textbf{++-B*} \end{tabular} & clp & inf & pnt & qdr & rel & skt & avg \\ \hline
    clp & - & 75.8 & 76.2 & 75.9 & 76.7 & 77.0 & 76.3 & clp & - & 76.5 & 77.2 & 76.6 & 77.5 & 77.8 & 77.1 & clp & - & 76.7 & 77.2 & 76.5 & 77.7 & 78.0 & \textbf{77.2} \\ 
    inf & 55.7 & - & 55.4 & 55.6 & 56.2 & 55.9 & 55.8 & inf & 55.1 & - & 55.0 & 54.6 & 55.3 & 55.2 & 55.0 & inf & 55.3 & - & 55.0 & 54.8 & 55.5 & 55.3 & 55.2  \\ 
    pnt & 72.2 & 71.9 & - & 72.1 & 73.1 & 73.5 & \textbf{72.6} & pnt & 72.3 & 71.5 & - & 69.4 & 72.5 & 72.6 & 71.7 & pnt & 71.9 & 71.4 & - & 69.2 & 72.4 & 71.9 & 71.4 \\
    qdr & 21.3 & 20.8 & 21.5 & - & 21.6 & 21.2 & 21.3 & qdr & 25.0 & 22.9 & 23.6 & - & 23.7 & 25.1 & 24.1 & qdr & 25.8 & 24.1 & 24.3 & - & 25.2 & 25.4 & 25.0 \\ 
    rel & 85.8 & 85.6 & 84.9 & 85.1 & - & 85.5 & 85.4  & rel & 86.0 & 85.9 & 85.8 & 85.5 & - & 85.9 & \textbf{85.8} & rel & 86.1 & 85.8 & 85.7 & 85.6 & - & 85.8 & \textbf{85.8} \\
    skt & 67.5 & 67.7 & 68.5 & 67.9 & 68.3 & - & 68.0 & skt & 68.5 & 67.8 & 68.2 & 67.5 & 68.0 & - & 68.0 & skt & 68.9 & 67.7 & 68.3 & 67.5 & 68.1 & - & \textbf{68.1} \\
    avg & 60.5 & 64.4 & 61.3 & 71.3 & 59.2 & 62.6 & \colorbox{lightgray}{63.2} & avg & 61.4 & 64.9 & 62.0 & 70.7 & 59.4 & 63.3 & \colorbox[RGB]{221,235,247}{63.6} & avg & 61.6 & 65.1 & 62.1 & 70.7 & 59.8 & 63.3  & \colorbox[RGB]{221,235,247} {\textbf{63.8}} \\ \hline

    \end{tabular}}
    \vspace{-8pt}
\end{table*}

\begin{table*}[!t]
    \caption{MSDA results on DomainNet and OfficeHome. Best results are marked in bold font. Methods with `*' are based on CLIP.}
    \centering
    \label{multisrc}
    \vspace{-5pt}
    \resizebox{0.85\linewidth}{!}{
    \setlength{\tabcolsep}{5.5pt}
    \begin{tabular}{l|l|ccccccc|ccccc}
    \toprule
    & & \multicolumn{7}{c|}{DomainNet (ResNet101)} & \multicolumn{5}{c}{OfficeHome (ResNet50)} \\ \cmidrule(l){3-14} 
    \multirow{-2}{*}{Method} & \multirow{-2}{*}{Venue} & $\rightarrow$clp & $\rightarrow$inf & $\rightarrow$pnt & $\rightarrow$qdr & $\rightarrow$rel & $\rightarrow$skt & Avg. & $\rightarrow$A & $\rightarrow$C & $\rightarrow$P & $\rightarrow$R & Avg. \\ \midrule
    M$^3$SDA~\cite{domainnet} & ICCV'19 & 58.6 & 26.0 & 52.3 & 6.3 & 62.7 & 49.5 & 42.6 & 67.2 & 63.5 & 79.1 & 79.4 & 72.3 \\
    LtC-MSDA~\cite{wang2020learning} & ECCV'20 & 63.1 & 28.7 & 56.1 & 16.3 & 66.1 & 53.8 & 47.4 & 67.4 & \textbf{64.1} & 79.2 & 80.1 & 72.7 \\
    CLIP*~\cite{clip} & ICML'21 & 61.3 & 42.0 & 56.1 & 10.3 & 79.3 & 54.1 & 50.5 & 71.7 & 51.7 & 81.5 & 82.3 & 71.8 \\
    DAPrompt*~\cite{daprompt} & TNNLS'23 & 62.4 & 43.8 & 59.3 & 10.6 & 81.5 & 54.6 & 52.0 & 72.8 & 51.9 & 82.6 & 83.7 & 72.8 \\
    MPA*~\cite{chen2023multi} & NeurIPS'23 & 65.2 & 47.3 & 62.0 & 10.2 & 82.0 & 57.9 & 54.1 & 74.8 & 54.9 & 86.2 & 85.7 & 75.4 \\
    DAMP*~\cite{du2024domain} & CVPR'24 & 69.7 & \textbf{51.0} & \textbf{67.5} & 14.7 & 82.5 & 61.5 & 57.8 & \textbf{77.7} & 61.2 & \textbf{90.1} & 87.7 & \textbf{79.2} \\
    \rowcolor[RGB]{221,235,247} 
    UniMoS*(TLR)~\cite{unimos} & CVPR'24 & 69.8 & 50.1 & 65.5 & 17.2 & 82.5 & 61.7 & 57.8 & 76.8 & 59.1 & 89.7 & 87.6 & 78.3 \\
    \rowcolor[RGB]{221,235,247} 
    \textbf{UniMoS++*} & TPAMI'25 & \textbf{70.6} & 50.7 & 65.8 & \textbf{19.0} & \textbf{82.8} & \textbf{62.5} & \textbf{58.6} & 77.1 & 59.7 & 89.9 & \textbf{88.0} & 78.7 \\ \bottomrule
    \end{tabular}}
    \vspace{-8pt}
\end{table*}

\begin{table*}[t]
    \caption{ADA and SFADA (methods with $\dagger$) results on Office-Home with ResNet50. Best results are marked bold. B is annotation budget.}
    \vspace{-5pt}
    \centering
    \label{active_officehome}
    \resizebox{0.9\linewidth}{!}{
    \setlength{\tabcolsep}{3.5pt}
    \begin{tabular}{p{2.3cm}|p{1.4cm}|c|cccccccccccc|c}
    \toprule
    Method & Venue & B & A$\to$C & A$\to$P & A$\to$R & C$\to$A & C$\to$P & C$\to$R & P$\to$A & P$\to$C & P$\to$R & R$\to$A & R$\to$C & R$\to$P & Avg. \\ \midrule
    CLIP \cite{clip} & ICML'21 & \multirow{13}{*}{5\%} & 51.7 & 81.5 & 82.3 & 71.7 & 81.5 & 82.3 & 71.7 & 51.7 & 82.3 & 71.7 & 51.7 & 81.5 & 71.8 \\ 
    CLUE~\cite{clue} & ICCV'21 &  & 63.0 & 81.7 & 81.1 & 63.2 & 79.3 & 76.2 & 64.6 & 59.7 & 81.5 & 73.1 & 63.5 & 86.8 & 72.8 \\
    EADA~\cite{eada} & AAAI'22 &  & 63.6 & 84.4 & 83.5 & 70.7 & 83.7 & 80.5 & 73.0 & 63.5 & 85.2 & 78.4 & 65.4 & 88.6 & 76.7 \\
    DAPM-TT~\cite{dapm} & NeurIPS'23 &  & 64.2 & 85.4 & 85.7 & 69.2 & 84.2 & 83.5 & 69.1 & 63.4 & 86.0 & 77.2 & 68.4 & 88.6 & 77.1 \\
    MHPL~\cite{wang2023mhpl} & CVPR'23 &  & 66.8 & 82.3 & 84.0 & 71.1 & 84.2 & 80.6 & 71.9 & 65.5 & 84.5 & 78.1 & 67.6 & 89.3 & 77.2 \\
    DiaNA~\cite{diana} & CVPR'23 &  & 64.5 & 86.0 & 84.9 & 72.3 & 84.6 & 82.5 & 73.3 & 63.7 & 85.6 & 78.5 & 67.2 & 89.5 & 77.7 \\
    DUC~\cite{duc} & ICLR'23 &  & 65.5 & 84.9 & 84.3 & 73.0 & 83.4 & 81.1 & 73.9 & 66.6 & 85.4 & 80.1 & \textbf{69.2} & 88.8 & 78.0 \\
    LAS~\cite{lada} & ICCV'23 &  & \textbf{68.6} & 86.7 & 85.0 & 72.1 & 84.6 & 80.6 & 71.4 & 65.6 & 85.5 & 79.4 & 68.4 & 89.9 & 78.2 \\
    \rowcolor[RGB]{221,235,247}  
    \textbf{UniMoS++} & TPAMI'25 &  & 67.3 & 93.1 & 89.6 & 79.3 & 92.8 & 89.1 & \textbf{80.1} & 66.7 & 89.7 & 80.6 & 67.2 & 93.5 & \textbf{82.4} \\ 
    \rowcolor[RGB]{221,235,247}  
    \textbf{\textit{w/} post\_prp.}~\cite{denseclip} & TPAMI'25 &  & 66.5 & 92.8 & 89.5 & 78.9 & 92.9 & 89.1 & 79.5 & \textbf{67.1} & \textbf{90.0} & 80.3 & 68.2 & 93.4 & \textbf{82.4} \\ 
    \rowcolor[RGB]{221,235,247}  
    \textit{w/} CoOp~\cite{coop} & TPAMI'25 &  & 66.7 & 92.9 & 89.5 & 78.0 & 92.8 & 89.5 & 79.7 & 67.0 & 89.5 & \textbf{80.9} & 67.1 & 93.4 & 82.3 \\
    \rowcolor[RGB]{221,235,247} 
    \textit{w/} LoRA~\cite{lora} & TPAMI'25 & & 66.3 & \textbf{93.2} & \textbf{89.7} & \textbf{79.4} & \textbf{93.2} & \textbf{89.6} & 79.5 & 66.4 & 89.9 & 80.0 & 66.6 & \textbf{93.6} & 82.3 \\
     \rowcolor[RGB]{217,217,217}  
    Supervised & - & & 77.9 & 91.4 & 84.4 & 74.5 & 91.4 & 84.4 & 74.5 & 77.9 & 84.4 & 74.5 & 77.9 & 91.4 & 82.0 \\\hline

    CLUE~\cite{clue} & ICCV'21 &  & 72.3 & 87.6 & 85.8 & 74.0 & 88.6 & 84.3 & 74.4 & 72.6 & 87.2 & 79.4 & 73.3 & 91.1 & 80.9 \\
    LAMDA~\cite{hwang2022combating} & ECCV'22 &  & 74.8 & 88.5 & 86.9 & 73.8 & 88.2 & 83.3 & 74.6 & 75.5 & 86.9 & 80.8 & 77.8 & 91.7 & 81.9 \\
    LAS~\cite{lada} & ICCV'23 &  & \textbf{77.8} & 91.8 & 88.4 & 77.7 & 91.5 & 87.7 & 78.1 & \textbf{79.1} & 89.5 & 83.4 & \textbf{79.8} & 94.1 & 84.9 \\
    \rowcolor[RGB]{221,235,247}
    UniMoS++ & TPAMI'25 & 10\%  & 72.6 & 94.8 & 91.4 & 82.4 & 94.7 & 91.4 & 82.9 & 72.7 & \textbf{91.9} & 83.2 & 72.0 & 95.5 & 85.5  \\ 
    \rowcolor[RGB]{221,235,247}
    \textbf{\textit{w/} post\_prp.}~\cite{denseclip} & TPAMI'25 &  & 72.9 & 94.6 & 91.8 & \textbf{82.8} & \textbf{95.4} & 91.4 & 82.5 & 73.5 & 91.7 & 83.9 & 72.5 & \textbf{95.7} & \textbf{85.7} \\ 
    \rowcolor[RGB]{221,235,247}  
    \textit{w/} CoOp~\cite{coop} & TPAMI'25 &  & 71.1 & 94.8 & \textbf{91.9} & 82.0 & 94.9 & 91.2 & 82.3 & 71.4 & 91.4 & \textbf{84.3} & 72.2 & 95.5 & 85.3 \\ 
    \rowcolor[RGB]{221,235,247}
    \textit{w/} LoRA~\cite{lora} & TPAMI'25 & & 71.1 & \textbf{95.6} & \textbf{91.9} & 82.6 & 95.0 & \textbf{92.0} & \textbf{83.7} & 71.6 & \textbf{91.9} & 84.1 & 72.8 & 95.1 & 85.6 \\\hline

    BADGE$\dagger$~\cite{badge} & ICLR'20 & \multirow{7}{*}{5\%} & 62.4 & 82.7 & 83.9 & 71.5 & 83.0 & 81.8 & 71.2 & 62.7 & 84.6 & 76.2 & 62.9 & 87.8 & 75.9 \\
    ELPT$\dagger$~\cite{li2022source} & MM'22 &  & 65.3 & 84.1 & 84.9 & 72.9 & 84.4 & 82.8 & 69.8 & 63.3 & 86.1 & 76.2 & 65.6 & 89.1 & 77.0 \\
    DAPM-TT$\dagger$~\cite{dapm} & NeurIPS'23 &  & 64.4 & 85.8 & 85.4 & 72.4 & 84.7 & 84.1 & 70.0 & 63.3 & 85.6 & 77.4 & 65.8 & 89.1 & 77.3 \\
    MHPL$\dagger$~\cite{wang2023mhpl} & CVPR'23 &  & \textbf{69.0} & 85.7 & 86.4 & 72.6 & 87.4 & 84.2 & 73.3 & \textbf{67.4} & 86.4 & \textbf{80.1} & \textbf{69.6} & 89.8 & 79.3 \\
    \rowcolor[RGB]{221,235,247}
    \textbf{UniMoS++$\dagger$} & TPAMI'25 &  & 65.3 & \textbf{92.3} & \textbf{89.4} & \textbf{77.8} & \textbf{93.0} & \textbf{89.4} & 77.7 & 63.1 & \textbf{89.2} & 78.6 & 64.6 & \textbf{92.9} & \textbf{81.1} \\ 
    \rowcolor[RGB]{221,235,247}
    \textbf{\textit{w/} post\_prp.}$\dagger$~\cite{denseclip} & TPAMI'25 &  & 65.6 & 91.9 & 89.1 & 77.7 & 92.3 & 89.2 & 77.9 & 64.0 & 88.9 & 78.4 & 65.3 & 92.5 & \textbf{81.1}  \\ 
    \rowcolor[RGB]{221,235,247}  
    \textit{w/} CoOp$\dagger$~\cite{coop} & TPAMI'25 &  & 63.8 & \textbf{92.3} & 89.2 & 76.8 & 92.2 & 89.2 & \textbf{78.1} & 64.2 & 88.9 & 78.4 & 65.2 & 92.2 & 80.9  \\ \bottomrule
    \end{tabular}}
\vspace{-8pt}
\end{table*}

\subsection{Main Results}
\label{sec_mainres}
1)  \textbf{Unsupervised domain adaptation (UDA).} \tref{officehome}, \tref{visda} and \tref{domainnet} present UDA results of UniMoS($w^*$) and UniMoS++ with different backbones, where the values $w^*$ in parentheses of UniMoS are preset ensemble weight, and `TLR' means manually tailored $w^*$ for each epoch.
As shown in \tref{officehome}, different $w^*$ in UniMoS bring significant performance fluctuation.
However, in UDA for practical scenarios, we cannot access target labels thus cannot decide the optimal $w^*$. Moreover, challenging tasks like VisDA in \tref{visda} would require manually tailored weights (TLR) for each epoch to achieve good performance, further reducing the practicability of UniMoS. The proposed UniMoS++ tackles such challenge by automatically deciding optimal $w^*$ with our MaE algorithm. In \tref{officehome} and \tref{visda}, UniMoS++ with MaE achieves comparable or better performance with the optimal manual weights in UniMoS, supporting its efficacy.
In late training epochs, we occasionally (every 3 epochs) set learned $w$ as $w^*$ as train-time weights also contain modality distribution information useful for inference.
The proposed UniMoS and UniMoS++ results significantly surpass previous ResNet- or CLIP-based methods, showcasing the effectiveness of modality-aware separation and adaptation. 
We notice that UniMoS++ is slightly behind PADCLIP~\cite{padclip} on VisDA, since PADCLIP \textbf{fully tunes} the CLIP-pretrained parameters to fill the domain gap, while our methods build on frozen CLIP encoders for computation efficiency. \cite{padclip} also mentions the domain gap caused by synthetic data in VisDA cannot be bridged without full fine-tuning. On OfficeHome UniMoS++  surpasses PADCLIP without any tuning, proving its efficiency.
On the most challenging DomainNet, UniMoS++ achieves the best overall performance, surpassing the recent SOTA DAMP~\cite{du2024domain}. On the hardest target domain \textit{qdr}, CLIP-based non-tuning methods cannot handle such large domain gap  well like full-tuning methods like \cite{pmtrans}. However, UniMoS achieves significant improvements over previous CLIP-based methods, proving the domain adaptation effects of modality-aware alignment. UniMoS++ further enhances the results (21.3\%$\to$25\%) with dynamically adjusted ensemble weights generated by MaE.

\begin{table*}[t]
    \caption{ADA results on Mini-DomainNet and VisDA with 5\% budget based on ResNet50. Best results are marked bold.}
    \vspace{-5pt}
    \centering
    \label{minidomainnet}
    \resizebox{0.9\linewidth}{!}{
    \setlength{\tabcolsep}{2.8pt}
    \begin{tabular}{p{2.1cm}|p{1.4cm}|ccccccccccccc|c}
    \toprule
    Method & Venue & cl$\to$pn & cl$\to$re & cl$\to$sk & pn$\to$cl & pn$\to$re & pn$\to$sk & re$\to$cl & re$\to$pn & re$\to$sk & sk$\to$cl & sk$\to$pn & sk$\to$re & Avg. & VisDA \\ \midrule
    BADGE~\cite{badge} & ICLR'20 & 64.3 & 80.8 & 63.5 & 65.2 & 80.2 & 63.8 & 65.9 & 65.4 & 63.4 & 66.7 & 63.3 & 79.2 & 68.5 & 84.3$\pm$0.3    \\
    CLIP~\cite{clip} & ICML'21 & 67.9 & 84.8 & 62.9 & 69.1 & 84.8 & 62.9 & 69.2 & 67.9 & 62.9 & 69.1 & 67.9 & 84.8 & 71.2 & 82.0$\pm$0.0 \\
    TQS~\cite{tqs} & CVPR'21 & 67.8 & 82.0 & 65.4 & 67.5 & 84.8 & 66.1 & 63.8 & 67.2 & 62.5 & 71.1 & 64.4 & 81.6 & 70.4 & 83.1$\pm$0.4    \\
    EADA~\cite{eada} & AAAI'22 & 66.0 & 80.8 & 63.5 & 69.4 & 83.0 & 65.1 & 71.1 & 68.6 & 65.7 & 71.0 & 64.3 & 81.0 & 70.8 & 88.3$\pm$0.1    \\
    DUC~\cite{duc} & ICLR'23 & 67.1 & 81.1 & 67.1 & 74.0 & 83.5 & 67.6 & 72.4 & 70.3 & 66.5 & 73.5 & 70.0 & 81.1 & 72.9 & 88.9$\pm$0.2    \\
    DAPM-TT~\cite{dapm} & NeurIPS'23 & 67.3 & 82.1 & 68.5 & 75.9 & 82.7 & 67.1 & 74.2 & 70.1 & 66.6 & 74.1 & 69.8 & 81.5 & 73.3 & 89.1$\pm$0.1    \\
    \rowcolor[RGB]{221,235,247}
    UniMoS++ & TPAMI'25 & 80.3 & 92.0 & 76.6 & \textbf{80.8} & 92.2 & 76.0 & \textbf{80.8} & \textbf{80.7} & 75.7 & 81.0 & 80.5 & 91.8 & \textbf{82.4} & 89.2$\pm$0.2  \\ 
    \rowcolor[RGB]{221,235,247}
    \textit{w/} post\_prp~\cite{denseclip} & TPAMI'25 & 80.0 & 91.7 & 76.5 & 80.5 & 92.0 & \textbf{76.2} & 80.6 & 80.2 & 75.5 & 80.9 & 80.4 & \textbf{92.2} & 82.2 & \textbf{89.6$\pm$0.2} \\ 
    \rowcolor[RGB]{221,235,247}
    \textit{w/} CoOp~\cite{coop} & TPAMI'25 & 79.8 & 91.9 & \textbf{76.7} & 80.2 & 92.2 & 75.9 & 80.1 & 80.0 & \textbf{76.1} & 80.9 & 79.8 & 91.8 & 82.1 & 88.1$\pm$0.2 \\ 
    \rowcolor[RGB]{221,235,247}
    \textit{w/} LoRA~\cite{lora} & TPAMI'25 & \textbf{80.5} & \textbf{92.1} & 76.6 & 80.2 & \textbf{92.4} & \textbf{76.2} & 80.4 & 80.3 & 75.7 & \textbf{81.3} & \textbf{80.6} & 92.0 & \textbf{82.4} & 88.8$\pm$0.2 \\\bottomrule
    \end{tabular}}
\vspace{-8pt}
\end{table*}

2) \textbf{Multi-source domain adaptation (MSDA).} \tref{multisrc} presents MSDA results on DomainNet and OfficeHome with different backbones following previous works. 
The UniMoS results are obtained by tailoring suitable $w^*$ weights for different datasets. 
However, such manual weights cannot handle the dynamic mixed distribution of multiple source domains during training and fall behind UniMoS++ that generates automatic weights.
With the help of MaE, UniMoS++ surpasses all recent SOTA on the most challenging MSDA benchmark DomainNet.

3) \textbf{Active domain adaptation (ADA).} \tref{active_officehome} presents ADA and source-free ADA results on OfficeHome with various annotation budgets. 
UniMoS++ is, to the best of our knowledge, the first method that solve ADA with VLMs. We also conduct experiments by replacing the manual prompts in UniMoS++ \textit{with} off-the-shelf prompt-tuning methods~\cite{denseclip, coop}. The overall performance among UniMoS++ and `\textit{w/}' prompt-tuning methods are similar, exceeding previous SOTA by up to 4.2\%. 
Although our method is designed to integrate with prompt-tuning methods, we also experiment with popular parameter-efficient fine-tuning methods like LoRA~\cite{lora}. We can find that LoRA also cooperates well with UniMoS++. Compared to prompt-tuning, incorporating LoRA performs better on easier target domains (P,R) and worse on easier ones (A,C), and achieves similar overall performances.
The averaged accuracy with 5\% annotation even surpasses that of fully-supervised training on target data, proving the superiority of VLM-based ADA. Source-free ADA (SFADA) results  are based on source-trained models without accessing source data during adaptation. 
The lack of source data leads to slightly decreased accuracies compared with ADA, but we still set new SOTA for the SFADA task. \tref{minidomainnet} presents ADA results on Mini-DomainNet and VisDA, where UniMoS++-based methods exhibit consistently outstanding performance. Especially on Mini-DomainNet, our method achieves a significant +9.1\% accuracy improvement, highlighting the effectiveness of utilizing multimodal pretrained knowledge.
The LoRA-based method achieves lower accuracies on VisDA, but as good performance on Mini-DomainNet as UniMoS++.

4) \textbf{Out-of-distribution generalization.} We  consider a more challenging and practical adaptation setting, where target domains are completely \textbf{unseen} during training. We remove the pseudo-labeling in UniMoS++ since we cannot access target data, and follow the experimentation setting in  the WILDS test-bed. We experiment on 3 WILDS \cite{koh2021wilds} benchmarks following \cite{chi2024adapting}, and report the original evaluation metrics suggested in WILDS. As shown in \tref{wild}, our method surpasses recent SOTA on CLIP \cite{chi2024adapting} in 2/3 datasets. Camelyon dataset includes medical tissue slides, where CLIP has very limited knowledge of, but we still improve the accuracy from 50.1\% (random guess) in CLIP's zero-shot inference to 89.5\%. The challenges posed by WILDS highlight the necessity of adapting to unseen expert domains, which we leave for future work to improve UniMoS++.


\begin{figure*}[!t]
  \centering
  \subfloat[]{\includegraphics[width=0.25\textwidth]{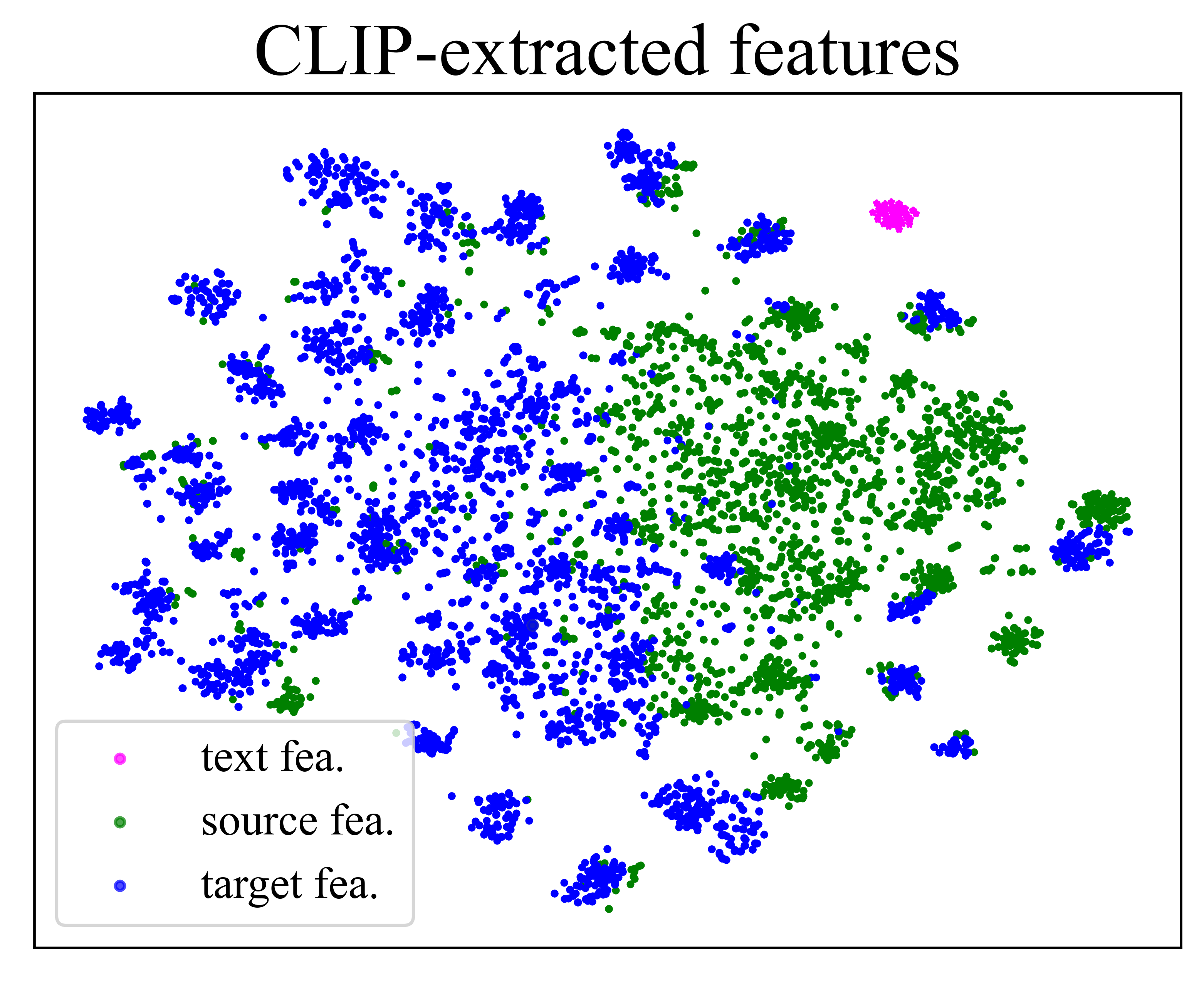}\label{fig6a}}
  \hfil
  \subfloat[]{\includegraphics[width=0.25\textwidth]{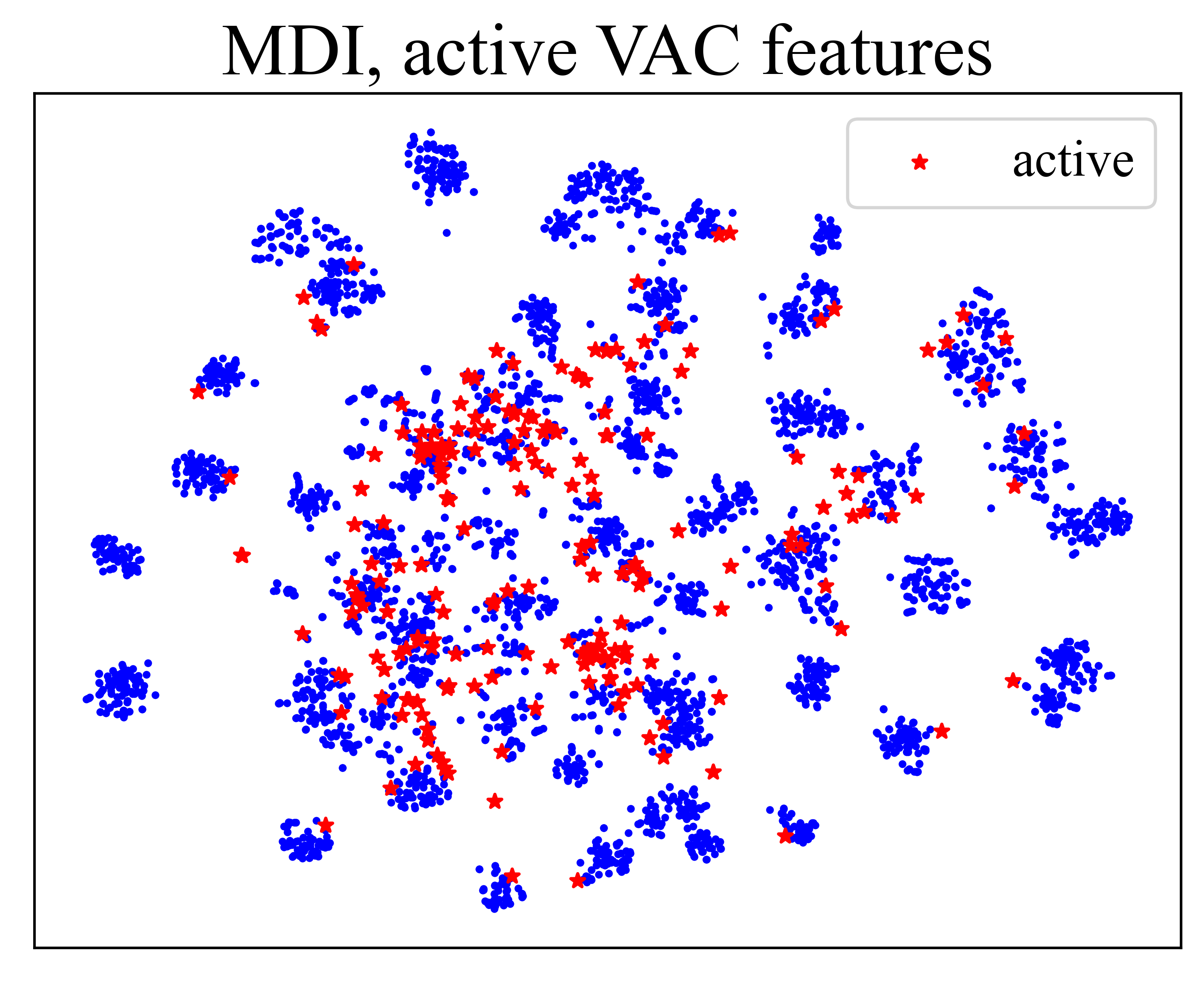}\label{fig6b}}
  \hfil
  \subfloat[]{\includegraphics[width=0.25\textwidth]{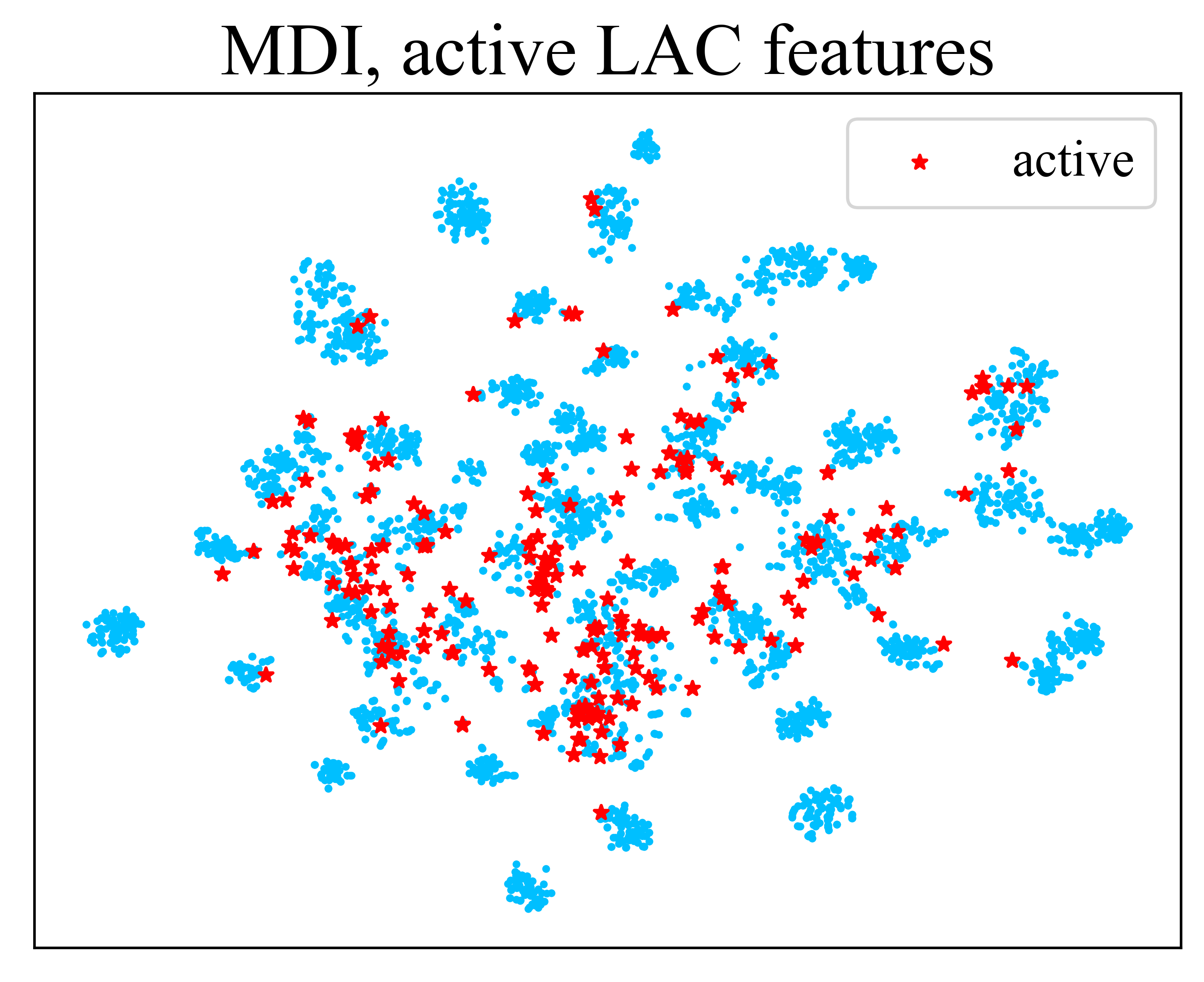}\label{fig6c}}
  \hfil
  \subfloat[]{\includegraphics[width=0.25\textwidth]{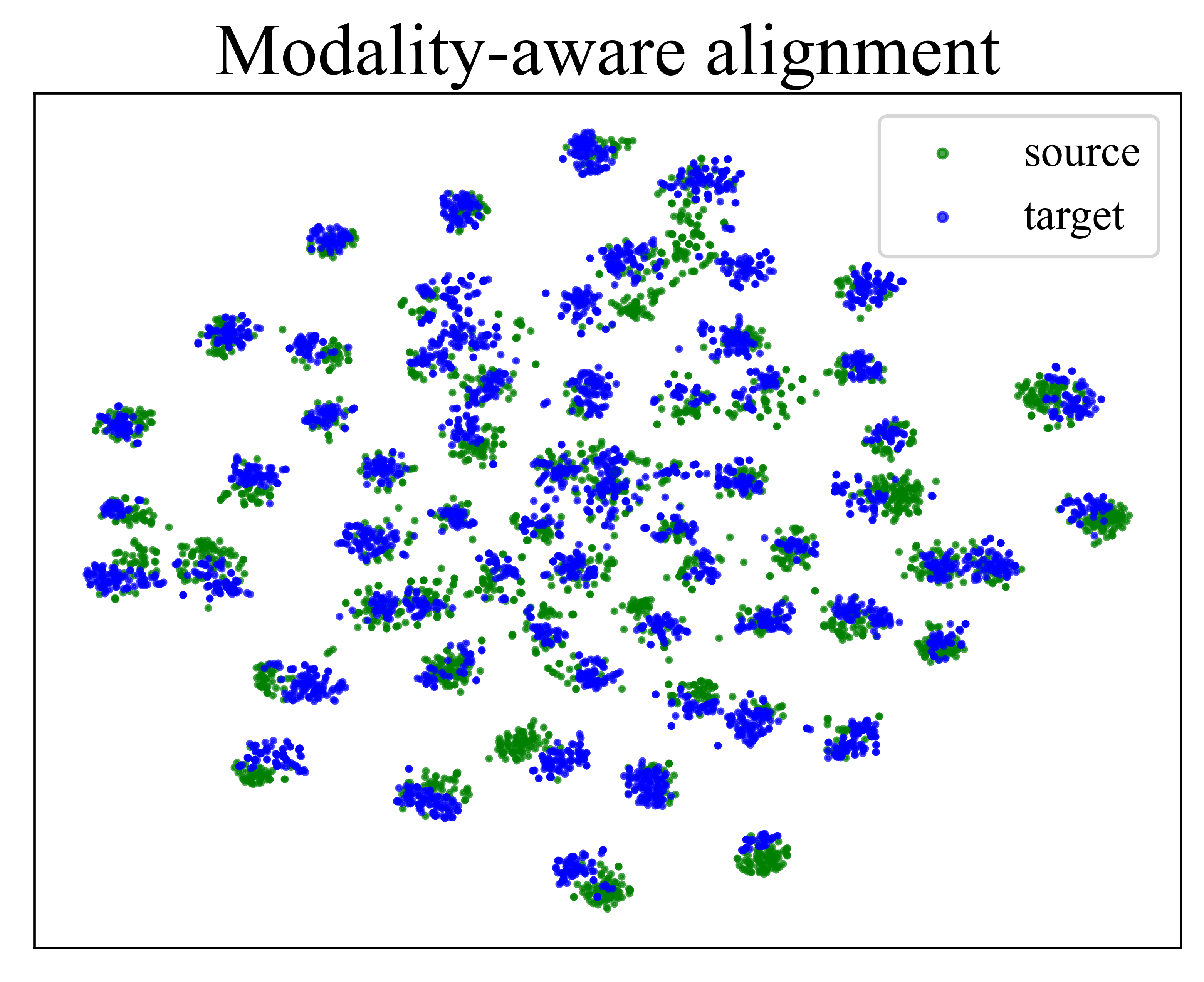}\label{fig6d}}
  \hfil
  \subfloat[]{\includegraphics[width=0.25\textwidth]{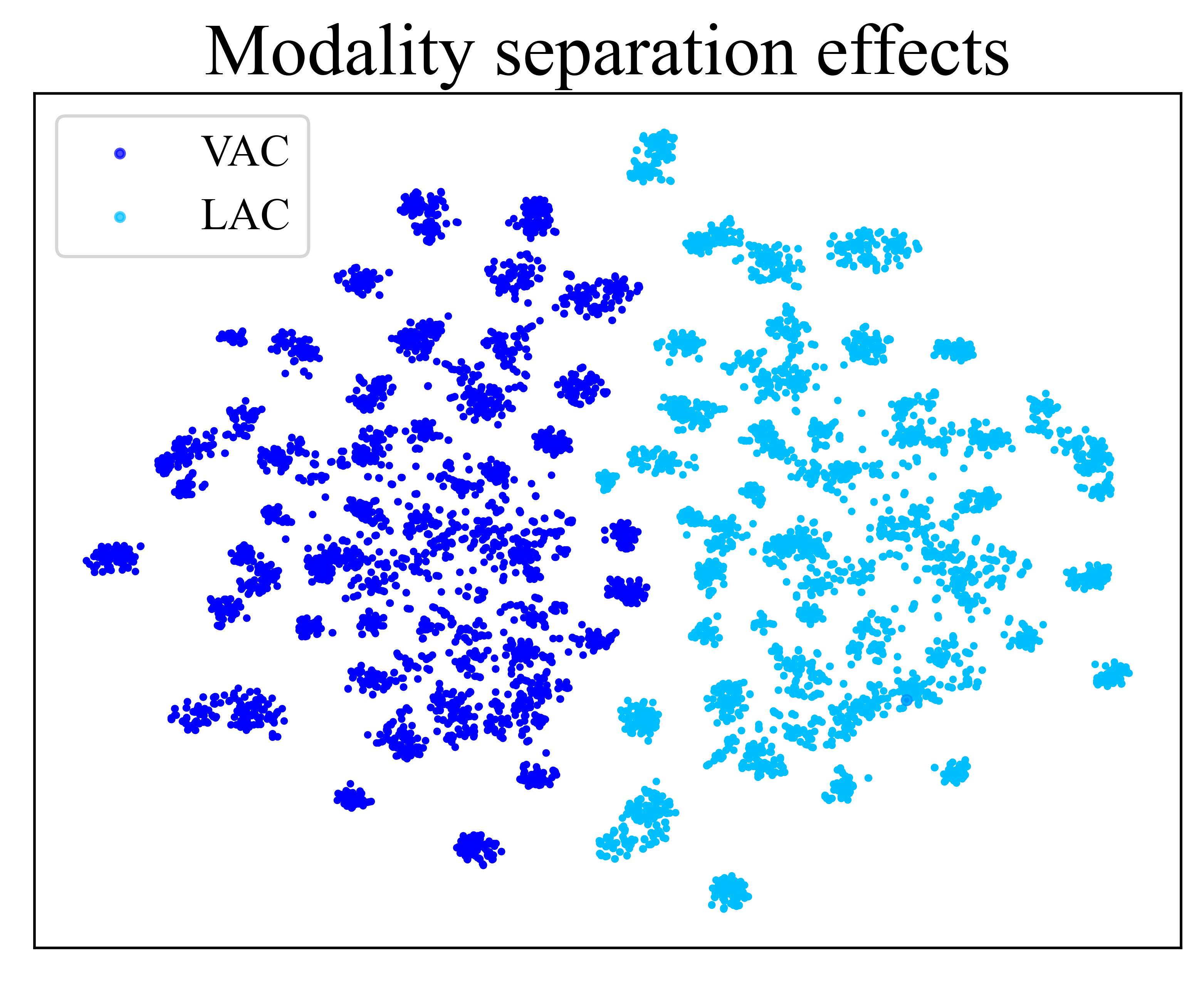}\label{fig6e}}
  \hfil
  \subfloat[]{\includegraphics[width=0.25\textwidth]{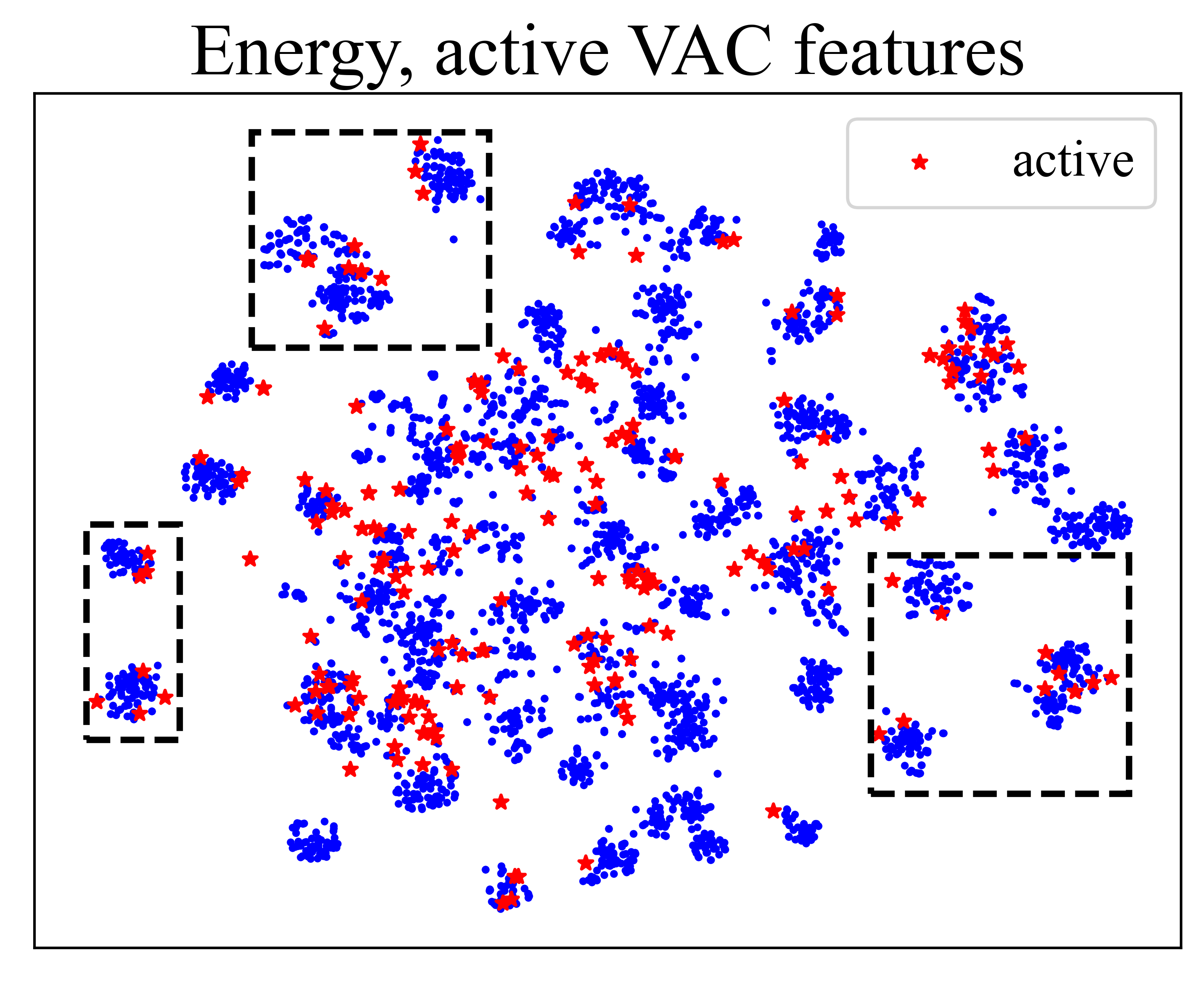}\label{fig6f}}
  \hfil
  \subfloat[]{\includegraphics[width=0.25\textwidth]{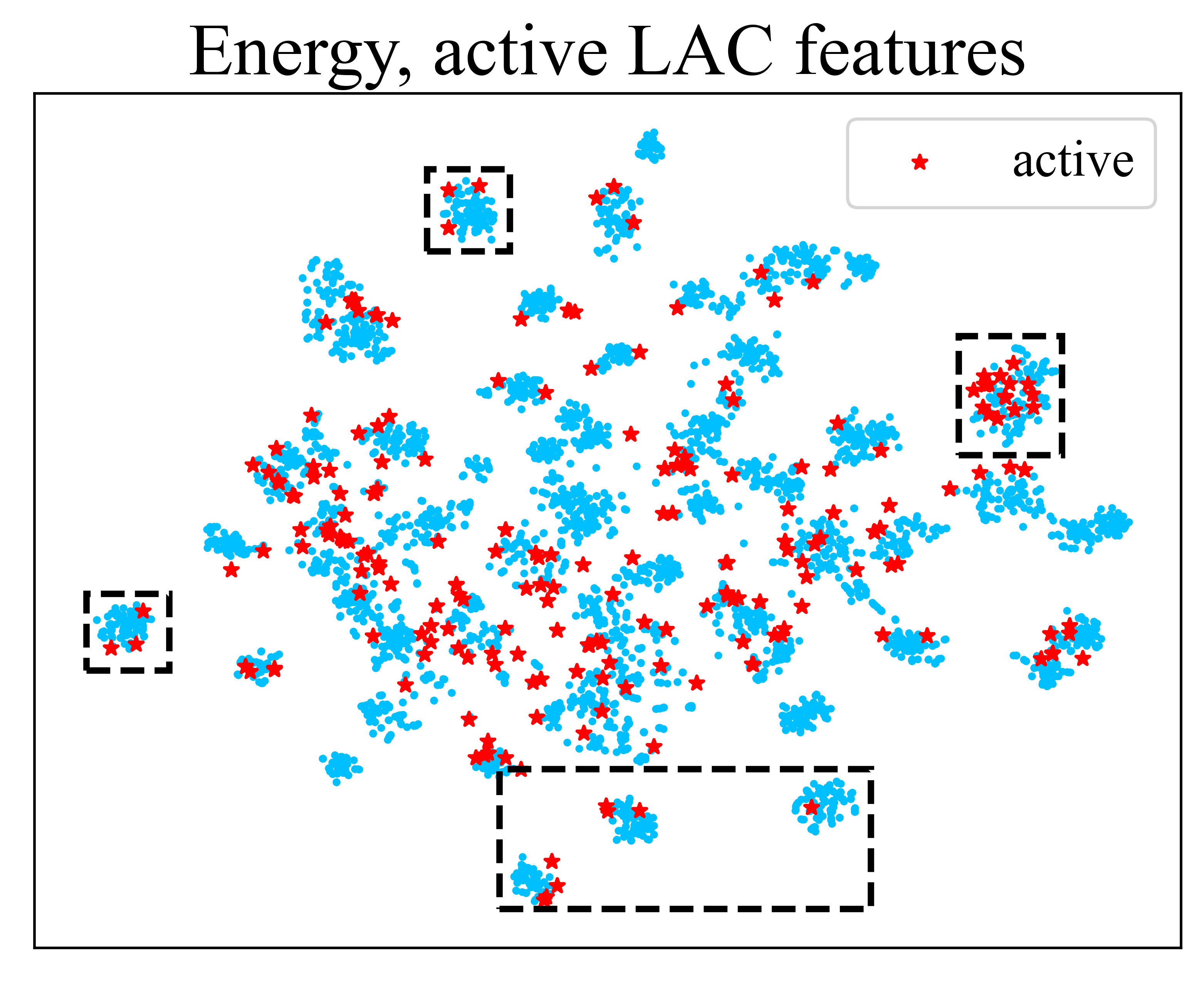}\label{fig6g}}
  \hfil
  \subfloat[]{\includegraphics[width=0.25\textwidth]{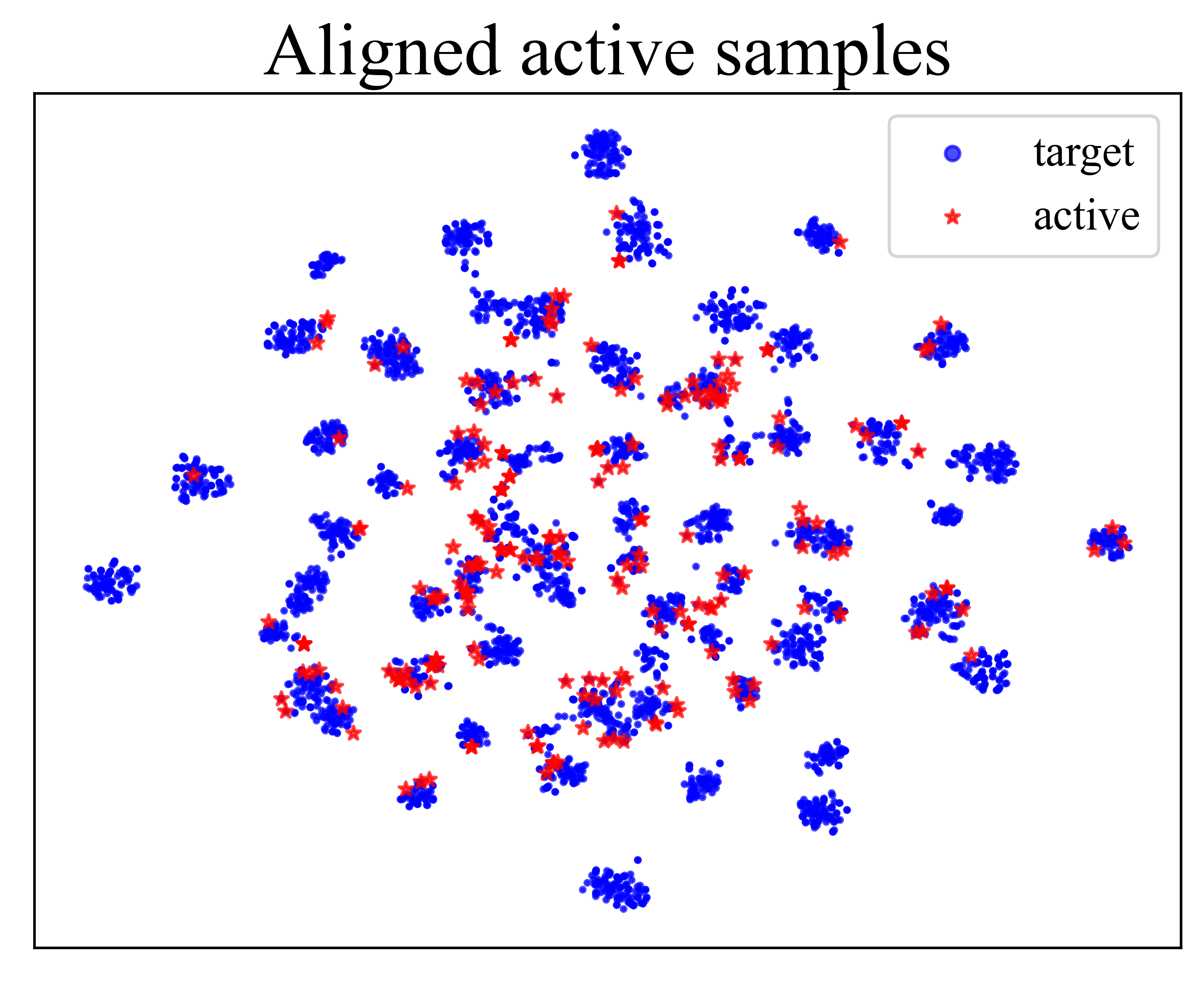}\label{fig6h}}
  \hfil
  \vspace{-5pt}
  \caption{Feature visualization on A-P, OfficeHome. (a) T-SNE visualization of CLIP-extracted vision and text features. (b,c) VAC and LAC distribution of active samples selected by MDI. (d) Aligned bottleneck feature distribution. (e) Distribution of separated target VAC and LAC. (f,g) VAC and LAC distribution of active samples selected by energy-based method. Boxes indicate low-quality active samples. (h) Aligned active sample distributions.}
  \label{fig6}
  \vspace{-12pt}
\end{figure*}

\subsection{Analysis}
1) \textbf{Ablation study.} \tref{abl} presents comprehensive ablation results on: a) Test-time ensemble weights. Weight $w^*$ in \eref{ensemble1} is set to 0.3 (optimal constant weight in \tref{officehome}), or selected by our MaE algorithm. b) Active annotation strategy. We compare energy-based active strategy (Ene.\cite{eada}) with our MDI. We also provide random selection results. c) Training process. We ablate over $\mathcal{L}_{ortho}$, $\mathcal{L}_{im}$, $\mathcal{L}_{d}$ (whether to introduce modality discriminator), whether to adopt learnable train-time weights (learn\_\textit{w}) or constant weight 0.5 in \eref{ensemble2}, whether to adopt knowledge distillation (KD) or standard cross-entropy loss in \eref{l_lac_t}.

We can draw the following  conclusions from the results in \tref{abl}. (1) When no training regularization terms are introduced, MDI+constant $w^*$ achieves the best result. Without modality-aware training and adaptation, most modality-specific information are lost, so there is no need for MaE. (2) With MaE, MDI and adding up each training component, the accuracy accordingly increases from 80.9\% to 82.4\%, suggesting that each term contributes positively. (3) With full training process, we investigate the effects of MaE and MDI. By disabling either MaE or MDI, the overall accuracy drops. The decreased performance of Energy on CLIP suggests that traditional single-modality active learning (AL) strategies cannot suit VLMs well, while our MDI provides pioneering insights for VLM-based AL methods. More analysis are given in \fref{fig6}.  (4) By replacing MDI with random selection, the accuracy drops by 2.7\%. (5) To demonstrate the necessity of modality separation and reunion, we design a simple ablation baseline where cross-modal connections (modality separation networks and modality-aware training) are removed. The two modality classifiers are adapted independently and assembled at inference. The degraded performance suggests that the cross-modal interplay during training  plays a significant role for VLMs' adaptation. 
(6) We conduct detailed analysis on alternative weighting strategies by replacing MaE with attention-based modules (variant 1) and RL-based modules (variant 2). Variant 1 requires supervised tuning on source data, but cannot well handle unlabeled target data in the presence of domain gap. Variant 2 encourages more diverse weighting strategies in an unsupervised RL-based manner \cite{shu2022hub}, but neglects modality information. Both variants exhibit reduced performance while introducing computation overheads, indicating the efficacy of MaE.

\begin{table}[t]
    \caption{OOD generalization results on WILDS. Best results are in bold. WC Acc refers to worst-case accuracies.}
    \vspace{-8pt}
    \centering
    \label{wild}
    \resizebox{\linewidth}{!}{
    \setlength{\tabcolsep}{4pt}
    \begin{tabular}{llcccc}
    \toprule
    \multicolumn{1}{l}{\multirow{2}{*}{Methods}} & \multicolumn{1}{l}{\multirow{2}{*}{Backbone}} & iWildCam\cite{beery2021iwildcam} & Camelyon17\cite{bandi2018detection} & FMoW\cite{christie2018functional} & \multicolumn{1}{c}{\multirow{2}{*}{Avg}} \\ \cmidrule(lr){3-5}
     &  & Macro F1 & Acc & WC Acc &  \\ \midrule 
     ERM & \multirow{3}{*}{CNN} & 31.0 & 70.3 & 32.3 & 44.5 \\
     IRM \cite{arjovsky2019invariant} &  & 32.8 & 64.2 & 30.0 & 42.3 \\
     Meta-DMoE \cite{zhong2022meta} &  & \textbf{34.0} & \textbf{91.4} & \textbf{35.4} & \textbf{53.6} \\ \midrule
     CLIP \cite{clip} & \multirow{3}{*}{ViT-B/16} & 9.7 & 50.1 & 14.5 & 24.8 \\
     VDPG \cite{chi2024adapting} &  & 30.1 & \textbf{93.2} & 37.8 & 53.7 \\
    \rowcolor[RGB]{221,235,247}
    UniMoS++ &  & \textbf{32.4} & 89.5 & \textbf{42.1} & \textbf{54.7} \\ \bottomrule
    \end{tabular}}
    \vspace{-12pt}
\end{table}
\begin{table}[!t]
    \caption{Ablation results of UniMoS++ on OfficeHome. Framework components in ensemble weight selection, active annotation strategy and training process are investigated. The best result (full design) is marked bold.}
    \centering
    \label{abl}
    \vspace{-5pt}
    \resizebox{\linewidth}{!}{
    \setlength{\tabcolsep}{2.5pt}
    \begin{tabular}{@{}cc|cc|ccccc|c@{}}
    \toprule
    \multicolumn{2}{c|}{Weight selection} & \multicolumn{2}{c|}{Active strategy} & \multicolumn{5}{c|}{Training} & \multicolumn{1}{c}{Result} \\ \midrule
    $w^*$=0.3 & MaE & Ene.\cite{eada} & MDI & $\mathcal{L}_{ortho}$ & $\mathcal{L}_{im}$ & $\mathcal{L}_{bce}$ & learn\_\textit{w} & KD & OH \\ \midrule
    \checkmark &  & \checkmark &  &  &  &  &  &  & 80.3 \\
    \checkmark &  &  & \checkmark &  &  &  &  &  & 81.2 \\
     & \checkmark &  & \checkmark &  &  &  &  &  & 80.7 \\ \midrule
     & \checkmark &  & \checkmark & \checkmark &  &  &  &  & 80.9 \\
     & \checkmark &  & \checkmark & \checkmark & \checkmark &  &  &  & 81.6 \\
     & \checkmark &  & \checkmark & \checkmark & \checkmark & \checkmark &  &  & 81.7 \\
     & \checkmark &  & \checkmark & \checkmark & \checkmark & \checkmark & \checkmark &  & 82.0 \\ \midrule
    \checkmark &  & \checkmark &  & \checkmark & \checkmark & \checkmark & \checkmark & \checkmark & 81.0 \\
    \checkmark &  &  & \checkmark & \checkmark & \checkmark & \checkmark & \checkmark & \checkmark & 81.9 \\
     & \checkmark & \checkmark &  & \checkmark & \checkmark & \checkmark & \checkmark & \checkmark & 81.0 \\
     \rowcolor[RGB]{221,235,247}
     & \checkmark &  & \checkmark & \checkmark & \checkmark & \checkmark & \checkmark & \checkmark & \textbf{82.4} \\ \midrule
     & \checkmark & \multicolumn{2}{c|}{Random} & \checkmark & \checkmark & \checkmark & \checkmark & \checkmark & 79.7 \\ \midrule
    \multicolumn{9}{l|}{Baseline: Adapt vision and text classifier independently then assemble} & 81.2 \\ \midrule
    \multicolumn{9}{l|}{\textit{Analysis on weight generation strategies (UDA tasks).}} & \\
    \multicolumn{9}{l|}{Variant 1: Replace MaE with cross-modality attention modules \cite{li2024agile}.} & 77.2 \\
    \multicolumn{9}{l|}{Variant 2: Replace MaE with maximum-entropy RL modules \cite{shu2022hub}.} & 76.8 \\ 
    \rowcolor[RGB]{221,235,247}
    \multicolumn{9}{l|}{Full design with MaE.} & \textbf{78.0} \\\bottomrule
    \end{tabular}}
\vspace{-10pt}
\end{table}

2) \textbf{Feature visualization.} \fref{fig6}  presents detailed feature visualization using t-SNE~\cite{van2008visualizing}. \fref{fig6a} shows CLIP-extracted vision features from source (green) and target (blue) domain, along with CLIP-extracted text features (pink) of naive prompts. The source and target features are misaligned, and distribution between modalities also diverges. \fref{fig6e} shows separated target VAC and LAC. A clear boundary can be observed between the features of both modalities, indicating the effects of modality separation. \fref{fig6}(b,c,f,g) visualizes the VAC (LAC) distributions of selected active samples  using different active strategies (MDI versus Energy\cite{eada}). By comparing \fref{fig6b} and \fref{fig6f}, we  observe that Energy-based method selects well-clustered samples (red dots in boxes). These samples have been correctly classified, thus \textit{cannot} contribute to the annotation informativeness. 
Similar observations can be made by comparing \fref{fig6c} and \fref{fig6g}.
The reason is that Energy-based method does not take modality information into consideration, therefore selecting well-learned modality-specific samples. 
The accuracy of active samples by MDI is 19.5\% lower than that by Energy, indicating more informative samples are retrieved.
\fref{fig6d} shows that vision features from both domains display a compact and aligned structure compared to \fref{fig6a}, proving the success of modality-aware adaptation. \fref{fig6h} presents the distribution of active samples after training. Compared to \fref{fig6b}, most active samples have been well learned.

\begin{figure*}[!t]
  \centering
  \subfloat[]{\includegraphics[width=0.25\textwidth]{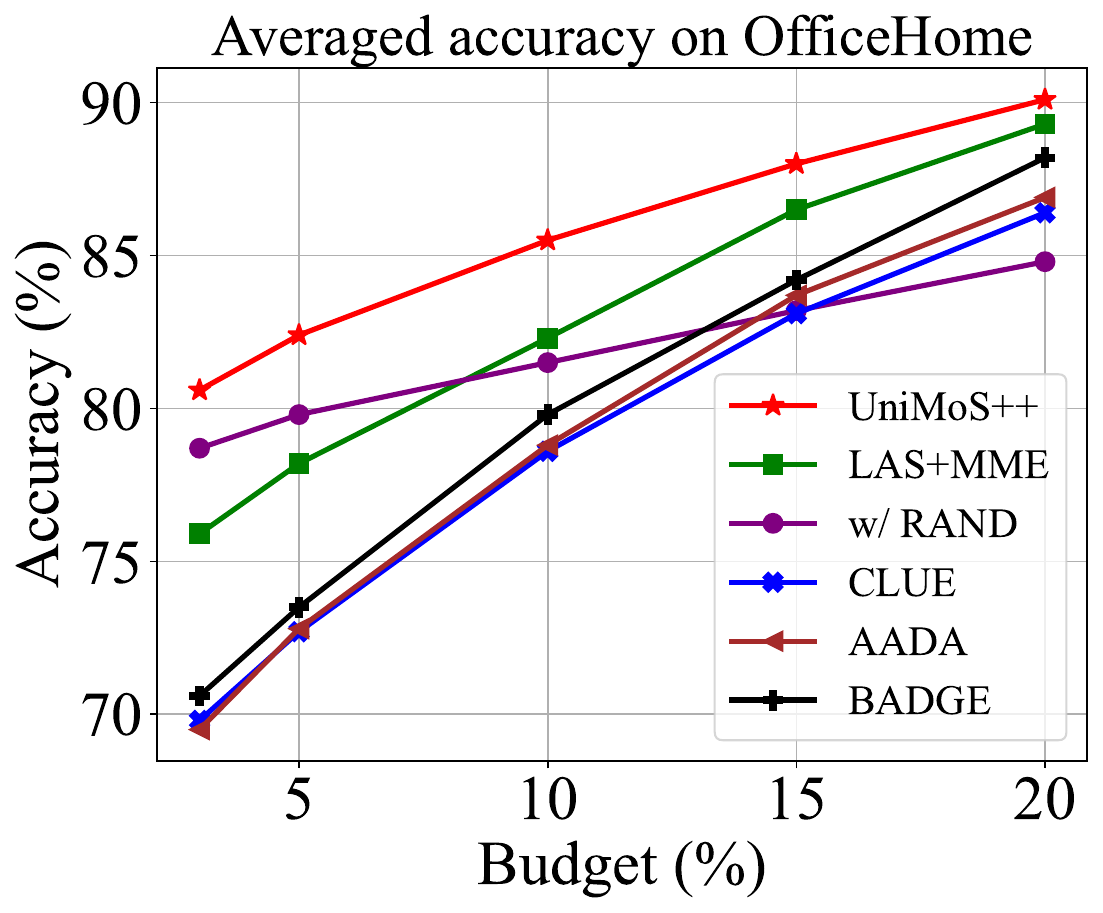}\label{fig7a}}
  \hfil
  \subfloat[]{\includegraphics[width=0.25\textwidth]{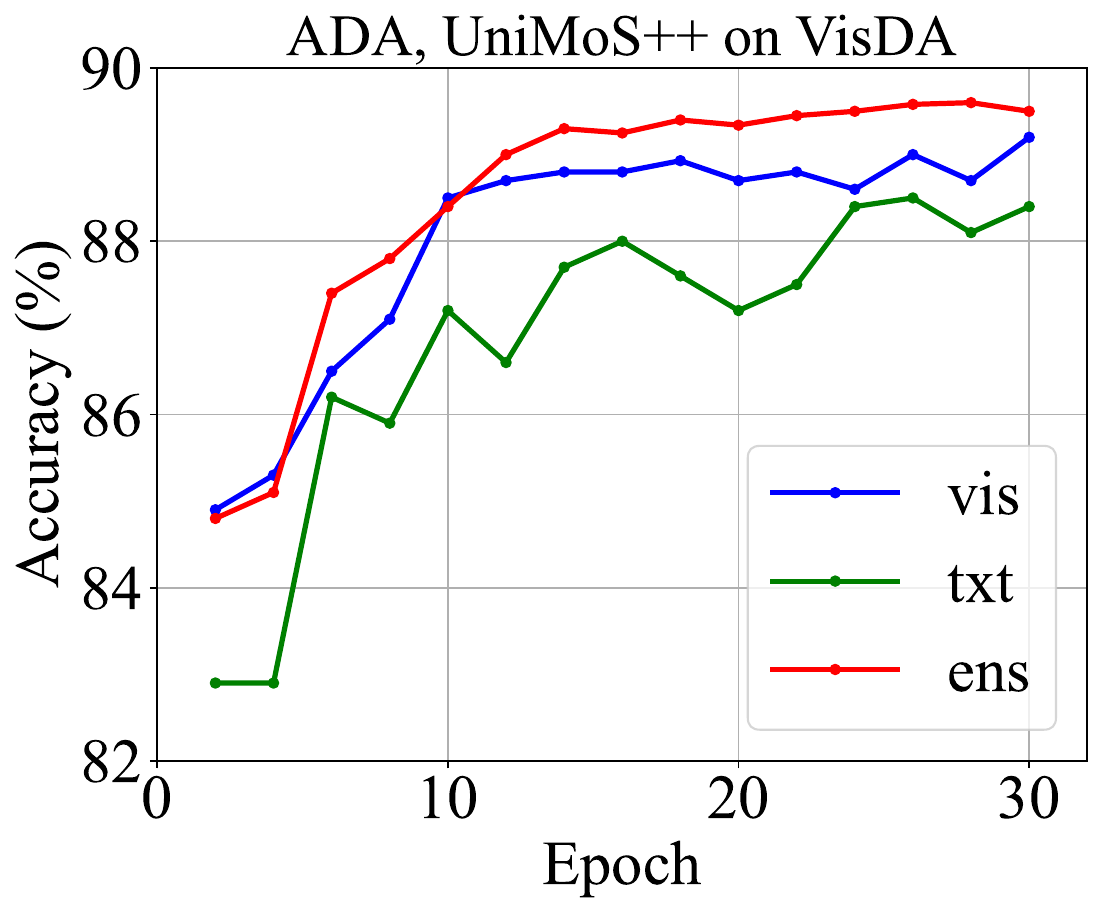}\label{fig7b}}
  \hfil
  \subfloat[]{\includegraphics[width=0.25\textwidth]{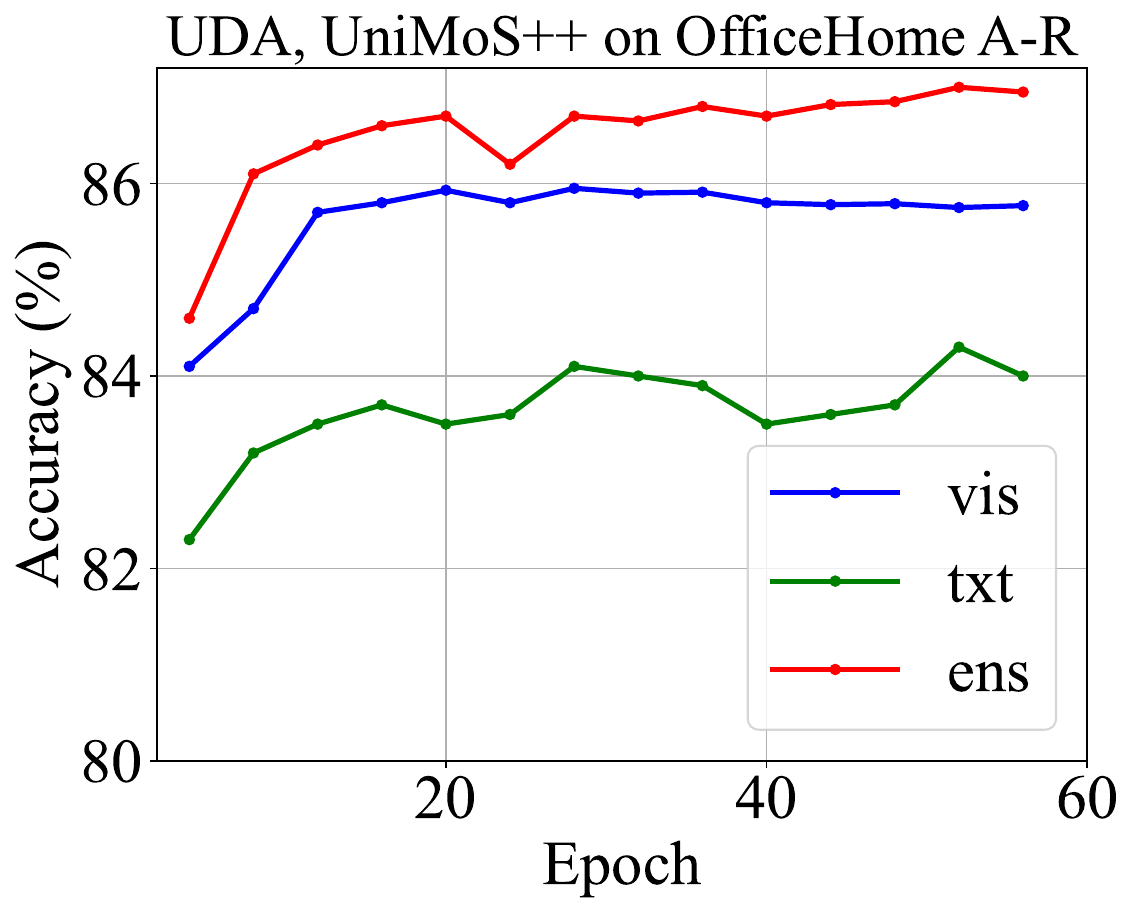}\label{fig7c}}
  \hfil
  \subfloat[]{\includegraphics[width=0.25\textwidth]{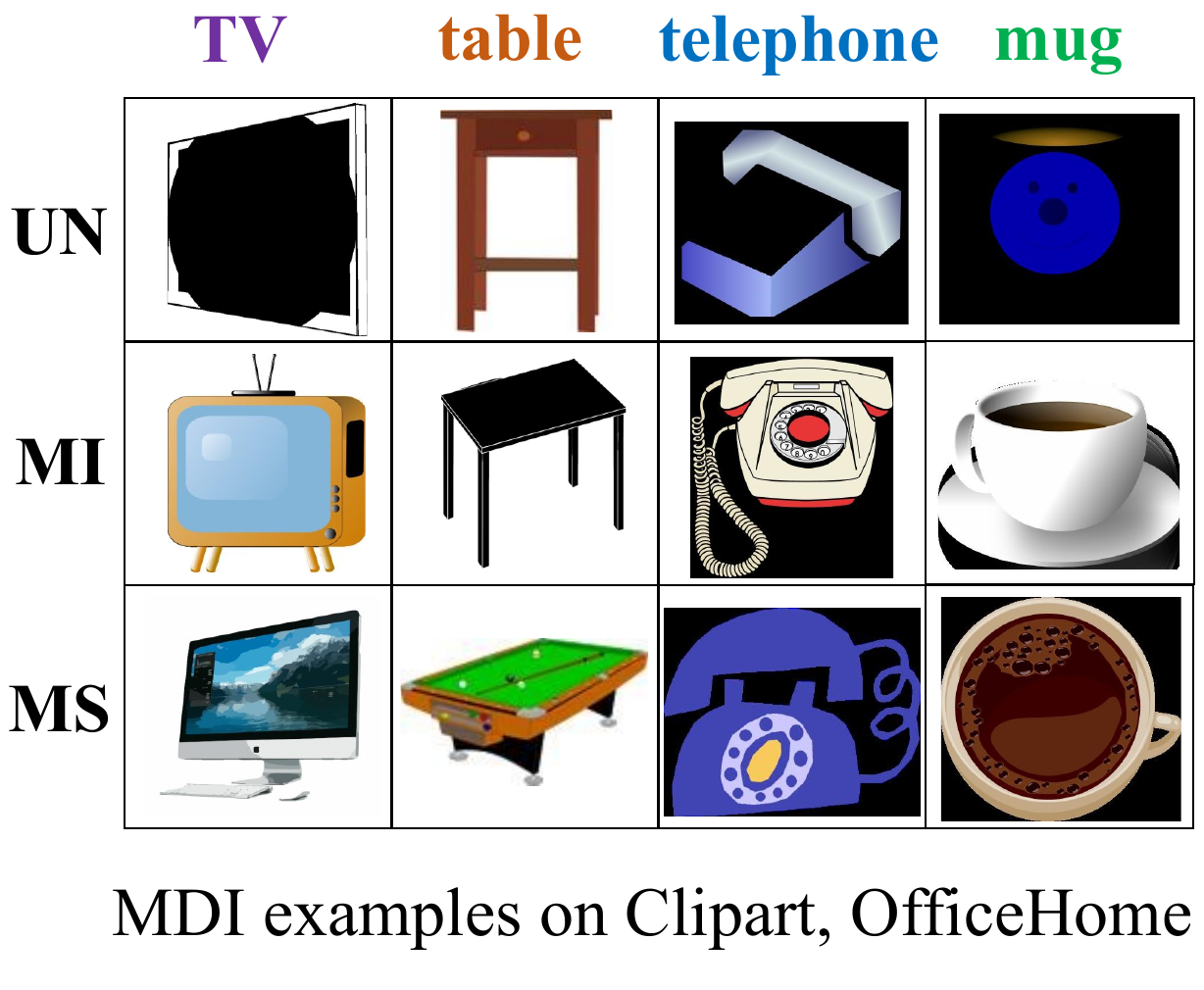}\label{fig7d}}
  \hfil
  \caption{(a) ADA results of different methods on OfficeHome with different active budgets. (b,c) Accuracies of vision, text and ensemble outputs of ADA/UDA tasks. (d) MDI examples of MS, MI and UN samples from domain C on Office-Home.}
  \label{fig7}
  \vspace{-10pt}
\end{figure*}
\begin{table}[!t]
    \caption{Computation analysis on OfficeHome. Best results are bolded.}
    \vspace{-8pt}
    \centering
    \label{computation}
    \resizebox{1\linewidth}{!}{
    \setlength{\tabcolsep}{4pt}
    \begin{tabular}{p{1.9cm}|ccccc}
    \toprule
    Method & Param.(M) & Throughput & Train time(H) & GFLOPs & Accuracy \\ \midrule
    DAPM-TT~\cite{du2023diffusion} & 50.71 & 102 & 1.10 & 16.53 & 77.1 \\
    EADA~\cite{eada} & 47.56 & 105 & 0.51 & 16.52 & 76.7 \\
    LAS~\cite{lada} & 24.05 & 158 & 0.35 & 4.13 & 78.2 \\
    \rowcolor[RGB]{221,235,247} 
    UniMoS++ & \textbf{2.65} & \textbf{1992} & \textbf{0.06} & \textbf{0.04} & \textbf{82.4} \\
    \rowcolor[RGB]{221,235,247} 
    w/ post\_prp & 9.75 & 676 & 0.17 & 0.45 & \textbf{82.4} \\
    \rowcolor[RGB]{221,235,247} 
    w/ coop & 2.68 & 623 & 0.13 & 1.97 & 82.3 \\ \bottomrule
    \end{tabular}}
    \vspace{-10pt}
\end{table}

3) \textbf{Training details and budget analysis.} \fref{fig7a} compares UniMoS++
and other ADA methods with active budget ranging from 3\% to 20\%, where our method exhibits clear superiority across all budgets. `UniMos++ w/ RAND'  performs poorly as the budgets increase, since it is harder to select more informative samples by chance. \fref{fig7b} presents the complete accuracy curves of the vision, text classifier and modality ensemble results for ADA task. Both vision and text accuracies climb as the training proceeds, with the ensemble accuracy consistently surpassing both modalities, suggesting the necessity to combine modality-specific information.
\fref{fig7c} shows accuracy curves on UDA tasks. Without active annotations, both modality accuracies first increase then remain stable. The ensemble accuracy climbs steadily with more suitable ensemble weight selected by MaE. \fref{fig7d} presents real examples of MDI categorization results. MI samples are representative among the class. MS samples usually contain richer semantic details but are atypical among the category, e.g., the MS table is actually a \textit{pool table}. UN samples are ambiguous and hard to recognize even for humans, making them suitable for annotation.

4) \textbf{Computation analysis.} \tref{computation}  evaluates the computational costs of UniMoS++-based methods and competing methods under the same platform. The `Train time' metric is evaluated on task A-P. It can be concluded our methods cost the least among all metrics, while achieving the best accuracy. UniMoS++ has significantly lower trainable parameters since no updates on CLIP's pretrained encoders are needed. Such design brings extremely high throughput, low training time and FLOPs as few forwards and backwards through CLIP encoders are needed. Equipped with any prompt-tuning method, UniMoS++ can agilely adapt large pretrained VLMs with few costs and annotation needs, providing a practically available solution for vision-language model adaptation with limited resources.

\begin{figure}[!t]
  \centering
  \subfloat[]{\includegraphics[width=0.22\textwidth]{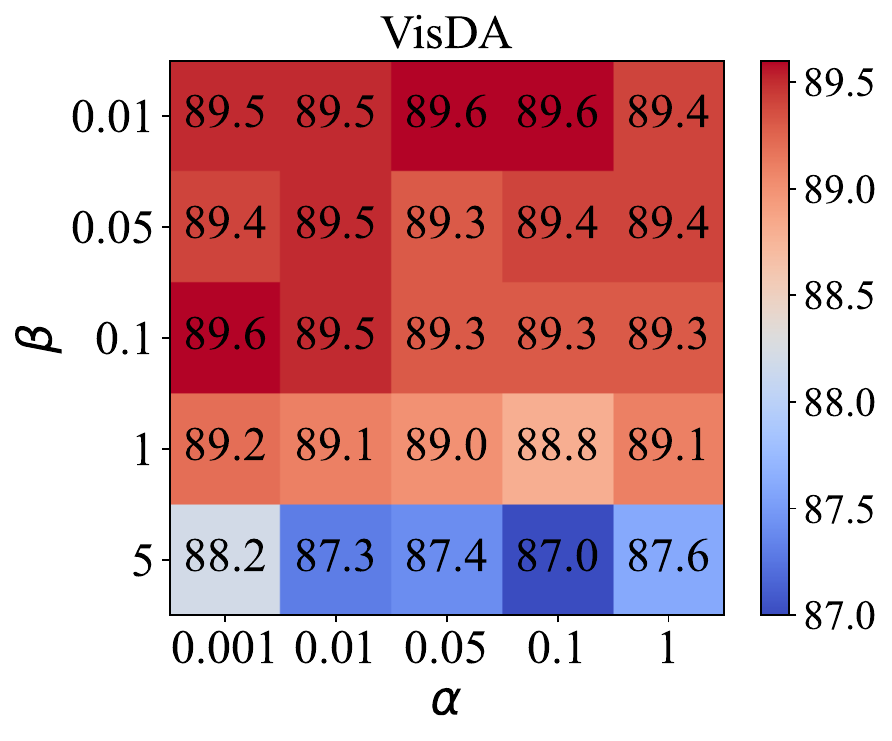}\label{fig8a}}
  \hfil
  \subfloat[]{\includegraphics[width=0.22\textwidth]{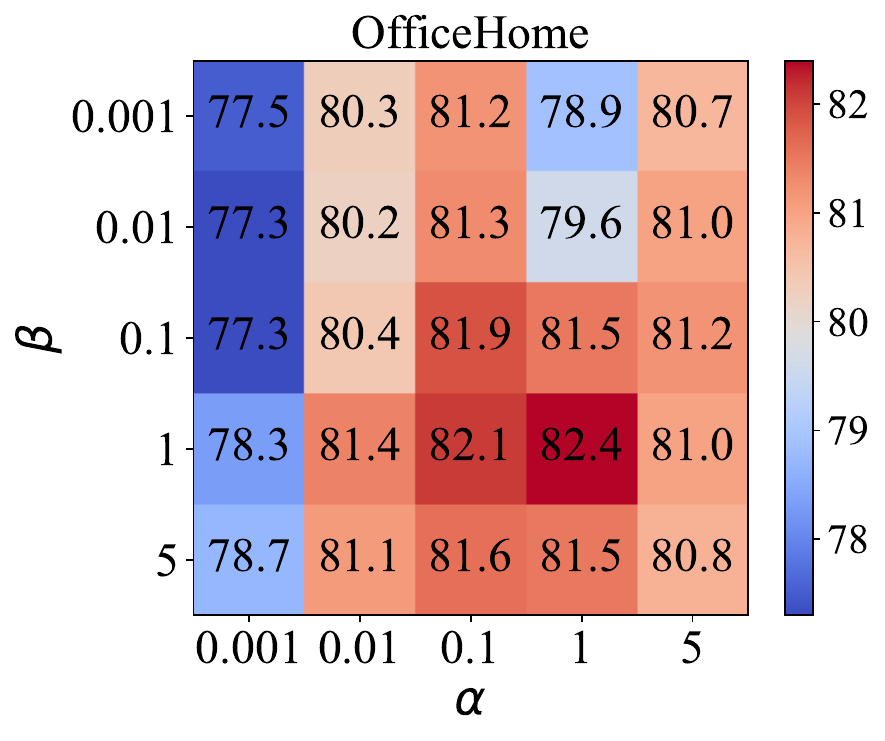}\label{fig8b}}
  \hfil
  \vspace{-5pt}
  \caption{Parameter sensitivity analysis on $\alpha$ and $\beta$ of UniMoS++ for ADA.}
  \label{fig8}
  \vspace{-10pt}
\end{figure}

5) \textbf{Parameter sensitivity.} 
We introduce hyperparameters $\alpha,\beta,\gamma$ in \eref{l_lac}, \eref{l_vac}, \eref{l_all},
among which $\alpha,\beta$ are pivotal to control source importance in domain adaptation. \fref{fig8} presents accuracies of UniMoS++ with different $\alpha,\beta$ combinations. On VisDA, smaller source weights contributes to  better outcomes. As described in Sec.\ref{sec_mainres}, pretrained CLIP struggles to identify the synthetic source data in VisDA. Therefore, down-weighting source supervision  proves beneficial. On OfficeHome, source and target data from both modalities are equally important. Simply setting $\alpha$=1 and $\beta$=1 leads to the best performances.

\section{Conclusion}
VLMs have shown promising success in unsupervised domain adaptation (UDA) tasks.
Based on our analysis and the modality gap theory, we discover the loss of modality-specific information in current VLM-based UDA methods. As a solution, we propose a  unified modality separation framework to  disentangle VLM-extracted features into vision- and language-associated components to protect and combine their modality strengths for adaptation. During training, different modality components are handled separately in a unified manner to exploit modality-specific knowledge, while at test-time a modality-adaptive ensemble algorithm  determines ensemble weights to maximize the cross-modal integration effects. We further design a modality discrepancy metric to evaluate sample modality characteristics, contributing to instance-level domain  adaptation and  active learning strategies.
Our methods are compatible with various prompt tuning techniques and a spectrum of domain adaptation settings, providing insights and effective solutions for effectively adapting VLMs.

\ifCLASSOPTIONcaptionsoff
  \newpage
\fi



\bibliographystyle{IEEEtran}
\bibliography{main}

\begin{thebibliography}{10}
\providecommand{\url}[1]{#1}
\csname url@samestyle\endcsname
\providecommand{\newblock}{\relax}
\providecommand{\bibinfo}[2]{#2}
\providecommand{\BIBentrySTDinterwordspacing}{\spaceskip=0pt\relax}
\providecommand{\BIBentryALTinterwordstretchfactor}{4}
\providecommand{\BIBentryALTinterwordspacing}{\spaceskip=\fontdimen2\font plus
\BIBentryALTinterwordstretchfactor\fontdimen3\font minus
  \fontdimen4\font\relax}
\providecommand{\BIBforeignlanguage}[2]{{%
\expandafter\ifx\csname l@#1\endcsname\relax
\typeout{** WARNING: IEEEtran.bst: No hyphenation pattern has been}%
\typeout{** loaded for the language `#1'. Using the pattern for}%
\typeout{** the default language instead.}%
\else
\language=\csname l@#1\endcsname
\fi
#2}}
\providecommand{\BIBdecl}{\relax}
\BIBdecl

\bibitem{ganin2015unsupervised}
Y.~Ganin and V.~Lempitsky, ``Unsupervised domain adaptation by
  backpropagation,'' in \emph{International conference on machine
  learning}.\hskip 1em plus 0.5em minus 0.4em\relax PMLR, 2015, pp. 1180--1189.

\bibitem{saito2018maximum}
K.~Saito, K.~Watanabe, Y.~Ushiku, and T.~Harada, ``Maximum classifier
  discrepancy for unsupervised domain adaptation,'' in \emph{Proceedings of the
  IEEE conference on computer vision and pattern recognition}, 2018, pp.
  3723--3732.

\bibitem{ben2010theory}
S.~Ben-David, J.~Blitzer, K.~Crammer, A.~Kulesza, F.~Pereira, and J.~W.
  Vaughan, ``A theory of learning from different domains,'' \emph{Machine
  learning}, vol.~79, pp. 151--175, 2010.

\bibitem{long2015learning}
M.~Long, Y.~Cao, J.~Wang, and M.~Jordan, ``Learning transferable features with
  deep adaptation networks,'' in \emph{International conference on machine
  learning}.\hskip 1em plus 0.5em minus 0.4em\relax PMLR, 2015, pp. 97--105.

\bibitem{long2017deep}
M.~Long, H.~Zhu, J.~Wang, and M.~I. Jordan, ``Deep transfer learning with joint
  adaptation networks,'' in \emph{International conference on machine
  learning}.\hskip 1em plus 0.5em minus 0.4em\relax PMLR, 2017, pp. 2208--2217.

\bibitem{goodfellow2014generative}
I.~Goodfellow, J.~Pouget-Abadie, M.~Mirza, B.~Xu, D.~Warde-Farley, S.~Ozair,
  A.~Courville, and Y.~Bengio, ``Generative adversarial nets,'' \emph{Advances
  in neural information processing systems}, vol.~27, 2014.

\bibitem{long2018conditional}
M.~Long, Z.~Cao, J.~Wang, and M.~I. Jordan, ``Conditional adversarial domain
  adaptation,'' \emph{Advances in neural information processing systems},
  vol.~31, 2018.

\bibitem{pmtrans}
J.~Zhu, H.~Bai, and L.~Wang, ``Patch-mix transformer for unsupervised domain
  adaptation: A game perspective,'' in \emph{Proceedings of the IEEE/CVF
  conference on computer vision and pattern recognition}, 2023, pp. 3561--3571.

\bibitem{cdtrans}
T.~Xu, W.~Chen, P.~Wang, F.~Wang, H.~Li, and R.~Jin, ``Cdtrans: Cross-domain
  transformer for unsupervised domain adaptation,'' \emph{arXiv preprint
  arXiv:2109.06165}, 2021.

\bibitem{clip}
A.~Radford, J.~W. Kim, C.~Hallacy, A.~Ramesh, G.~Goh, S.~Agarwal, G.~Sastry,
  A.~Askell, P.~Mishkin, J.~Clark \emph{et~al.}, ``Learning transferable visual
  models from natural language supervision,'' in \emph{International conference
  on machine learning}.\hskip 1em plus 0.5em minus 0.4em\relax PMLR, 2021, pp.
  8748--8763.

\bibitem{align}
C.~Jia, Y.~Yang, Y.~Xia, Y.-T. Chen, Z.~Parekh, H.~Pham, Q.~Le, Y.-H. Sung,
  Z.~Li, and T.~Duerig, ``Scaling up visual and vision-language representation
  learning with noisy text supervision,'' in \emph{International conference on
  machine learning}.\hskip 1em plus 0.5em minus 0.4em\relax PMLR, 2021, pp.
  4904--4916.

\bibitem{lester2021power}
B.~Lester, R.~Al-Rfou, and N.~Constant, ``The power of scale for
  parameter-efficient prompt tuning,'' \emph{arXiv preprint arXiv:2104.08691},
  2021.

\bibitem{du2024domain}
Z.~Du, X.~Li, F.~Li, K.~Lu, L.~Zhu, and J.~Li, ``Domain-agnostic mutual
  prompting for unsupervised domain adaptation,'' in \emph{Proceedings of the
  IEEE/CVF Conference on Computer Vision and Pattern Recognition}, 2024, pp.
  23\,375--23\,384.

\bibitem{daprompt}
C.~Ge, R.~Huang, M.~Xie, Z.~Lai, S.~Song, S.~Li, and G.~Huang, ``Domain
  adaptation via prompt learning,'' \emph{IEEE Transactions on Neural Networks
  and Learning Systems}, 2023.

\bibitem{padclip}
Z.~Lai, N.~Vesdapunt, N.~Zhou, J.~Wu, C.~P. Huynh, X.~Li, K.~K. Fu, and C.-N.
  Chuah, ``Padclip: Pseudo-labeling with adaptive debiasing in clip for
  unsupervised domain adaptation,'' in \emph{Proceedings of the IEEE/CVF
  International Conference on Computer Vision}, 2023, pp. 16\,155--16\,165.

\bibitem{liang2022mind}
V.~W. Liang, Y.~Zhang, Y.~Kwon, S.~Yeung, and J.~Y. Zou, ``Mind the gap:
  Understanding the modality gap in multi-modal contrastive representation
  learning,'' \emph{Advances in Neural Information Processing Systems},
  vol.~35, pp. 17\,612--17\,625, 2022.

\bibitem{shot}
J.~Liang, D.~Hu, and J.~Feng, ``Do we really need to access the source data?
  source hypothesis transfer for unsupervised domain adaptation,'' in
  \emph{International conference on machine learning}.\hskip 1em plus 0.5em
  minus 0.4em\relax PMLR, 2020, pp. 6028--6039.

\bibitem{menonvisual}
S.~Menon and C.~Vondrick, ``Visual classification via description from large
  language models,'' in \emph{The Eleventh International Conference on Learning
  Representations}.

\bibitem{zhang2024connect}
Y.~Zhang, E.~Sui, and S.~Yeung-Levy, ``Connect, collapse, corrupt: Learning
  cross-modal tasks with uni-modal data,'' \emph{arXiv preprint
  arXiv:2401.08567}, 2024.

\bibitem{gao2024clip}
P.~Gao, S.~Geng, R.~Zhang, T.~Ma, R.~Fang, Y.~Zhang, H.~Li, and Y.~Qiao,
  ``Clip-adapter: Better vision-language models with feature adapters,''
  \emph{International Journal of Computer Vision}, vol. 132, no.~2, pp.
  581--595, 2024.

\bibitem{selvaraju2017grad}
R.~R. Selvaraju, M.~Cogswell, A.~Das, R.~Vedantam, D.~Parikh, and D.~Batra,
  ``Grad-cam: Visual explanations from deep networks via gradient-based
  localization,'' in \emph{Proceedings of the IEEE international conference on
  computer vision}, 2017, pp. 618--626.

\bibitem{xie2022active}
B.~Xie, L.~Yuan, S.~Li, C.~H. Liu, X.~Cheng, and G.~Wang, ``Active learning for
  domain adaptation: An energy-based approach,'' in \emph{Proceedings of the
  AAAI conference on artificial intelligence}, vol.~36, no.~8, 2022, pp.
  8708--8716.

\bibitem{du2023diffusion}
Z.~Du and J.~Li, ``Diffusion-based probabilistic uncertainty estimation for
  active domain adaptation,'' \emph{Advances in Neural Information Processing
  Systems}, vol.~36, pp. 17\,129--17\,155, 2023.

\bibitem{unimos}
X.~Li, Y.~Li, Z.~Du, F.~Li, K.~Lu, and J.~Li, ``Split to merge: Unifying
  separated modalities for unsupervised domain adaptation,'' in
  \emph{Proceedings of the IEEE/CVF Conference on Computer Vision and Pattern
  Recognition}, 2024, pp. 23\,364--23\,374.

\bibitem{pan2009survey}
S.~J. Pan and Q.~Yang, ``A survey on transfer learning,'' \emph{IEEE
  Transactions on knowledge and data engineering}, vol.~22, no.~10, pp.
  1345--1359, 2009.

\bibitem{li2020maximum}
J.~Li, E.~Chen, Z.~Ding, L.~Zhu, K.~Lu, and H.~T. Shen, ``Maximum density
  divergence for domain adaptation,'' \emph{IEEE transactions on pattern
  analysis and machine intelligence}, vol.~43, no.~11, pp. 3918--3930, 2020.

\bibitem{yan2017mind}
H.~Yan, Y.~Ding, P.~Li, Q.~Wang, Y.~Xu, and W.~Zuo, ``Mind the class weight
  bias: Weighted maximum mean discrepancy for unsupervised domain adaptation,''
  in \emph{Proceedings of the IEEE conference on computer vision and pattern
  recognition}, 2017, pp. 2272--2281.

\bibitem{yang2021exploiting}
S.~Yang, J.~Van~de Weijer, L.~Herranz, S.~Jui \emph{et~al.}, ``Exploiting the
  intrinsic neighborhood structure for source-free domain adaptation,''
  \emph{Advances in neural information processing systems}, vol.~34, pp.
  29\,393--29\,405, 2021.

\bibitem{yang2023trust}
S.~Yang, Y.~Wang, J.~Van~de Weijer, L.~Herranz, S.~Jui, and J.~Yang, ``Trust
  your good friends: Source-free domain adaptation by reciprocal neighborhood
  clustering,'' \emph{IEEE Transactions on pattern analysis and machine
  intelligence}, 2023.

\bibitem{vit}
A.~Dosovitskiy, L.~Beyer, A.~Kolesnikov, D.~Weissenborn, X.~Zhai,
  T.~Unterthiner, M.~Dehghani, M.~Minderer, G.~Heigold, S.~Gelly \emph{et~al.},
  ``An image is worth 16x16 words: Transformers for image recognition at
  scale,'' \emph{arXiv preprint arXiv:2010.11929}, 2020.

\bibitem{ssrt}
T.~Sun, C.~Lu, T.~Zhang, and H.~Ling, ``Safe self-refinement for
  transformer-based domain adaptation,'' in \emph{Proceedings of the IEEE/CVF
  conference on computer vision and pattern recognition}, 2022, pp. 7191--7200.

\bibitem{liu2021swin}
Z.~Liu, Y.~Lin, Y.~Cao, H.~Hu, Y.~Wei, Z.~Zhang, S.~Lin, and B.~Guo, ``Swin
  transformer: Hierarchical vision transformer using shifted windows,'' in
  \emph{Proceedings of the IEEE/CVF international conference on computer
  vision}, 2021, pp. 10\,012--10\,022.

\bibitem{deit}
H.~Touvron, M.~Cord, M.~Douze, F.~Massa, A.~Sablayrolles, and H.~J{\'e}gou,
  ``Training data-efficient image transformers \& distillation through
  attention,'' in \emph{International conference on machine learning}.\hskip
  1em plus 0.5em minus 0.4em\relax PMLR, 2021, pp. 10\,347--10\,357.

\bibitem{tqs}
B.~Fu, Z.~Cao, J.~Wang, and M.~Long, ``Transferable query selection for active
  domain adaptation,'' in \emph{Proceedings of the IEEE/CVF conference on
  computer vision and pattern recognition}, 2021, pp. 7272--7281.

\bibitem{eada}
B.~Xie, L.~Yuan, S.~Li, C.~H. Liu, X.~Cheng, and G.~Wang, ``Active learning for
  domain adaptation: An energy-based approach,'' in \emph{Proceedings of the
  AAAI conference on artificial intelligence}, vol.~36, no.~8, 2022, pp.
  8708--8716.

\bibitem{lecun2006tutorial}
Y.~LeCun, S.~Chopra, R.~Hadsell, M.~Ranzato, and F.~Huang, ``A tutorial on
  energy-based learning,'' \emph{Predicting structured data}, vol.~1, no.~0,
  2006.

\bibitem{duc}
M.~Xie, S.~Li, R.~Zhang, and C.~H. Liu, ``Dirichlet-based uncertainty
  calibration for active domain adaptation,'' \emph{arXiv preprint
  arXiv:2302.13824}, 2023.

\bibitem{dapm}
Z.~Du and J.~Li, ``Diffusion-based probabilistic uncertainty estimation for
  active domain adaptation,'' \emph{Advances in Neural Information Processing
  Systems}, vol.~36, 2024.

\bibitem{diana}
D.~Huang, J.~Li, W.~Chen, J.~Huang, Z.~Chai, and G.~Li, ``Divide and adapt:
  Active domain adaptation via customized learning,'' in \emph{Proceedings of
  the IEEE/CVF Conference on Computer Vision and Pattern Recognition}, 2023,
  pp. 7651--7660.

\bibitem{lada}
T.~Sun, C.~Lu, and H.~Ling, ``Local context-aware active domain adaptation,''
  in \emph{Proceedings of the IEEE/CVF International Conference on Computer
  Vision}, 2023, pp. 18\,634--18\,643.

\bibitem{wang2023mhpl}
F.~Wang, Z.~Han, Z.~Zhang, R.~He, and Y.~Yin, ``Mhpl: Minimum happy points
  learning for active source free domain adaptation,'' in \emph{Proceedings of
  the IEEE/CVF Conference on Computer Vision and Pattern Recognition}, 2023,
  pp. 20\,008--20\,018.

\bibitem{li2022source}
X.~Li, Z.~Du, J.~Li, L.~Zhu, and K.~Lu, ``Source-free active domain adaptation
  via energy-based locality preserving transfer,'' in \emph{Proceedings of the
  30th ACM international conference on multimedia}, 2022, pp. 5802--5810.

\bibitem{he2016deep}
K.~He, X.~Zhang, S.~Ren, and J.~Sun, ``Deep residual learning for image
  recognition,'' in \emph{Proceedings of the IEEE conference on computer vision
  and pattern recognition}, 2016, pp. 770--778.

\bibitem{coop}
K.~Zhou, J.~Yang, C.~C. Loy, and Z.~Liu, ``Learning to prompt for
  vision-language models,'' \emph{International Journal of Computer Vision},
  vol. 130, no.~9, pp. 2337--2348, 2022.

\bibitem{cocoop}
------, ``Conditional prompt learning for vision-language models,'' in
  \emph{Proceedings of the IEEE/CVF conference on computer vision and pattern
  recognition}, 2022, pp. 16\,816--16\,825.

\bibitem{maple}
M.~U. Khattak, H.~Rasheed, M.~Maaz, S.~Khan, and F.~S. Khan, ``Maple:
  Multi-modal prompt learning,'' in \emph{Proceedings of the IEEE/CVF
  Conference on Computer Vision and Pattern Recognition}, 2023, pp.
  19\,113--19\,122.

\bibitem{denseclip}
Y.~Rao, W.~Zhao, G.~Chen, Y.~Tang, Z.~Zhu, G.~Huang, J.~Zhou, and J.~Lu,
  ``Denseclip: Language-guided dense prediction with context-aware prompting,''
  in \emph{Proceedings of the IEEE/CVF Conference on Computer Vision and
  Pattern Recognition}, 2022, pp. 18\,082--18\,091.

\bibitem{houlsby2019parameter}
N.~Houlsby, A.~Giurgiu, S.~Jastrzebski, B.~Morrone, Q.~De~Laroussilhe,
  A.~Gesmundo, M.~Attariyan, and S.~Gelly, ``Parameter-efficient transfer
  learning for nlp,'' in \emph{International conference on machine
  learning}.\hskip 1em plus 0.5em minus 0.4em\relax PMLR, 2019, pp. 2790--2799.

\bibitem{sung2022vl}
Y.-L. Sung, J.~Cho, and M.~Bansal, ``Vl-adapter: Parameter-efficient transfer
  learning for vision-and-language tasks,'' in \emph{Proceedings of the
  IEEE/CVF conference on computer vision and pattern recognition}, 2022, pp.
  5227--5237.

\bibitem{clip-adapter}
P.~Gao, S.~Geng, R.~Zhang, T.~Ma, R.~Fang, Y.~Zhang, H.~Li, and Y.~Qiao,
  ``Clip-adapter: Better vision-language models with feature adapters,''
  \emph{International Journal of Computer Vision}, pp. 1--15, 2023.

\bibitem{grave2017unbounded}
E.~Grave, M.~M. Cisse, and A.~Joulin, ``Unbounded cache model for online
  language modeling with open vocabulary,'' \emph{Advances in neural
  information processing systems}, vol.~30, 2017.

\bibitem{tip-adapter}
R.~Zhang, R.~Fang, W.~Zhang, P.~Gao, K.~Li, J.~Dai, Y.~Qiao, and H.~Li,
  ``Tip-adapter: Training-free clip-adapter for better vision-language
  modeling,'' \emph{arXiv preprint arXiv:2111.03930}, 2021.

\bibitem{udandarao2023sus}
V.~Udandarao, A.~Gupta, and S.~Albanie, ``Sus-x: Training-free name-only
  transfer of vision-language models,'' in \emph{Proceedings of the IEEE/CVF
  International Conference on Computer Vision}, 2023, pp. 2725--2736.

\bibitem{fahim2024its}
A.~Fahim, A.~Murphy, and A.~Fyshe, ``Its not a modality gap: Characterizing and
  addressing the contrastive gap,'' \emph{arXiv preprint arXiv:2405.18570},
  2024.

\bibitem{ramasinghe2024accept}
S.~Ramasinghe, V.~Shevchenko, G.~Avraham, and A.~Thalaiyasingam, ``Accept the
  modality gap: An exploration in the hyperbolic space,'' in \emph{Proceedings
  of the IEEE/CVF Conference on Computer Vision and Pattern Recognition}, 2024,
  pp. 27\,263--27\,272.

\bibitem{bousmalis2016domain}
K.~Bousmalis, G.~Trigeorgis, N.~Silberman, D.~Krishnan, and D.~Erhan, ``Domain
  separation networks,'' \emph{Advances in neural information processing
  systems}, vol.~29, 2016.

\bibitem{debiaspl}
X.~Wang, Z.~Wu, L.~Lian, and S.~X. Yu, ``Debiased learning from naturally
  imbalanced pseudo-labels,'' in \emph{Proceedings of the IEEE/CVF Conference
  on Computer Vision and Pattern Recognition}, 2022, pp. 14\,647--14\,657.

\bibitem{deepcluster}
M.~Caron, P.~Bojanowski, A.~Joulin, and M.~Douze, ``Deep clustering for
  unsupervised learning of visual features,'' in \emph{Proceedings of the
  European conference on computer vision (ECCV)}, 2018, pp. 132--149.

\bibitem{ahmed2021unsupervised}
S.~M. Ahmed, D.~S. Raychaudhuri, S.~Paul, S.~Oymak, and A.~K. Roy-Chowdhury,
  ``Unsupervised multi-source domain adaptation without access to source
  data,'' in \emph{Proceedings of the IEEE/CVF conference on computer vision
  and pattern recognition}, 2021, pp. 10\,103--10\,112.

\bibitem{msgd}
H.~Xia, T.~Jing, and Z.~Ding, ``Maximum structural generation discrepancy for
  unsupervised domain adaptation,'' \emph{IEEE Transactions on Pattern Analysis
  and Machine Intelligence}, vol.~45, no.~3, pp. 3434--3445, 2022.

\bibitem{kuda}
T.~Sun, C.~Lu, and H.~Ling, ``Prior knowledge guided unsupervised domain
  adaptation,'' in \emph{European Conference on Computer Vision}.\hskip 1em
  plus 0.5em minus 0.4em\relax Springer, 2022, pp. 639--655.

\bibitem{eidco}
Y.~Zhang, Z.~Wang, J.~Li, J.~Zhuang, and Z.~Lin, ``Towards effective instance
  discrimination contrastive loss for unsupervised domain adaptation,'' in
  \emph{Proceedings of the IEEE/CVF International Conference on Computer
  Vision}, 2023, pp. 11\,388--11\,399.

\bibitem{yue2023make}
Z.~Yue, Q.~Sun, and H.~Zhang, ``Make the u in uda matter: Invariant consistency
  learning for unsupervised domain adaptation,'' \emph{Advances in Neural
  Information Processing Systems}, vol.~36, pp. 26\,991--27\,004, 2023.

\bibitem{bai2024prompt}
S.~Bai, M.~Zhang, W.~Zhou, S.~Huang, Z.~Luan, D.~Wang, and B.~Chen,
  ``Prompt-based distribution alignment for unsupervised domain adaptation,''
  in \emph{Proceedings of the AAAI Conference on Artificial Intelligence},
  vol.~38, no.~2, 2024, pp. 729--737.

\bibitem{tvt}
J.~Yang, J.~Liu, N.~Xu, and J.~Huang, ``Tvt: Transferable vision transformer
  for unsupervised domain adaptation,'' in \emph{Proceedings of the IEEE/CVF
  Winter Conference on Applications of Computer Vision}, 2023, pp. 520--530.

\bibitem{venkateswara2017deep}
H.~Venkateswara, J.~Eusebio, S.~Chakraborty, and S.~Panchanathan, ``Deep
  hashing network for unsupervised domain adaptation,'' in \emph{Proceedings of
  the IEEE conference on computer vision and pattern recognition}, 2017, pp.
  5018--5027.

\bibitem{peng2017visda}
X.~Peng, B.~Usman, N.~Kaushik, J.~Hoffman, D.~Wang, and K.~Saenko, ``Visda: The
  visual domain adaptation challenge,'' \emph{arXiv preprint arXiv:1710.06924},
  2017.

\bibitem{domainnet}
X.~Peng, Q.~Bai, X.~Xia, Z.~Huang, K.~Saenko, and B.~Wang, ``Moment matching
  for multi-source domain adaptation,'' in \emph{Proceedings of the IEEE/CVF
  international conference on computer vision}, 2019, pp. 1406--1415.

\bibitem{saito2019semi}
K.~Saito, D.~Kim, S.~Sclaroff, T.~Darrell, and K.~Saenko, ``Semi-supervised
  domain adaptation via minimax entropy,'' in \emph{Proceedings of the IEEE/CVF
  international conference on computer vision}, 2019, pp. 8050--8058.

\bibitem{litrico2023guiding}
M.~Litrico, A.~Del~Bue, and P.~Morerio, ``Guiding pseudo-labels with
  uncertainty estimation for source-free unsupervised domain adaptation,'' in
  \emph{Proceedings of the IEEE/CVF Conference on Computer Vision and Pattern
  Recognition}, 2023, pp. 7640--7650.

\bibitem{koh2021wilds}
P.~W. Koh, S.~Sagawa, H.~Marklund, S.~M. Xie, M.~Zhang, A.~Balsubramani, W.~Hu,
  M.~Yasunaga, R.~L. Phillips, I.~Gao \emph{et~al.}, ``Wilds: A benchmark of
  in-the-wild distribution shifts,'' in \emph{International conference on
  machine learning}.\hskip 1em plus 0.5em minus 0.4em\relax PMLR, 2021, pp.
  5637--5664.

\bibitem{chi2024adapting}
Z.~Chi, L.~Gu, T.~Zhong, H.~Liu, Y.~Yu, K.~N. Plataniotis, and Y.~Wang,
  ``Adapting to distribution shift by visual domain prompt generation,''
  \emph{arXiv preprint arXiv:2405.02797}, 2024.

\bibitem{wang2020learning}
H.~Wang, M.~Xu, B.~Ni, and W.~Zhang, ``Learning to combine: Knowledge
  aggregation for multi-source domain adaptation,'' in \emph{Computer
  Vision--ECCV 2020: 16th European Conference, Glasgow, UK, August 23--28,
  2020, Proceedings, Part VIII 16}.\hskip 1em plus 0.5em minus 0.4em\relax
  Springer, 2020, pp. 727--744.

\bibitem{chen2023multi}
H.~Chen, X.~Han, Z.~Wu, and Y.-G. Jiang, ``Multi-prompt alignment for
  multi-source unsupervised domain adaptation,'' \emph{Advances in Neural
  Information Processing Systems}, vol.~36, pp. 74\,127--74\,139, 2023.

\bibitem{clue}
V.~Prabhu, A.~Chandrasekaran, K.~Saenko, and J.~Hoffman, ``Active domain
  adaptation via clustering uncertainty-weighted embeddings,'' in
  \emph{Proceedings of the IEEE/CVF international conference on computer
  vision}, 2021, pp. 8505--8514.

\bibitem{lora}
E.~J. Hu, Y.~Shen, P.~Wallis, Z.~Allen-Zhu, Y.~Li, S.~Wang, L.~Wang, W.~Chen
  \emph{et~al.}, ``Lora: Low-rank adaptation of large language models.''
  \emph{ICLR}, vol.~1, no.~2, p.~3, 2022.

\bibitem{hwang2022combating}
S.~Hwang, S.~Lee, S.~Kim, J.~Ok, and S.~Kwak, ``Combating label distribution
  shift for active domain adaptation,'' in \emph{European Conference on
  Computer Vision}.\hskip 1em plus 0.5em minus 0.4em\relax Springer, 2022, pp.
  549--566.

\bibitem{badge}
J.~T. Ash, C.~Zhang, A.~Krishnamurthy, J.~Langford, and A.~Agarwal, ``Deep
  batch active learning by diverse, uncertain gradient lower bounds,'' in
  \emph{International Conference on Learning Representations}, 2020.

\bibitem{shu2022hub}
Y.~Shu, Z.~Cao, Z.~Zhang, J.~Wang, and M.~Long, ``Hub-pathway: transfer
  learning from a hub of pre-trained models,'' \emph{Advances in Neural
  Information Processing Systems}, vol.~35, pp. 32\,913--32\,927, 2022.

\bibitem{beery2021iwildcam}
S.~Beery, A.~Agarwal, E.~Cole, and V.~Birodkar, ``The iwildcam 2021 competition
  dataset,'' \emph{arXiv preprint arXiv:2105.03494}, 2021.

\bibitem{bandi2018detection}
P.~Bandi, O.~Geessink, Q.~Manson, M.~Van~Dijk, M.~Balkenhol, M.~Hermsen, B.~E.
  Bejnordi, B.~Lee, K.~Paeng, A.~Zhong \emph{et~al.}, ``From detection of
  individual metastases to classification of lymph node status at the patient
  level: the camelyon17 challenge,'' \emph{IEEE transactions on medical
  imaging}, vol.~38, no.~2, pp. 550--560, 2018.

\bibitem{christie2018functional}
G.~Christie, N.~Fendley, J.~Wilson, and R.~Mukherjee, ``Functional map of the
  world,'' in \emph{Proceedings of the IEEE Conference on Computer Vision and
  Pattern Recognition}, 2018, pp. 6172--6180.

\bibitem{arjovsky2019invariant}
M.~Arjovsky, L.~Bottou, I.~Gulrajani, and D.~Lopez-Paz, ``Invariant risk
  minimization,'' \emph{arXiv preprint arXiv:1907.02893}, 2019.

\bibitem{zhong2022meta}
T.~Zhong, Z.~Chi, L.~Gu, Y.~Wang, Y.~Yu, and J.~Tang, ``Meta-dmoe: Adapting to
  domain shift by meta-distillation from mixture-of-experts,'' \emph{Advances
  in Neural Information Processing Systems}, vol.~35, pp. 22\,243--22\,257,
  2022.

\bibitem{li2024agile}
X.~Li, J.~Li, F.~Li, L.~Zhu, and K.~Lu, ``Agile multi-source-free domain
  adaptation,'' in \emph{Proceedings of the AAAI Conference on Artificial
  Intelligence}, vol.~38, no.~12, 2024, pp. 13\,673--13\,681.

\bibitem{van2008visualizing}
L.~Van~der Maaten and G.~Hinton, ``Visualizing data using t-sne.''
  \emph{Journal of machine learning research}, vol.~9, no.~11, 2008.

\end{thebibliography}
%

%

\begin{IEEEbiography}[{\includegraphics[width=1in,height=1.25in,clip,keepaspectratio]{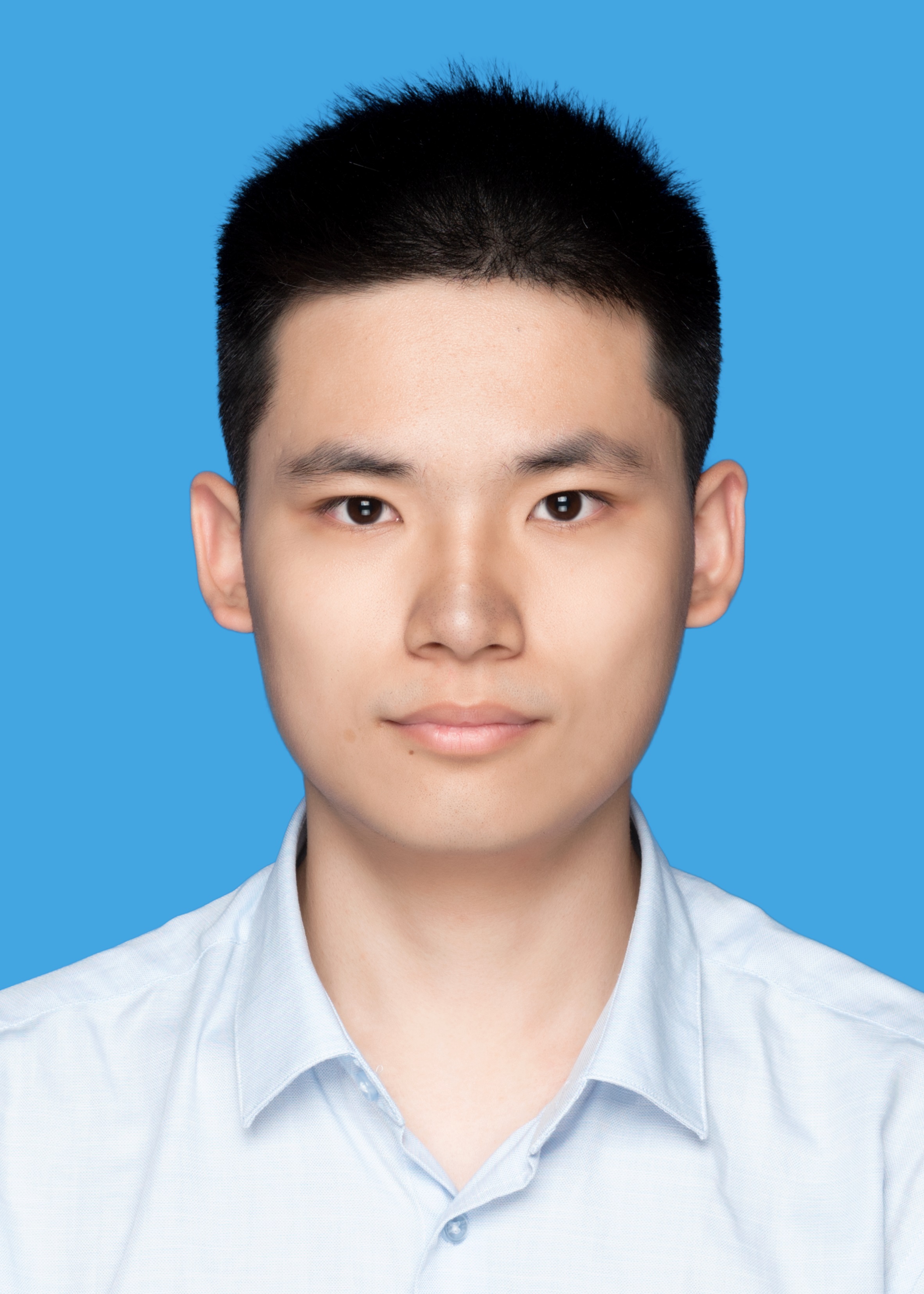}}]{Xinyao Li}
is currently working towards the PhD degree with School of Computer Science and Engineering, University of Electronic Science and Technology of China. His current research interest  focuses on transfer learning, deep learning, parameter-efficient fine-tuning and vision-language models. 
\end{IEEEbiography}

\begin{IEEEbiography}[{\includegraphics[width=1in,height=1.25in,clip,keepaspectratio]{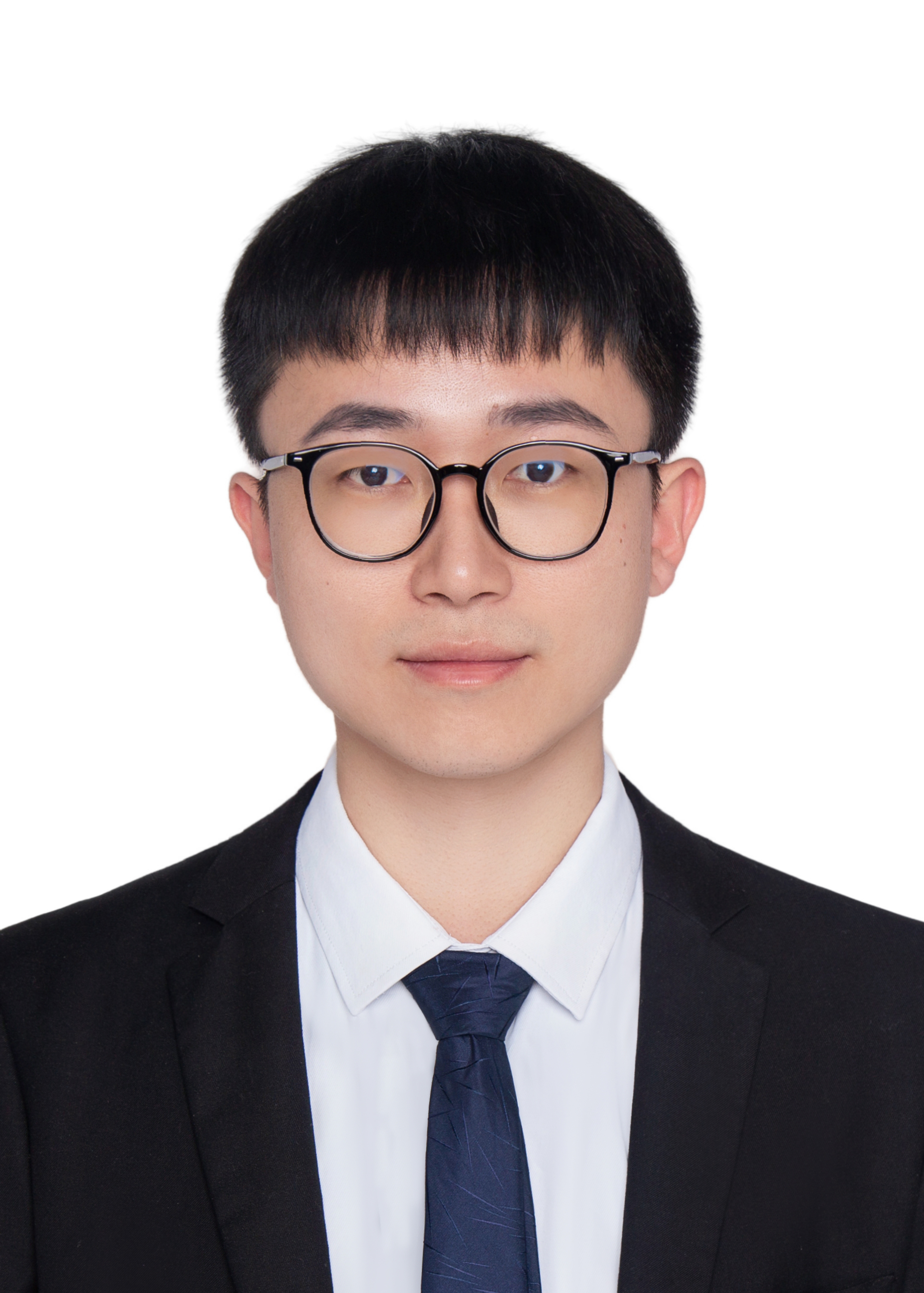}}]{Zhekai Du}
is currently a last-year Ph.D. student with the School of Computer Science and Engineering, University of Electronic Science and Technology of China (UESTC). His research interests are domain adaptation, domain generalization and their applications in computer vision. He received the bachelor degree from UESTC in 2018. 
\end{IEEEbiography}

\begin{IEEEbiography}[{\includegraphics[width=1in,height=1.25in,clip,keepaspectratio]{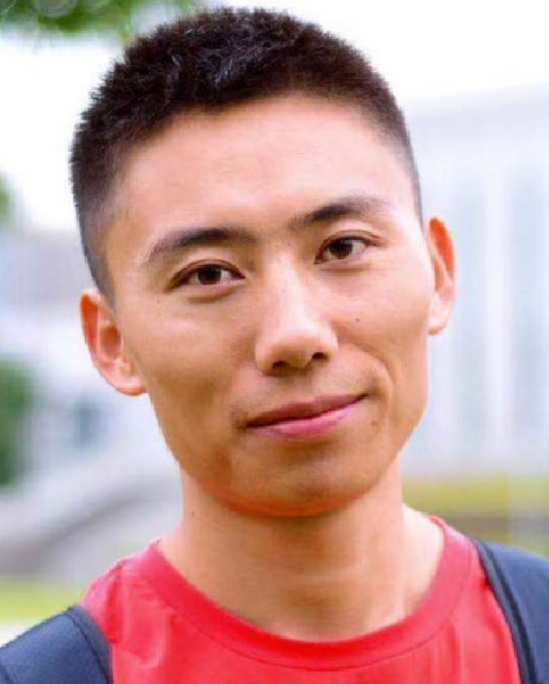}}]{Jingjing Li}
received the MSc and PhD degrees in computer science from the University of Electronic Science and Technology of China, in 2013 and 2017, respectively. Now, he is a national Postdoctoral Program for Innovative Talents research fellow with the School of Computer Science and Engineering, University of Electronic Science and Technology of China. He has great interest in machine learning, especially transfer learning, domain adaptation, and recommender systems.
\end{IEEEbiography}

\begin{IEEEbiography}[{\includegraphics[width=1in,height=1.25in,clip,keepaspectratio]{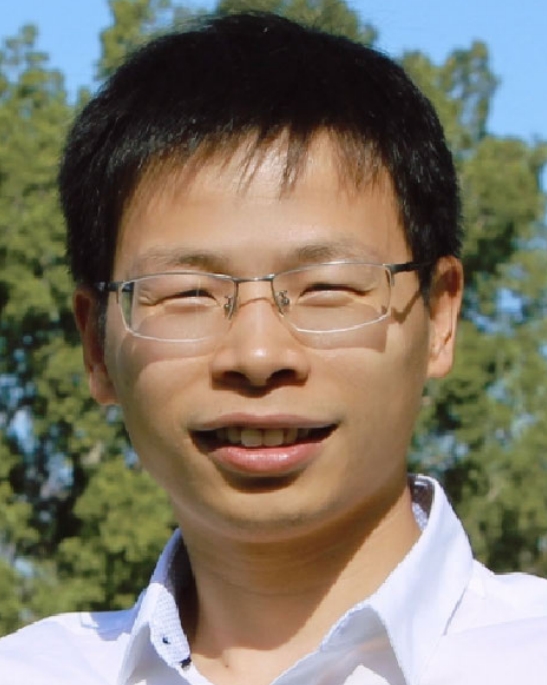}}]{Lei Zhu}
received the BEng degree from the Wuhan University of Technology, in 2009, and the PhD degree from Huazhong University Science and Technology, in 2015. He is currently a professor with the School of Computer Science and Technology, Tongji University. He was a postdoctoral research fellow with the University of Queensland (2016–2017). His research interests include the research area of large-scale multimedia content analysis and retrieval. He has co-/authored more than 100 peer-reviewed papers, such as ACM SIGIR, ACM MM, IEEE TPAMI, IEEE TIP, IEEE TKDE, and ACM TOIS. 
At present, he serves as the associate editor of the IEEE Transactions on Big Data and ACM Transactions on Multimedia Computing, Communications, and Applications. He has served as the area chair for ACM MM and IEEE ICME, Senior Program Committee for SIGIR, AAAI, and CIKM. He won ACM SIGIR 2019 Best Paper Honorable Mention Award, ADMA 2020 Best Paper Award, ChinaMM 2022 Best Student Paper Award, ACM China SIGMM Rising Star Award, Shandong Provincial Entrepreneurship Award for Returned Students, and Shandong Provincial AI Outstanding Youth Award.
\end{IEEEbiography}

\begin{IEEEbiography}[{\includegraphics[width=1in,height=1.25in,clip,keepaspectratio]{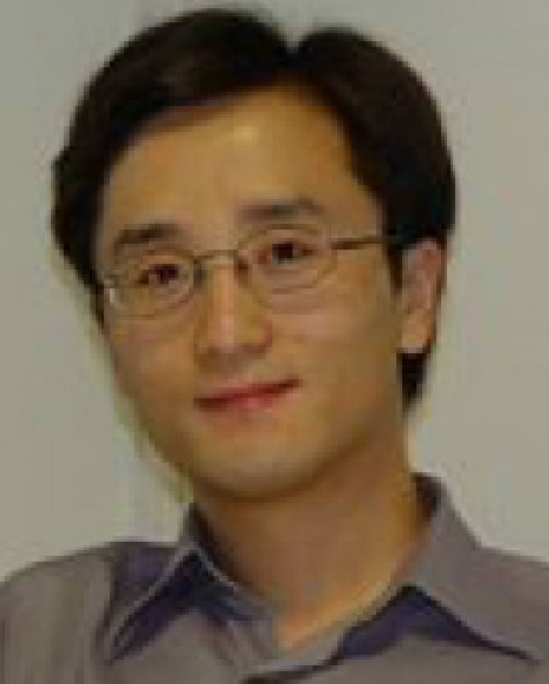}}]{Heng Tao Shen}
(Fellow, IEEE) received the BSc (1st class honours) and PhD degrees from the Department of Computer Science, National University of Singapore, in 2000 and 2004, respectively. He is distinguished professor and dean with the School of Computer Science and Engineering, Executive Dean of AI Research Institute, and director of Centre for Future Media, University of Electronic Science and Technology of China. He then joined the University of Queensland and became a professor in late 2011. His research interests mainly include multimedia search, computer vision, artificial intelligence, and Big Data management.  He has published more than 350 peer-reviewed papers, including more than 130 IEEE/ACM Transactions, and received seven Best Paper Awards from international conferences, including the Best Paper Award from ACM Multimedia 2017 and Best Paper Award - Honorable Mention from ACM SIGIR 2017. He is/was an associate editor of the ACM Transactions of Data Science, IEEE TIP, IEEE TMM, and IEEE TKDE.
He is an OSA fellow and ACM fellow.
\end{IEEEbiography}


\end{document}